\RequirePackage{letltxmacro}
\LetLtxMacro{\LaTeXtextbf}{\textbf}
\documentclass{ieeeaccess}
\LetLtxMacro{\textbf}{\LaTeXtextbf}
\usepackage{cite}
\usepackage{amsmath,amssymb,amsfonts}
\usepackage{algorithmic}
\usepackage{graphicx}
\usepackage{textcomp}

\usepackage{bm}
\makeatletter
\AtBeginDocument{\DeclareMathVersion{bold}
\SetSymbolFont{operators}{bold}{T1}{times}{b}{n}
\SetSymbolFont{NewLetters}{bold}{T1}{times}{b}{it}
\SetMathAlphabet{\mathrm}{bold}{T1}{times}{b}{n}
\SetMathAlphabet{\mathit}{bold}{T1}{times}{b}{it}
\SetMathAlphabet{\mathbf}{bold}{T1}{times}{b}{n}
\SetMathAlphabet{\mathtt}{bold}{OT1}{pcr}{b}{n}
\SetSymbolFont{symbols}{bold}{OMS}{cmsy}{b}{n}
\renewcommand\boldmath{\@nomath\boldmath\mathversion{bold}}}
\makeatother

\def\BibTeX{{\rm B\kern-.05em{\sc i\kern-.025em b}\kern-.08em
    T\kern-.1667em\lower.7ex\hbox{E}\kern-.125emX}}

\usepackage[font={sf,scriptsize},           % new
            labelfont={bf,color=accessblue},% new
            caption=false]                  % new
           {subfig} 

\usepackage{colortbl}
\usepackage{amsthm}
\usepackage{booktabs}
\newtheorem{definition}{Definition}
\usepackage{microtype}
\usepackage{pifont}
\usepackage[vlined,ruled,linesnumbered,noend]{algorithm2e} % Import algorithm2e package
\usepackage{cleveref}
\usepackage{nicefrac}
\usepackage{url}
\usepackage{siunitx}
\usepackage{multirow}
\usepackage[flushleft]{threeparttable}
\usepackage{longtable}
\usepackage{bbm}

\newcommand{\Cts}{x}  % ts
\newcommand{\Ctab}{z}  % tabular data
\newcommand{\Csal}{\psi}  % saliency map
\newcommand{\Cfeatimp}{\phi}  % feature importance
\newcommand{\Cmean}{\mu} 
 
\newcommand{\Cstd}{\sigma} 
 
\newcommand{\Cvector}[1]{\textbf{#1}}
\newcommand{\Cmatrix}[1]{\MakeUppercase{#1}}
\newcommand{\Ctensor}[1]{\mathcal{\MakeUppercase{#1}}}

\newcommand{\Csignal}{\Cvector{\Cts}}
\newcommand{\Cmts}{\Cmatrix{\Cts}}
\newcommand{\Ctsdata}{\Ctensor{\Cts}}

\newcommand{\Csaliencymatrix}{\Psi}

\newcommand{\Cfeatimpmatrix}{\Phi}

\newcommand{\Cmeans}{\boldsymbol{\Cmean}}
\newcommand{\Cstds}{\boldsymbol{\Cstd}}

\newcommand{\Cnpoints}{m}  % # of observations
\newcommand{\Cnsignals}{c}  % # of signals
\newcommand{\Cnmts}{n}  % # of time series 
\newcommand{\Cnsubseq}{p}  % # of receptive fields
\newcommand{\Cwindowsize}{w}  % widow size (or subsequence length)
\newcommand{\Cwindowsizes}{\Cvector{\Cwindowsize}}  % widow sizes (or subsequence lengths)
\newcommand{\Cdilation}{d}  % dilation
\newcommand{\Cdilations}{\Cvector{\Cdilation}}  % dilations
\newcommand{\Cstride}{s}  % stride
\newcommand{\Cwordlength}{l}  % word length (# of segments)
\newcommand{\Cwordlengths}{\Cvector{\Cwordlength}}  % word lengths (# of segments)
\newcommand{\Cseglength}{q}  % length of each segment in a word (# of points in a segment)

\newcommand{\Calphabet}{\alpha}  % # of SAX symbols

\newcommand{\Clabel}{y}  % label
\newcommand{\Clabels}{\Cvector{\Clabel}}  % labels vector
\newcommand{\Ctabdata}{\Cmatrix{\Ctab}}  % tabular dataset
\newcommand{\Ctabtensor}{\Ctensor{\Ctab}}  % tensor

\newcommand{\Cmethod}[1]{\uppercase{\texttt{#1}}}

\newcommand{\Ccdot}{\cdot}

\newcommand{\Cnwindows}{v}

\newcommand{\Cmethodname}{{\Cmethod{BORF}}}

\newcommand{\cmark}{\ding{51}}%
\newcommand{\xmark}{\ding{55}}%

\begin{document}
%\history{Date of publication xxxx 00, 0000, date of current version xxxx 00, 0000.}
%\doi{10.1109/ACCESS.2024.0429000}
\doi{10.1109/ACCESS.2024.3464743}
\vol{12}
\year{2024}
\history{This article has been accepted for publication in IEEE Access. Content may differ from the final edited publication.}

\title{Fast, Interpretable, and Deterministic Time Series Classification With a Bag-of-Receptive-Fields}
\author{\uppercase{Francesco Spinnato}\authorrefmark{1,2}, %\IEEEmembership{Fellow, IEEE},
\uppercase{Riccardo Guidotti}\authorrefmark{1,2}, \uppercase{Anna Monreale}\authorrefmark{1}, and \uppercase{Mirco Nanni}\authorrefmark{2}
%\IEEEmembership{Member, IEEE}
}

\address[1]{Department of Computer Science, University of Pisa, Largo B. Pontecorvo, 3, 56127 Pisa, Italy  (e-mail: \{name.surname\}@unipi.it)}
\address[2]{ISTI-CNR, Via Giuseppe Moruzzi, 1, 56124 Pisa, Italy (e-mail: \{name.surname\}@isti.cnr.it)}

\tfootnote{This work has been partially supported by the EU H2020 programme under the funding schemes ERC-2018-ADG G.A. 834756 ``XAI: Science and technology for the eXplanation of AI decision making'', ``SoBigData++: European Integrated Infrastructure for Social Mining and Big Data Analytics'', G.A. 101120763 TANGO, by the European Commission under the NextGeneration EU programme – National Recovery and Resilience Plan (Piano Nazionale di Ripresa e Resilienza, PNRR) – Project: ``SoBigData.it – Strengthening the Italian RI for Social Mining and Big Data Analytics'' – Prot. IR0000013 – Avviso n. 3264 del 28/12/2021, and M4C2 - Investimento 1.3, Partenariato Esteso PE00000013 - ``FAIR - Future Artificial Intelligence Research'' - Spoke 1 ``Human-centered AI'', and by the Italian Project Fondo Italiano per la Scienza FIS00001966 MIMOSA.}

% \markboth
% {Author \headeretal: Preparation of Papers for IEEE TRANSACTIONS and JOURNALS}
% {Author \headeretal: Preparation of Papers for IEEE TRANSACTIONS and JOURNALS}

\markboth
{F. Spinnato \headeretal: Fast, Interpretable, and Deterministic Time Series Classification With a Bag-of-Receptive-Fields}
{F. Spinnato \headeretal: Fast, Interpretable, and Deterministic Time Series Classification With a Bag-of-Receptive-Fields}

\corresp{Corresponding author: Francesco Spinnato (e-mail: francesco.spinnato@di.unipi.it).}

\begin{abstract}
    The current trend in the literature on Time Series Classification is to develop increasingly accurate algorithms by combining multiple models in ensemble hybrids, representing time series in complex and expressive feature spaces, and extracting features from different representations of the same time series. As a consequence of this focus on predictive performance, the best time series classifiers are black-box models, which are not understandable from a human standpoint. Even the approaches that are regarded as interpretable, such as shapelet-based ones, rely on randomization to maintain computational efficiency. This poses challenges for interpretability, as the explanation can change from run to run.
    Given these limitations, we propose the Bag-Of-Receptive-Field (BORF), a fast, interpretable, and deterministic time series transform. Building upon the classical Bag-Of-Patterns, we bridge the gap between convolutional operators and discretization, enhancing the Symbolic Aggregate Approximation (SAX) with dilation and stride, which can more effectively capture temporal patterns at multiple scales. We propose an algorithmic speedup that reduces the time complexity associated with SAX-based classifiers, allowing the extension of the Bag-Of-Patterns to the more flexible Bag-Of-Receptive-Fields, represented as a sparse multivariate tensor. The empirical results from testing our proposal on more than 150 univariate and multivariate classification datasets demonstrate good accuracy and great computational efficiency compared to traditional SAX-based methods and state-of-the-art time series classifiers, while providing easy-to-understand explanations.
    
\end{abstract}


\titlepgskip=-21pt

\maketitle

\section{Introduction}
\label{sec:introduction}
The availability of high-dimensional time series data has significantly expanded across critical domains such as finance, healthcare, and environmental science, capturing dynamic changes over time and presenting both opportunities and challenges for time series classification (TSC).

The current trend in this field, highlighted by the most recent ``bake-off'' survey paper~\cite{middlehurst2024bakeredux}, is to discover increasingly better-performing algorithms. This is achieved through various techniques, including combining multiple models in ensembles~\cite{dempster2023hydra,tan2022multirocket,middlehurst2021hive}, representing time series in expressive feature spaces~\cite{dempster2020rocket,dempster2021minirocket,middlehurst2020canonical}, and extracting features from different representations of the same time series signal, such as differencing and Fourier transform~\cite{schafer2023weasel}, as well as autoregressive coefficients~\cite{middlehurst2023extracting}.

As a consequence of this accuracy arms race, currently, the top-performing time series classifiers are black-box models. These models lack interpretability either due to their inherently complex structures, such as neural networks~\cite{ismail2020inceptiontime}, or because, even when using relatively simple classifiers like tree ensembles or linear models, they leverage an uninterpretable feature space~\cite{dempster2020rocket,tan2022multirocket}. 
Even the most well-known interpretable models for TSC, such as shapelet and interval-based algorithms~\cite{guillaume2022random,cabello2024fast} rely on randomization to maintain reasonable computational efficiency, which poses challenges for interpretability as explanations can vary significantly between different runs~\cite{theissler2022explainable}.

A classical interpretable and deterministic approach for TSC is the so-called Bag-Of-Patterns (BOP)~\cite{baydogan2013bag}, analogous to the Bag-Of-Words approach~\cite{harris1954distributional} in text analysis, adapted for the time series domain. BOP utilizes the Symbolic Aggregate approXimation (SAX)~\cite{lin2007experiencing} with a sliding window to transform time series data into symbolic subsequences, also called words. The counts of these words serve as input features for predicting categorical outputs. This window-wise discretization enhances data representation expressiveness~\cite{senin2013sax,le2019interpretable,nguyen2022fast}, maintains interpretability, but imposes higher computational demands~\cite{bagnall2017great}, particularly for longer sequences, thereby making it less feasible for very large datasets.
In summary, to the best of our knowledge, the literature lacks a \textit{fast}, \textit{interpretable}, and above all, \textit{deterministic} time series classifier capable of achieving good accuracy across a broad spectrum of tasks. 

Given these challenges, our contributions are as follows.
%\textit{(i)} We 
First, we formalize window-wise SAX, a technique widely adopted in experimental settings in the literature~\cite{senin2013sax,nguyen2022fast,le2019interpretable}, but lacking systematic presentation and definition. 
We extend SAX with \textit{dilation} and \textit{stride} convolutional operators to extract \textit{receptive fields}. 
Unlike standard subsequences, receptive fields capture the temporal evolution of a time series at multiple resolutions, emphasizing both local and global characteristics. Dilation enables skipping points within a subsequence, allowing for the detection of broader temporal patterns without additional computational overhead, while stride defines the overlap between consecutive receptive fields.

Second, we show that current Piecewise Aggregate Approximation (PAA)~\cite{hung2007combining} and SAX-based classifiers exhibit quadratic worst-case time complexity with respect to the number of observations in a time series. 
We propose a deterministic speedup, enhancing the algorithm scalability over long time series.

Third, leveraging this speedup, we introduce an efficient and deterministic transform, named Bag-Of-Receptive-Fields (\Cmethodname{}), capable of converting both univariate and multivariate time series datasets into a human-understandable tabular representation. Unlike the original Bag-Of-Patterns approach~\cite{baydogan2013bag}, \Cmethodname{} represents subsequence counts as a sparse multivariate tensor rather than a dense matrix, offering greater flexibility for handling multivariate data.

Finally, we demonstrate that the integration of convolutional operators and multi-resolution symbolic patterns in \Cmethodname{} achieves competitive accuracy in TSC, while our optimization significantly enhances computational efficiency. 
We evaluate \Cmethodname{} across the complete UEA and UCR machine learning repositories, comprising over 150 datasets, benchmarking against SAX-based methods initially and subsequently against several state-of-the-art classifiers. To conclude, we assess the interpretability of our proposed approach, which is a combination of an interpretable feature space, i.e., the Bag-Of-Receptive-Fields, paired with a simple linear classifier.

The rest of this paper is structured as follows: \Cref{sec:related} discusses related work and principal competitors, while \Cref{sec:background} provides a background overview of our proposed methodology. The \Cmethodname{} approach is detailed in \Cref{sec:method}, followed by experimental results and analysis in \Cref{sec:experiments}. Finally, conclusions are drawn in \Cref{sec:conclusion}.

\section{Related Work}
\label{sec:related}
Time Series Classification (TSC) has been widely studied in recent years, with several so-called ``bake-offs,''~\cite{bagnall2017great,ruiz2021great,middlehurst2024bakeredux} i.e., survey papers benchmarking the performance of the most famous time series classifier from different categories. 
Our proposal belongs to the dictionary-based family, which extracts features from time series by recording characteristics of discretized subpatterns~\cite{bagnall2017great}. These methods segment time series into subsequences, convert them into symbolic words, and create histograms of feature counts~\cite{baydogan2013bag}. 
For this reason, they rely heavily on effective time series approximation methods such as the Symbolic Aggregate approXimation (SAX)~\cite{lin2007experiencing} and Symbolic Fourier Approximation (SFA)~\cite{schafer2012sfa}. 
SAX uses Piecewise Aggregate Approximation (PAA)~\cite{keogh2001dimensionality} for segmentation and binning, while SFA focuses on spectral properties using Fourier coefficients \cite{schafer2012sfa}.

SAX and SFA are often used in a window-based manner; that is, they are not applied to discretize the entire time series, but instead, a window is slid over the time series, converting each corresponding subsequence into a distinct word. 
This approach allows the extraction of scale-invariant local patterns in the sequence, given that each subsequence is normalized independently.
The first method using this approach is Bag-of-Patterns (\Cmethod{bop})~\cite{lin2012rotation}, an unsupervised transformation that uses SAX to generate discretized subsequences, counting their frequency.
Other methods rely on supervised techniques. For example, \Cmethod{sax-vsm}~\cite{senin2013sax} introduces a tf-idf weighting of features to improve classification accuracy.
\Cmethod{mr-seql}~\cite{le2019interpretable} uses the feature space of all subsequences in the training data to find useful features within SAX or SFA words. It employs greedy feature selection and a gradient bound to prune non-promising features. 
To improve time efficiency, \Cmethod{mr-sqm}~\cite{nguyen2022fast} also concatenates multiple symbolic representations at multiple resolutions, but uses a Chi-squared bound and random sampling to filter subsequences extracted with SAX and SFA. There are several issues with these approaches. First, their performance is still subpar with respect to the state-of-the-art~\cite{bagnall2017great,middlehurst2024bakeredux}. Second, their implementation does not consider multivariate time series. Finally, as shown in \Cref{sec:method}, their worst-case time complexity, is quadratic with respect to the length of the time series, which negates the possibility of using them for very long time series.
Better results in terms of accuracy are achieved by \Cmethod{weasel}, \Cmethod{weasel+muse}, and \Cmethod{weasel-V2}~\cite{schafer2017fast, schafer2017multivariate, schafer2023weasel}, which focus on the usage of SFA. However, although informative for certain applications, SFA has higher computational complexity than SAX~\cite{schafer2012sfa, nguyen2022fast}, also losing temporal information, which makes it difficult to interpret from a human standpoint. The state-of-the-art for dictionary-based approaches is a black-box called \Cmethod{hydra}~\cite{dempster2023hydra}, which convolves the time series with kernels and organizes the resulting activation maps into groups, counting the number of best matches in terms of kernel activation with the time series.

The most famous interpretable TSC approaches are based on shapelets~\cite{theissler2022explainable}. Shapelet-based approaches focus on finding subsequences that define a class independently of their position~\cite{ye2011time}. The discriminative features are distances, which can be used to convert the time series dataset into a tabular one through the so-called \textit{shapelet transform}~\cite{lines2012shapelet}. These approaches use different methods for extracting shapelets and building the subsequent classifier~\cite{grabocka2014learning, karlsson2016generalized}. 
Until recently, the Shapelet Transform (\Cmethod{st})~\cite{bostrom2015binary} was the most accurate shapelet approach, with the drawback of high computational complexity resulting in rather long training and inference times. 
However, according to~\cite{middlehurst2024bakeredux}, the newer Random Dilated Shapelet Transform (\Cmethod{rdst}) achieves better performance at a faster runtime, with the drawback of more randomness in the approach, which can hinder its interpretability~\cite{theissler2022explainable}.

Interval-based classifiers, such as Time Series Forest (\Cmethod{tsf})~\cite{deng2013time} divide time series into random intervals and extract summary statistics like mean, standard deviation, and slope, training a random forest on the resulting dataset. One of the most famous methods in this category is the Canonical Interval Forest (\Cmethod{cif})~\cite{middlehurst2020canonical} and its extension (\Cmethod{DrCIF}), which use the \Cmethod{catch22} feature set~\cite{lubba2019catch22} on randomly selected intervals from different transformations of the original time series, training a tree ensemble on this representation. The interpretability of these approaches depends on the feature set used, and, in general, explanations are in the form of highlighting the most relevant features extracted from discriminatory intervals~\cite{cabello2024fast}. Similar to the best-performing shapelet classifiers, these approaches heavily rely on randomization, causing variations in the areas of interest used by the classifier across runs. In contrast, our proposal is entirely deterministic, generating simple, repeatable representations.

As an interesting note, standard deep learning approaches often fall behind in general time series classification benchmarks~\cite{middlehurst2024bakeredux}, given that they usually require ad-hoc fine-tuning, often dependent on the specific task and dataset. An exception to this is \Cmethod{H-InceptionTime}~\cite{ismail2022deep}, an extension of \Cmethod{InceptionTime}~\cite{ismail2020inceptiontime}, which is an ensemble of deep Convolutional Neural Networks (CNNs). According to~\cite{middlehurst2024bakeredux}, it achieves state-of-the-art performance. Convolution is also used in \Cmethod{rocket}~\cite{dempster2020rocket} and its numerous variants~\cite{dempster2021minirocket,tan2022multirocket}, which are considered some of the best-performing models in TSC.
They generate many random convolutional kernels to transform time series into feature maps, which are pooled and used in a linear classifier. Even with large and complex datasets, \Cmethod{rocket} often achieves high accuracy and processing speed~\cite{middlehurst2024bakeredux}. 

Finally, the absolute top-performing models in TSC are ensemble hybrids of the aforementioned methods, such as \Cmethod{MultiRocket-Hydra}~\cite{dempster2023hydra}, \Cmethod{Hive-Cote-2}~\cite{middlehurst2021hive}, \Cmethod{TS-CHIEF}~\cite{shifaz2020ts}, and \Cmethod{RIST}~\cite{middlehurst2023extracting}. By their nature, these models are very complex and thus not interpretable from a human standpoint.

In summary, the current trend in the literature is to build increasingly complex and diverse feature spaces extracted from the original time series signal and several signal transformations, heavily relying on randomization to achieve accurate and fast classification. With \Cmethodname{}, we go against this trend by demonstrating that it is still possible to build an efficient, completely deterministic, easy-to-interpret representation grounded only in the time domain, while achieving good classification performance.

\section{Background}
\label{sec:background}
This section provides all the necessary concepts to understand our proposal. We begin by defining time series data\footnote{A summary of the notation is available in \Cref{sec:notation}.}.

\begin{definition}[Time Series Data]
    A \textit{time series dataset}, $\Ctsdata=\{\Cmts_{1}, \dots, \Cmts_{\Cnmts}\} \in \mathbb{R}^{\Cnmts \times \Cnsignals \times \Cnpoints}$, is a collection of $\Cnmts$ time series. A time series, $\Cmts$, is a collection of $\Cnsignals$ signals (or channels), $\Cmts = \{\Csignal_{1}, \dots, \Csignal_{\Cnsignals}\} \in {\mathbb{R}}^{\Cnsignals \times \Cnpoints}$. A signal, $\Csignal$, is a sequence of $\Cnpoints$ real-valued observations sampled regularly at equal time intervals,
    $\Csignal = [\Cts_{1}, \dots, \Cts_{\Cnpoints}] \in {\mathbb{R}}^{\Cnpoints}$.
\end{definition}
\noindent
When $\Cnsignals=1$, the time series is \textit{univariate}, for $\Cnsignals > 1$ it is \textit{multivariate}. 
Time series datasets can be used in a variety of tasks. 
This work focuses on supervised learning, particularly Time Series Classification (TSC), through a Bag-Of-Patterns (\Cmethod{bop}) representation~\cite{baydogan2013bag}.
\begin{definition}[TSC]
Given a time series dataset, $\Ctsdata$, \textit{Time Series Classification} is the task of training a model, $f$, to predict a categorical output, $\Clabel$, for each input time series, $\Cmts$, i.e., $f(\Ctsdata) = [f(\Cmts_1), \dots, f(\Cmts_\Cnmts)] = \Clabels\in\mathbb{N}^{\Cnmts}$. 
\end{definition}

\noindent Specifically, \Cmethod{bop} tackles classification by extracting symbolic patterns (or words) from the time series, converting them into a tabular representation, where the rows represent the time series, and the columns are the extracted patterns. Values in this matrix correspond to the count of appearances of each pattern in each time series. Formally:

\begin{definition}[Bag-of-Patterns]
\label{def:bop}
Given a time series dataset $\Ctsdata$  
and a set of $\Cnsubseq$ patterns, a Bag-of-Patterns is a dataset $\Ctabdata\in\mathbb{N}^{\Cnmts \times \Cnsubseq}$, where $\Ctab_{i,j}$ is the number of appearances of pattern $j$ in the time series $i$.
\end{definition}

\noindent Any classifier can use this dataset for TSC.
A common way of building a Bag-Of-Patterns is by time series discretization, which essentially converts signals into a symbolic form by approximating both the time and value axes. 
One of the most famous discretization algorithms is the Symbolic Aggregate Approximation (SAX)~\cite{lin2007experiencing}. 
SAX uses the Piecewise Aggregate Approximation (PAA)~\cite{keogh2001dimensionality} to segment a time series signal into equal-sized segments and then compute the mean value for each segment. 
Formally, the average of the $i$-th segment is:

\begin{equation}
\label{eq:paa}
    \Cmean_i = \frac{\Cwordlength}{\Cnpoints} \sum_{j= \frac{\Cnpoints}{\Cwordlength} (i-1)  + 1}^{ \frac{\Cnpoints}{\Cwordlength} i } \Cts_j,
\end{equation}
\noindent
where $\Cwordlength$ is defined as the word (or pattern) length, and $i\in[1, \Cwordlength]$ \cite{lin2007experiencing}. Therefore, the transformed signal is $\Cmeans=[\Cmean_1, \dots, \Cmean_\Cwordlength]\in\mathbb{R}^{\Cwordlength}$. 
This vector is then standardized into $\Cmeans^*$, with each element computed as $\Cmean_i^*=\textit{zscore}(\Cmean_i, \Cmean^\Csignal, \Cstd^\Csignal)=(\Cmean_i-\Cmean^\Csignal)/{\Cstd^\Csignal}$, where $\Cmean^\Csignal$ and $\Cstd^\Csignal$ are the mean and standard deviation of the full signal.

The last step in SAX is discretization through equal-frequency binning.
Values in $\Cmeans^*$ are quantized using a set of breakpoints, $\mathbf{b}=[b_1, \dots, b_{\Calphabet-1}]$, obtained on quantiles of the standard Gaussian distribution, which bin values in $\Calphabet\geq 2$ equiprobable symbols. The final SAX approximation is stored in $\tilde{\Cmeans}$, where each element is computed as:

\begin{equation}
\label{eq:sax}
    \tilde{\Cmean}_i=\textit{bin}(\Cmean_i^*, \Calphabet, \mathbf{b})=
\begin{cases} 
0 & \text{if } \Cmean^*_i \leq b_1, \\
k-1 & \text{if } b_{k-1} < \Cmean^*_i \leq b_k,\\
\Calphabet-1 & \text{if } \Cmean^*_i > b_{\Calphabet-1}.
\end{cases}
\end{equation}
\noindent
Since each signal element is accessed once, the cost of applying SAX to a time series signal of length $\Cnpoints$ is $O(\Cnpoints)$~\cite{lin2007experiencing}. 

However, several TSC algorithms in the literature~\cite{baydogan2013bag,senin2013sax,le2019interpretable,nguyen2022fast} iterate the aforementioned process using a sliding window, thus converting each time series signal into a collection of SAX words, which become the patterns used to build of a Bag-Of-Patterns.

\section{Bag-of-Receptive-Fields}
\label{sec:method}
In this section, we present \Cmethodname{}, Bag-Of-Receptive-Fields, our proposal to efficiently extract an interpretable feature space that any machine learning classifier can use for TSC.
\Cmethodname{} extends the notion of Bag-Of-Patterns, representing time series as a sparse multivariate tensor instead of a simple matrix. 
We introduce here a connection between convolutional operators and pattern extraction by generalizing the concept of a window to that of a \textit{receptive field}. 
In convolutional neural networks theory, the receptive field refers to the spatial extent of input data that influences the activation of a particular neuron in the network~\cite{dumoulin2016guide}.
Applied to our setting, a receptive field can be viewed as a generalized time series subsequence. %, formally:

\begin{definition}[Receptive Field]
    Given a signal $\Csignal$, the $i$-th \textit{receptive field} of length $\Cwindowsize \geq 1$ and dilation $\Cdilation \geq 1$, is an ordered sequence of values, 
    $[\Cts_{i}, \Cts_{i+\Cdilation}, \dots, \Cts_{i+\Cdilation\cdot(\Cwindowsize-1)}]$.
\end{definition}

\noindent The dilation, $\Cdilation$, plays an essential role in convolution, offering control over receptive field size and determining the spacing elements, enabling the capture of long-range dependencies and patterns. 
By introducing gaps between observations, dilation expands the receptive field and facilitates the incorporation of distant information. 
By iterating over the signal $\Csignal$, $\Cnwindows$ receptive fields can be extracted~\cite{dumoulin2016guide}, where:
\begin{align}
    \Cnwindows=1+\left\lfloor\frac{{(\Cnpoints-\Cwindowsize-(\Cdilation-1)(\Cwindowsize-1))}}{{\Cstride}}\right\rfloor.
\end{align}
In this formulation, the stride, $\Cstride\geq 1$, is essentially the number of elements skipped when iterating over the signal, i.e., the distance between consecutive receptive fields.
A receptive field simplifies to a subsequence obtained via sliding window when $\Cdilation=1$ and $\Cstride=1$. 
Without loss of generality, we use the terms receptive field, pattern, subsequence, and window interchangeably.

In the following, first, we connect the notion of receptive fields to SAX discretization, proposing a way to lower the time complexity of extracting SAX words (\Cref{sec:wsax}). 
Then, in \Cref{sec:borf}, we introduce the Bag-Of-Receptive-Fields, extending \Cref{def:bop}, by considering receptive fields instead of standard subsequences for extracting symbolic patterns, and by building this representation as a sparse 3D tensor, which is able to represent multivariate time series data. 
This extension of SAX through dilation and stride allows for building a representative feature space while maintaining interpretability. In fact, dilation and stride only change the shape of a subsequence, such as by skipping time series observations, but they maintain their semantics. This ensures that these generalized subsequences remain meaningful while capturing different resolutions of the original time series data.
Finally, in \Cref{sec:compl,sec:tsc,sec:xai}, we discuss the complexity of the overall approach in detail and its applicability to TSC, also from an interpretability standpoint.

\begin{figure}[t]
    \centering
    \includegraphics[width=0.9\linewidth, trim = 0mm 0mm 0mm 0mm, clip]{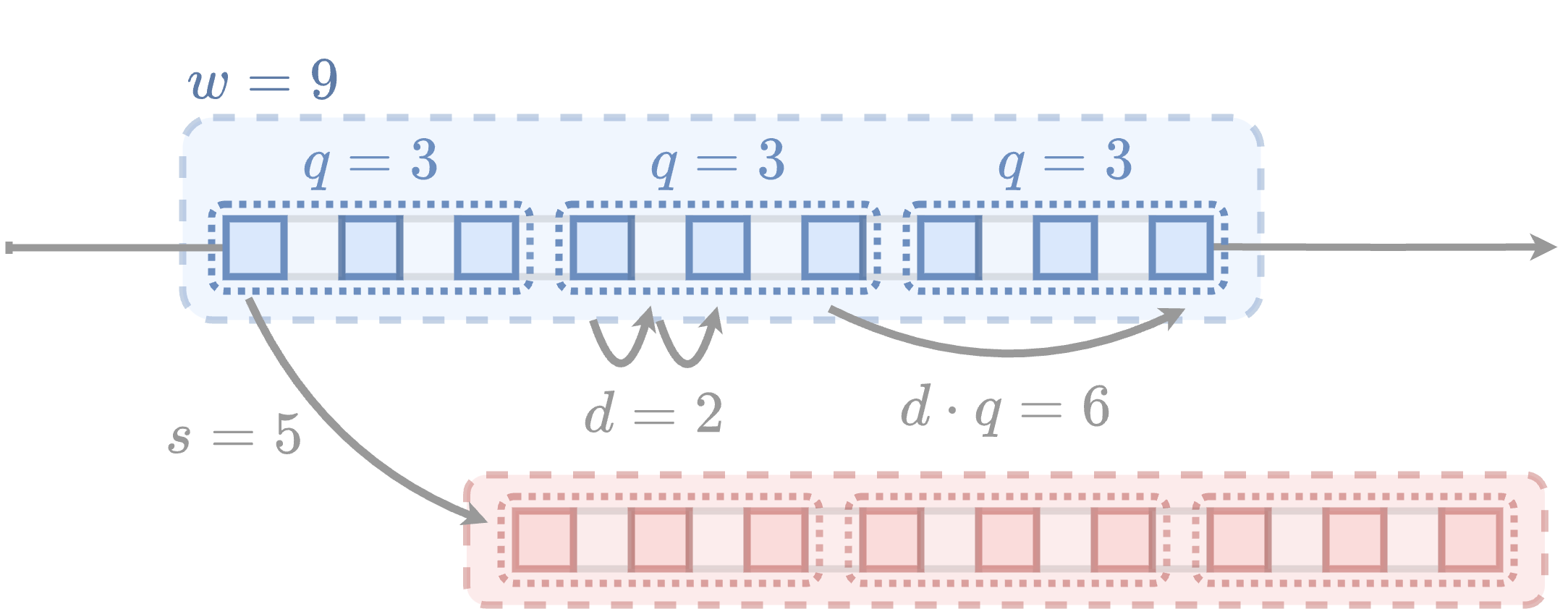}
    \caption{Two receptive fields (blue and red) with a length of $\Cwindowsize=9$, divided into segments of length $\Cseglength=3$. Dilation is the number of steps between consecutive observations within the receptive field. $\Cdilation\Ccdot\Cseglength$ is the segment hop, i.e., the number of steps between consecutive segments. The parameter $\Cstride=5$ is the stride, i.e., the distance between consecutive receptive fields.}
    \label{fig:example-window}
\end{figure}

\subsection{wSAX}
\label{sec:wsax}

Although window-wise PAA and SAX are used in many works ~\cite{baydogan2013bag,senin2013sax,le2019interpretable,nguyen2022fast}, to the best of our knowledge, the formulation of~\cite{lin2007experiencing}, reported in \Cref{eq:paa}, was never formally extended for such a case. 
Thus, as a first step, we formalize window-wise PAA (wPAA), also incorporating stride and dilation operators. 
The output of wPAA can be represented as a matrix \( M \in \mathbb{R}^{\Cnwindows \times \Cwordlength} \), where the rows correspond to the receptive fields, and the columns to the averaged segments:

\begin{align}
M &=
\left[
\begin{array}{ccc}
\Cmean_{1,1} & \cdots & \Cmean_{1,\Cwordlength} \\
\vdots   & \ddots & \vdots  \\
\Cmean_{\Cnwindows,1} & \cdots & \Cmean_{\Cnwindows,\Cwordlength} \\
\end{array}
\right]\in\mathbb{R}^{\Cnwindows\times\Cwordlength}, \label{eq:wpaa}\\
\text{with } \Cmean_{i,j} &= \frac{1}{\Cseglength} \sum_{k=1}^{ \Cseglength} \Cts_{1 + (i-1) \Ccdot \Cstride + (j-1)\Ccdot \Cdilation\Ccdot\Cseglength + (k-1)\Ccdot\Cdilation}, \label{eq:paadil}
\end{align}

\noindent where $\Cseglength={\Cwindowsize}/{\Cwordlength}$ is the segment size and with $i\in[1, \Cnwindows], \label{eq:indices} \text{ and }j \in [1, \Cwordlength]$. 
In summary, a time series signal is divided into (possibly) overlapping windows, which are then divided into non-overlapping segments. wPAA takes the average of each of these segments.
In \Cref{fig:example-window}, we show an example of a receptive field of length $\Cwindowsize=9$, containing $\Cwordlength=3$ segments of length $\Cseglength=3$. Dilation, $\Cdilation=2$, is the number of steps between consecutive observations in the receptive field. $\Cdilation\Ccdot\Cseglength$ is the segment hop, i.e., the number of steps between consecutive segments. Finally, the stride, $\Cstride=5$, is the window hop, i.e., the number of steps between consecutive receptive fields.

\paragraph{Naive Approach.}
With this formulation, it follows that the complexity of computing $M$, by repeating \Cref{eq:paadil} for each cell, as is currently done in the literature, is $O(\Cnwindows\Ccdot\Cwordlength\Ccdot\Cseglength)=O(\Cnwindows\Ccdot\Cwindowsize)$, given that for each cell of $M$, we have to aggregate $\Cseglength$ elements. 
The highest number of receptive fields we can have is when $\Cdilation=1$ and $\Cstride=1$~\cite{dumoulin2016guide}. 
Therefore, the complexity simplifies to $O((\Cnpoints-\Cwindowsize+1)\Ccdot\Cwindowsize$). 
This is mostly fine while $\Cwindowsize \ll \Cnpoints$, but, in the worst case, when $\Cwindowsize=\nicefrac{(\Cnpoints+1)}{2}$, the complexity is $O(\Cnpoints^2)$, i.e., quadratic in the number of points of the time series signal\footnote{A more detailed proof is reported in \Cref{sec:complapp}.}. This reduces the possibility of using bigger window sizes, as the time series signal gets longer. 

In the following, we propose a solution for this long-standing problem regarding SAX-based methods and show that it is possible to compute wPAA efficiently, independently from the window size, with a complexity that depends only on the word length. This opens the possibility of using longer window sizes, which can capture long-range patterns, thus increasing the expressivity of words in the Bag-Of-Receptive-Fields.

\paragraph{Speed Up.}
In \Cref{alg:wsax}, we report the pseudo-code of wSAX describing our proposed speed-up. 
Given $\Cseglength={\Cwindowsize}/{\Cwordlength}$ with $\Cwindowsize \equiv 0 \, (\text{mod} \, \Cwordlength)$, i.e., with the window size divisible by the word length, to optimize computation, we can precompute the averages for all the segments in the time series in a moving fashion (line 1), storing the result for later. Formally:

\begin{align}
\Cmeans^{\textit{seg}} &= \textit{ma}(\Csignal, \Cseglength, \Cdilation) =  [\Cmean_1^{\textit{seg}}, \dots, \Cmean_{\Cnpoints+\Cdilation-\Cdilation\Ccdot\Cseglength}^{\textit{seg}}] \label{eq:maseg}\\
\text{with } \Cmean^{\textit{seg}}_i &=   \frac{1}{\Cseglength} \sum_{j=1}^{\Cseglength} \Cts_{i + (j-1)\Ccdot\Cdilation} \label{eq:masegsing},
\end{align}
\noindent where $\Cmean_i$ is the mean of a single segment in $\Csignal$, with $i \in  [1,\Cnpoints+\Cdilation-\Cdilation\Ccdot\Cseglength]$, where $\Cnpoints+\Cdilation-\Cdilation\Ccdot\Cseglength$ is the total number of segments. 
The main advantage of this formalization is that we do not have to independently calculate each $\Cmean_i^{\textit{seg}}$ from \Cref{eq:masegsing}. Instead, we can employ moving algorithms, which compute $\Cmeans^\textit{seg}$ in $O(\Cnpoints)$~\cite{knuth2014art}. 
In this context, a simple moving average algorithm suffices, where:
\begin{align}
    \Cmean^{\textit{seg}}_{i}=\Cmean^{\textit{seg}}_{i-\Cdilation} + \frac{1}{\Cseglength}(\Cts_{i-\Cdilation + \Cdilation\Ccdot\Cseglength} - \Cts_{i-\Cdilation}).
\end{align}
The vector $\Cmeans^{\textit{seg}}$ contains all the information necessary for wPAA.
Instead of computing \Cref{eq:paadil} for each cell of $M$, we can iterate over $i$, $j$ as in \Cref{alg:wsax} (lines 4-7), filling each entry of $M$ as follows: 
\begin{equation}
    \Cmean_{i,j} = \Cmean^{\textit{seg}}_{1 + (i-1) \Ccdot \Cstride +(j-1)\Ccdot\Cdilation\Ccdot\Cseglength} \label{eq:filling}
\end{equation}

\begin{figure}[t]
    \centering
    \includegraphics[width=.99\linewidth, trim = 0mm 30mm 0mm 0mm, clip]{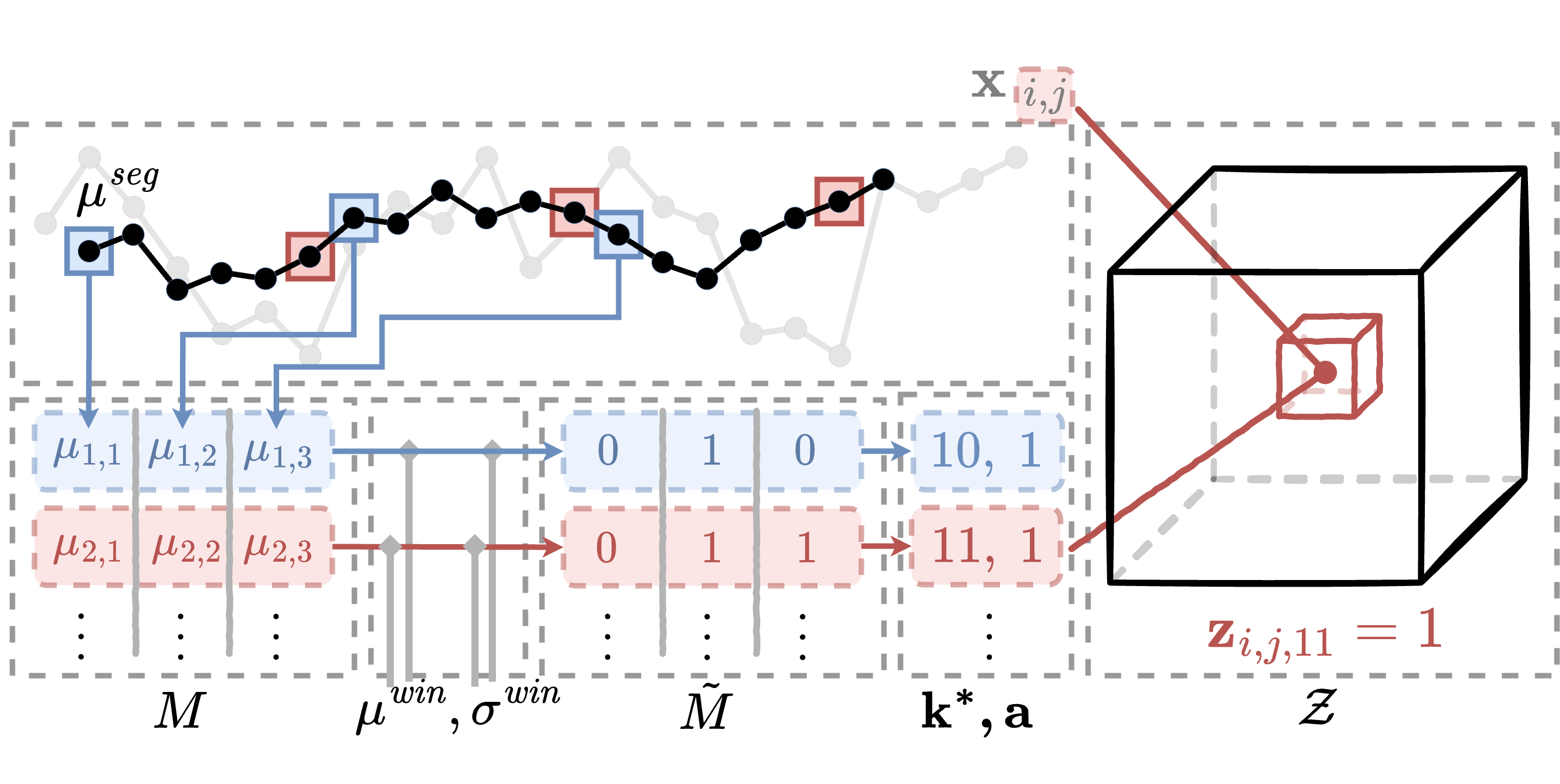}
    \caption{A simplified schema of \Cref{alg:borf} for the two receptive fields of \Cref{fig:example-window}, extracted from signal $\Csignal_{i,j}$. First, the moving average $\Cmeans^{\mathit{seg}}$ is computed. Values in $\Cmeans^{\mathit{seg}}$ are used to fill $M$ as in \Cref{eq:filling}. The two receptive fields are then normalized, binned into the SAX words $[0,1,0]$ and $[0,1,1]$, and hashed into the integers and respective counts $10,1$ and $11,1$. This allows the update of $\Ctabtensor$. E.g., for the red word,  $\Ctab_{i,j,11}=1$.}  
    \label{fig:borf}
\end{figure}

\noindent In \Cref{fig:borf}, we report this step for the two receptive fields depicted in \Cref{fig:example-window}. Performing wPAA basically simplifies to iterating over each cell of $M$, collecting the pre-computed moving averages, based on \Cref{eq:filling}. For example, the blue receptive field is segmented into $[\Cmean_{1,1}, \Cmean_{1,2}, \Cmean_{1,3}] = [\Cmean^{\textit{seg}}_{1}, \Cmean^{\textit{seg}}_{7}, \Cmean^{\textit{seg}}_{13}]$.

\begin{algorithm}[t]
    \footnotesize
    \caption{wSAX}
    \label{alg:wsax}
    \SetAlgoLined
    \KwIn{$\Csignal$ - time series signal, $\Cwindowsize$ - window size, $\Cseglength$ - segment size, $\Cwordlength$ - word length, $\Calphabet$ - alphabet size, $\textbf{b}$ - breakpoints,  $\Cdilation$ - dilation, $\Cstride$ - stride}
    \KwOut{$\Tilde{M}$ - A matrix of SAX words} 
    \BlankLine
    $\Cmeans^{\textit{seg}} = \textit{ma}(\Csignal, \Cseglength, \Cdilation)$; \tcp*[f]{\scriptsize Moving avg for segments}\\
    $\Cmeans^{\textit{win}} = \textit{ma}(\Csignal, \Cwindowsize, \Cdilation)$; \tcp*[f]{\scriptsize Moving avg for windows}\\
    $\Cstds^{\textit{win}} = \textit{mstd}(\Csignal, \Cwindowsize, \Cdilation)$; \tcp*[f]{\scriptsize Moving std for windows}\\
    \textbf{init} $\tilde{M}\in\mathbb{N}^{\Cnwindows\times\Cwordlength}$; \tcp*[f]{\scriptsize Initialize empty matrix}\\
    \For(\tcp*[f]{\scriptsize For each window index}){$1\leq i \leq \Cnwindows$}{ 
        \For(\tcp*[f]{\scriptsize For each segment index}){$1\leq j \leq \Cwordlength$}{ %
            $\Cmean_{i,j} = \Cmean^{\textit{seg}}_{1 + (i-1) \Ccdot \Cstride +(j-1)\Ccdot\Cdilation\Ccdot\Cseglength}$; \tcp*[f]{\scriptsize wPAA} \\
            $\Cmean^*_{i, j} = \textit{zscore}(\Cmean_{i,j}, \Cmean^{\textit{win}}_{i\Ccdot\Cstride},  \Cstd^{\textit{win}}_{i\Ccdot\Cstride})$; \tcp*[f]{\scriptsize Standardize} \\
            $\tilde{\Cmean}_{i, j}=\textit{bin}(\Cmean^*_{i, j}, \Calphabet, \textbf{b})$; \tcp*[f]{\scriptsize Discretize}\\
        }
    }
    \Return{$\tilde{M}$}
\end{algorithm}

\paragraph{Discretization.}
To compute wSAX, each segment average needs to be standardized based on the window to which it belongs. 
Therefore, we compute the mean of each window, $\Cmeans^{\textit{win}}=\textit{ma}(\Csignal, \Cwindowsize, \Cdilation)$ similarly to \Cref{eq:maseg}, and their standard deviation (Alg. \ref{alg:wsax}, lines 2-3):

\begin{align}
\Cstds^{\textit{win}} &= \textit{mstd}(\Csignal, \Cwindowsize, \Cdilation) = [\Cstd_1^{\textit{win}}, \dots, \Cstd_{\Cnwindows}^{\textit{win}}] \\
\text{with } \Cstd_i^{\textit{win}} &=   \sqrt{\frac{1}{\Cwindowsize} \sum_{j=1}^{\Cwindowsize} (\Cts_{i + (j-1)\Ccdot\Cdilation} - \Cmean^{\textit{win}}_i)^2},
\end{align} 
\noindent
which again, can be computed in $O(\Cnpoints)$, using a moving standard deviation algorithm such as~\cite{welford1962note}.
Then, standardization is applied to each element of $M$ (Alg. \ref{alg:wsax}, line 8): 

\begin{equation}
\Cmean^*_{i, j} = \textit{zscore}(\Cmean_{i,j}, \Cmean^{\textit{win}}_{i\Ccdot\Cstride},  \Cstd^{\textit{win}}_{i\Ccdot\Cstride}) = \frac{\Cmean_{i,j} - \Cmean^{\textit{win}}_{i\Ccdot\Cstride}}{\Cstd^{\textit{win}}_{i\Ccdot\Cstride}}.
\end{equation}
\noindent
Finally, discretization is performed as described in \Cref{eq:sax}: $\tilde{\Cmean}_{i, j}=\textit{bin}(\Cmean^*_{i, j}, \Calphabet, \textbf{b})$ (Alg. \ref{alg:wsax}, line 9). The transformed segments, $\tilde{\Cmean}_{i, j}$, are collected in $\tilde{M}\in\mathbb{N}^{\Cnwindows\times\Cwordlength}$, which is the wSAX conversion of signal $\Csignal$. In \Cref{fig:borf}, the blue and red receptive fields are converted into the word vectors $[0,1,0]$ and $[0,1,1]$.

\subsection{Bag-Of-Receptive-Fields}

\label{sec:borf}
Once the SAX words are extracted, we store them in our proposed Bag-Of-Receptive-Fields, as described in \Cref{alg:borf}.

\begin{definition}[Bag-Of-Receptive-Fields]
Given a time series dataset $\Ctsdata \in {\mathbb{R}}^{\Cnmts \times \Cnsignals \times \Cnpoints}$ and a set of $\Cnsubseq$ patterns, a Bag-Of-Receptive-Fields is a tensor $\Ctabtensor\in\mathbb{N}^{\Cnmts\times \Cnsignals \times \Cnsubseq}$, where $\Ctab_{i,j,k}$ is the number of appearances of SAX word $k$ in the signal $j$ of time series $i$.
\end{definition}

\noindent In summary, to build $\Ctabtensor$, we need to repeat wSAX for each time series signal in our dataset (Alg. \ref{alg:borf}, lines 2-3), storing the extracted words and respective counts.
We view this problem as a way of hashing the extracted word vectors into integers. We propose an easy solution, which is to convert each row in $\tilde{M}$ into an integer obtained through concatenating the SAX symbols: 
\begin{equation}
    k_i = \sum_{j=1}^{\Cwordlength} \tilde{\Cmean}_{i,j} \Ccdot \Calphabet^{\Cwordlength - j},
\end{equation}

\noindent where $\Calphabet$ is the alphabe size. Thus, $\textbf{k}=[k_1, \dots, k_{\Cnwindows}]$ (Alg. \ref{alg:borf}, line 5).
Given $\textbf{k}$, through a hashmap, we group-by and count the appearance frequencies of each SAX word, i.e., $\textbf{k}^*, \textbf{a} = \textit{unique}(\textbf{k})$, where $\textbf{k}^*$ contains the unique hashed words, and $\textbf{a}$ contains their count (Alg. \ref{alg:borf}, line 6). Basically, SAX words become integers, used directly as the feature indices in $\Ctabtensor$. This operation is performed for each time series $i$, for each signal $j$. 
Therefore, $\Ctab_{i,j,k} = a$, where $a$ represents how many times the SAX word $k$ was found in $\Csignal_{i, j}$. 
In our example in \Cref{fig:borf}, the discretized red receptive field, $[0,1,1]$, is converted to the integer $11$ and appears only $1$ time in our series signal. 
Its value and count are therefore stored in $\Ctabtensor$, i.e., $\Ctab_{i,j,11} = 1$.

\begin{algorithm}[t]
    \footnotesize
    \caption{\Cmethodname{}}
    \label{alg:borf}
    \SetAlgoLined
    \KwIn{$\Ctsdata$ - time series dataset, $\Cwindowsize$ - window size, $\Cseglength$ - segment size, $\Cwordlength$ - word length, $\Calphabet$ - alphabet size, $\textbf{b}$ - breakpoints, $\Cdilation$ - dilation, $\Cstride$ - stride}
    \KwOut{$\Ctabtensor$ - Bag-Of-Receptive-Fields} 
    \BlankLine
    \textbf{init} $\Ctabtensor\in\mathbb{N}^{\Cnmts\times\Cnsignals\times\Cnsubseq}$; \tcp*[f]{\scriptsize Initialize empty tensor}\\
    \For(\tcp*[f]{\scriptsize For each time series index}){$1\leq i \leq \Cnmts$}{ %
        \For(\tcp*[f]{\scriptsize For each signal index}){$1\leq j \leq \Cnsignals$}{ 
            $\tilde{M} = \text{wSAX}(\Csignal_{i,j}, \Cwindowsize, \Cseglength, \Cwordlength, \Calphabet, \textbf{b}, \Cdilation, \Cstride)$; \tcp*[f]{\scriptsize wSAX} \\
            $\textbf{k} = \textit{hash}(\tilde{M})$; \tcp*[f]{\scriptsize Hash SAX words}\\
            $\textbf{k}^*, \textbf{a} = \textit{unique}(\textbf{k})$; \tcp*[f]{\scriptsize Group-by and count}\\
            \For(\tcp*[f]{\scriptsize For each word,count}){$(k, a) \in \{(\textbf{k}^*, \textbf{a})\}$}{
                $\Ctab_{i, j, k} = a$; \tcp*[f]{\scriptsize Store the word and count}\\
            }

        }
    }
    
    \Return{$\Ctabtensor$}
\end{algorithm}

\subsection{Complexity}
\label{sec:compl}

\paragraph{Time.}
The time complexity of computing the averages for the segments and window means is $O(\Cnpoints)$ and can be calculated using moving average algorithms~\cite{knuth2014art}. The same applies to the moving standard deviation~\cite{welford1962note}. 
Once these quantities are computed, they need to be accessed a number of times equal to the dimension of $M$.
Therefore, the complexity is 
$O(\Cnwindows\Ccdot\Cwordlength)=O((\Cnpoints-\Cwindowsize+1)\Ccdot\Cwordlength)$. 
The worst case would be when $\Cwordlength \in O(\Cwindowsize)$ and $\Cwindowsize \in O(\Cnpoints)$, e.g., $\Cwindowsize = \Cnpoints/2$, and $\Cwordlength = \Cwindowsize/2$, thus leading to an hypothetical overall complexity of $O(\Cnpoints^2)$.
However, it is important to stress that this becomes a pathological parameter choice as the signal length increases. 
In fact, in the literature of SAX-based predictors, $\Cwindowsize$ is increased as $\Cnpoints$ grows, whereas the word length, $\Cwordlength$, is always kept very low to avoid the combinatorial explosion in possible words~\cite{le2019interpretable,nguyen2022fast}. 
For this reason, compared to the naive one, the advantage of our implementation is that, within a reasonable parameter choice, the complexity does not increase with larger window sizes. 
Hence, the final time complexity of repeating this process for each time series signal in our dataset is \(O(\Cnmts \Ccdot \Cnsignals \Ccdot \Cnpoints \Ccdot \Cwordlength)\), while the naive approach is \(O(\Cnmts \Ccdot \Cnsignals \Ccdot \Cnpoints^2)\). 
To further prove this point, in \Cref{sec:experiments}, we show that a word length of $8$ is sufficient to achieve good classification performance, comparable to the state-of-the-art, and empirically benchmark the scalability of the approach.

\paragraph{Space.}
Space complexity is tricky when dealing with SAX patterns, as their number grows fast as the word length and alphabet size increase. The complexity is dominated either by the total number of theoretically possible words, or the number of receptive fields, and is $O(\Cnsignals\Ccdot\min(\Calphabet^\Cwordlength,\Cnmts\Ccdot\Cnpoints))$. 
This seems to require a lot of memory. However, in our case, $\Ctabtensor$ is mostly sparse, given that the majority of entries are $0$, i.e., only a few patterns appear in each time series. 
Choosing a sparse coordinate format representation, where each entry in $\Ctabtensor$ is stored as a 4-valued tuple $(i, j, k, a)$, the total complexity is the number of non-zero elements in the sparse representation, which can be represented as a function of the dataset size ($\Cnmts\Ccdot\Cnsignals\Ccdot\Cnpoints$), but also depend on the intrinsic characteristics of the data. We investigate this relationship in \Cref{sec:experiments}, and provide a heuristic for estimating the number of non-zero elements, given the total dataset size.

\subsection{Time Series Classification}
\label{sec:tsc}
\Cmethodname{} is an unsupervised method, however we benchmark it on Time Series Classification to assess its performance in extracting meaningful features for this task. Given that classification algorithms usually have a 2D input, the Bag-Of-Receptive field can be reshaped naively by concatenating the second and third dimension, leaving the first unchanged, going from a tensor $\Ctabtensor\in\mathbb{N}^{\Cnmts\times \Cnsignals \times \Cnsubseq}$ to a standard matrix dataset $\Ctabdata\in\mathbb{N}^{\Cnmts\times (\Cnsignals \times \Cnsubseq)}$. 
Having previous knowledge of the dataset, an alternative could be to sum the counts of patterns extracted from different groups of time series signals. In general, the flexibility in aggregation could be of great use when domain experts know which signal to aggregate and which signal to keep separate based on their semantics. We explore this possibility in a case study in \Cref{sec:experiments}. 

As highlighted in many recent works~\cite{le2019interpretable,nguyen2022fast,dempster2020rocket}, building multiple representations of the same input data is extremely important for classification performance. In our setting, this is equivalent to building many Bag-Of-Receptive-Fields, each generated with different parameters (e.g., dilation, window size, word length), and then stacking each resulting $\Ctabdata_i$ horizontally: $\Ctabdata=[\Ctabdata_1 | \Ctabdata_2 | \Ctabdata_3| \dots ]$.
Any classification model that takes sparse matrices as input can then use the resulting dataset for training and inference.

\subsection{Explainability}
\label{sec:xai}
The interesting characteristic of a Bag-Of-Receptive-Fields is that, contrary to many state-of-the-art time series classifiers, the meaning of a feature value is very simple to interpret from a human standpoint. 
Given $\Ctab_{i,j,k} = a$, this simply translates into ``The pattern $k$ is contained $a$ times into signal $j$ of time series $i$''. 
Therefore, paired with an interpretable classifier, this feature space can provide easy-to-understand explanations. Interpretability is achieved through both an easily comprehensible feature representation and a simple classifier that is utilized to perform predictions.
For example, using a decision tree, one could extract the decision paths, i.e., rules with conditions based on the number of appearances of discriminative patterns. 
However, in this work, we focus on simple linear models, which are also used in many state-of-the-art competitors as they achieve very good performance. 

\begin{figure}[t]
    \centering
    \subfloat[Cylinder]{%
    \includegraphics[width=.32\linewidth,  trim = 0mm 0mm 0mm 0mm, clip]{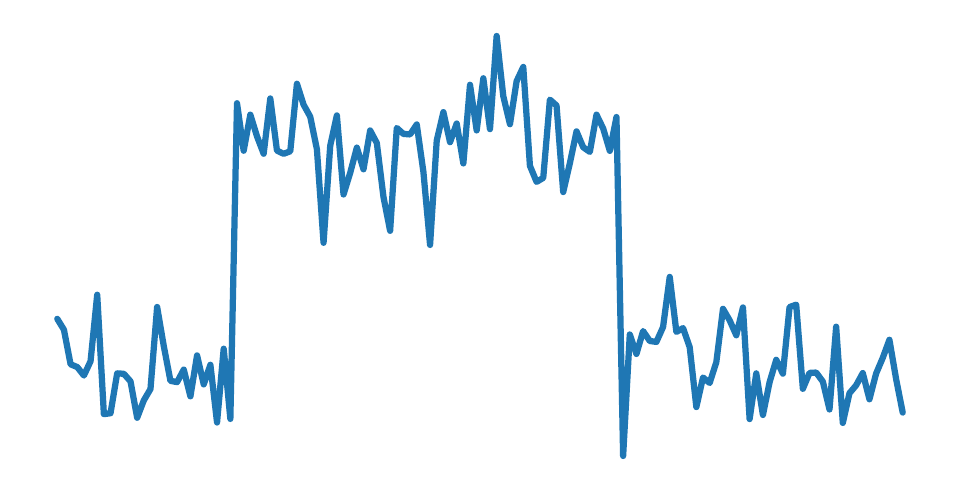}}%
    ~
    \subfloat[Bell]{%
    \includegraphics[width=.32\linewidth,  trim = 0mm 0mm 0mm 0mm, clip]{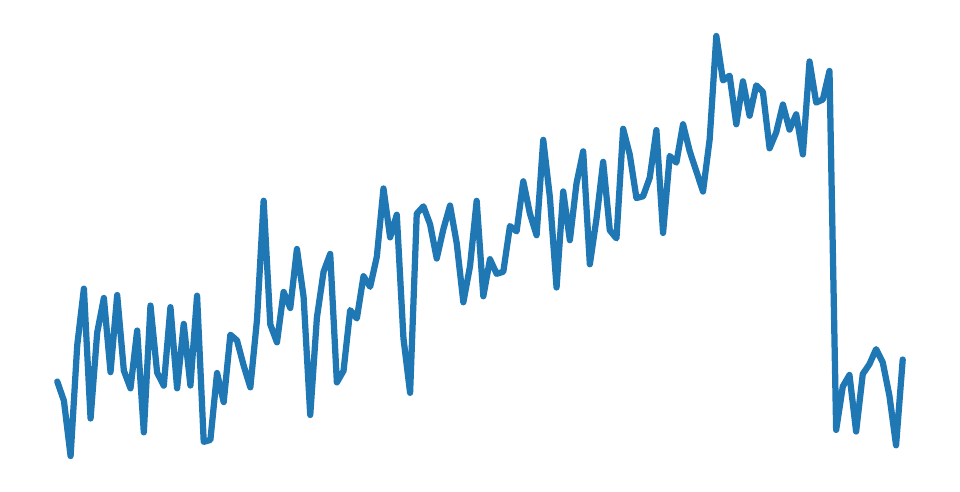}}%
    ~
    \subfloat[Funnel]{%
    \includegraphics[width=.32\linewidth,  trim = 0mm 0mm 0mm 0mm, clip]{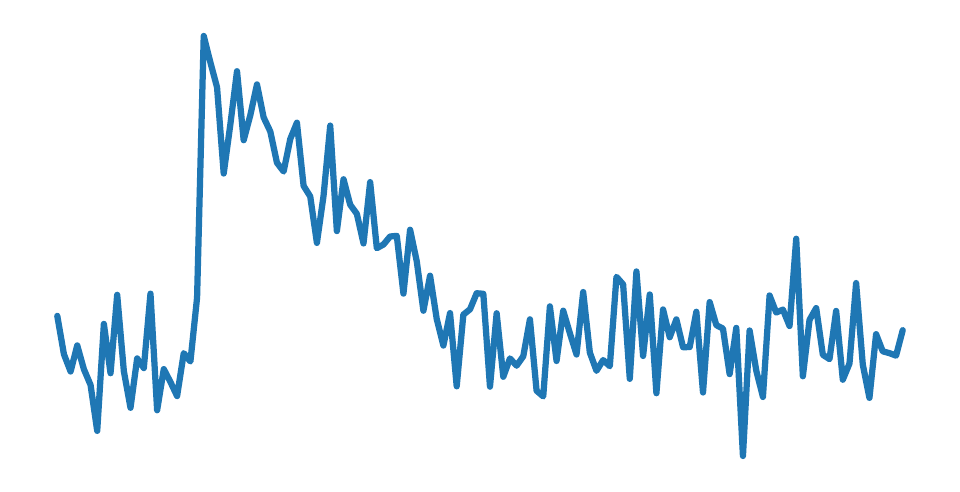}}%
    \caption{Example instances from the three classes of the ``Cylinder-Bell-Funnel'' dataset.}
    \label{fig:cbf}
\end{figure}

In particular, we pair a ridge classifier with SHAP~\cite{lundberg2017unified} to highlight the contribution, $\phi_{i,j,k}$, of each pattern count, $\Ctab_{i,j,k} = a$, towards the classification. 
$\phi_{i,j,k}$ is called SHAP value, and if positive indicates a contribution to the positive class, a negative SHAP value contributes toward the negative class (or, in general, all the other classes if the dataset is multiclass), while a SHAP value close to zero indicates that the feature value is irrelevant. 
Given this semantic, in a Bag-Of-Receptive-Field, $\phi_{i,j,k}$ is the contribution of $\Ctab_{i,j,k} = a$ toward a given class $\Clabel$, which translates to ``The fact that pattern $k$ of signal $j$ of time series $i$ is contained $a$ times, pushes the classification toward class $\Clabel$''.
This allows for building both \textit{global} and \textit{local} explanations, which can give several insights into the classifier behavior. 
In the following subsections, we show examples of these kinds of explanations on the classical ``\textit{Cylider-Bell-Funnel}'' (CBF) dataset~\cite{saito1994local}, containing time series having three distinct shapes, as shown in \Cref{fig:cbf}.

\begin{figure}[t]
    \centering
    \includegraphics[width=.99\linewidth,  trim = 0mm 0mm 0mm 0mm, clip]{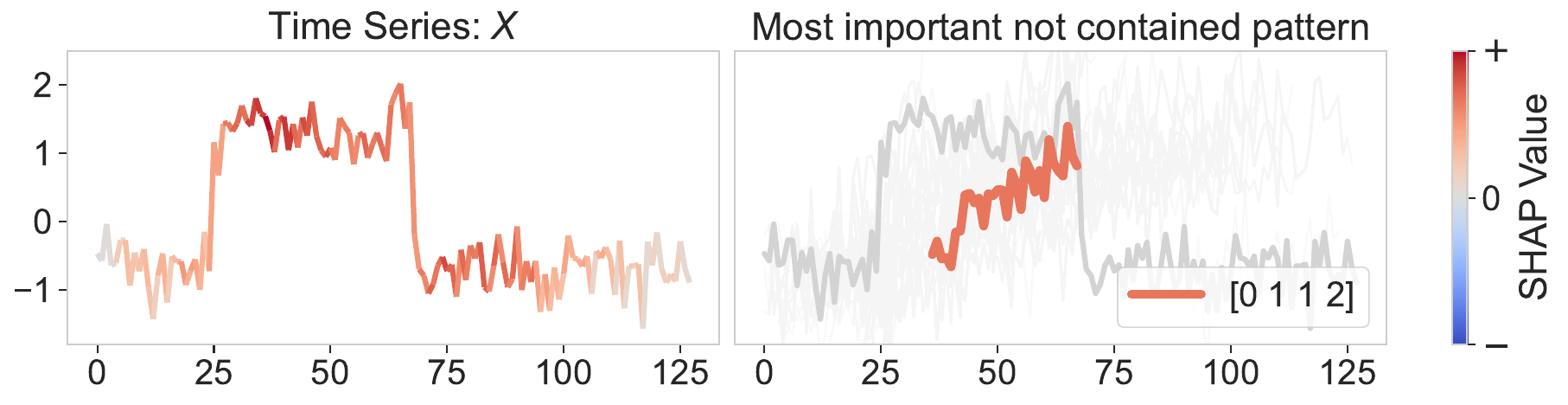}%
    \caption{Local explanation for a \textit{Cylinder} instance of the CBF dataset. On the left, the original time series colored based on the saliency map, highlighting the most relevant observations for the classification. On the right, the medoid of the most important pattern that is not contained in the time series. The grey area represents all the possible alignments of the pattern `0,1,1,2' found in the dataset.}
    \label{fig:cbf_local}
\end{figure}

\paragraph{Local Explanation.}
The local explanation can be built by treating the feature importance of contained and not-contained patterns in different ways. 
Given a time series, $\Cmts\in\mathbb{R}^{\Cnsignals\times\Cnpoints}$, its \Cmethodname{} conversion, $\Ctabdata\in\mathbb{N}^{\Cnsignals\times\Cnsubseq}$, and a feature importance matrix, $\Cfeatimpmatrix\in\mathbb{R}^{\Cnsignals\times\Cnsubseq}$, obtained through some local explainer, such as SHAP, we propose a way to build a saliency map, $\Csaliencymatrix\in\mathbb{R}^{\Cnsignals\times\Cnpoints}$.
A single feature importance, $\Cfeatimp_{j,k}$, for a single SAX word, $\Ctab_{j,k}$, can be mapped back to the original timesteps only if $\Ctab_{j,k}$ is contained in the time series, i.e., $\Ctab_{j,k} > 0$.
If that is the case, we can retrieve the set $T_{j, k}=\{(j,i_1), (j,i_2), \dots\}$, containing all the unique alignment indices\footnote{We take the unique indices, as $\Ctab_{j,k}$ can have overlapping alignments on $\Csignal_j$.} for $\Ctab_{j,k}$ on $\Cmts$, and in particular, on signal, $\Csignal_j$.
The saliency matrix is then defined as:
\begin{align}
    \Csaliencymatrix&=
\begin{bmatrix}
\Csal_{1,1}  & \cdots & \Csal_{1,\Cnpoints} \\
\vdots  & \ddots & \vdots \\
\Csal_{\Cnsignals,1} & \cdots & \Csal_{\Cnsignals,\Cnpoints}
\end{bmatrix}
\in\mathbb{R}^{\Cnsignals\times\Cnpoints},\\
\text{with } \Csal_{j, i} &= \sum_{k=1}^{p} \Cfeatimp_{j, k} \cdot \mathbbm{1}[(j,i) \in T_{j,k}].
\end{align}

\noindent Basically, the importance of each pattern is ``spread'' on the original indices from which it was extracted. For this reason, the saliency map has to be scaled, in order to sum to the original contribution, and is therefore multiplied by a scaling factor:
\begin{align}
    \Csaliencymatrix^* = \Csaliencymatrix\cdot \frac{\sum_{j=1}^\Cnsignals\sum_{k=1}^p{\Cfeatimp_{j,k}\cdot\mathbbm{1}[\Ctab_{j,k}>0]}}{\sum_{j=1}^\Cnsignals\sum_{i=1}^\Cnpoints{\Csal_{j,i}}}
\end{align}
In practice, the sum of the saliency values is scaled to equal the sum of the feature importances of contained patterns. 

An example of these saliency maps can be viewed in \Cref{fig:cbf_local} (left), plotted on a \textit{Cylinder} time series. The more important the observation, the more intense the color will be. In this case, we can easily see that the classifier is ``looking'' mostly at the central part of the time series (from the $30th$ to the $90th$ timestep). These are the most relevant parts of the time series, as the extremes usually contain noise, and thus tend towards a grey color.
On the other hand, when $\Ctab_{j,k}=0$, i.e., when pattern $\Ctab_{j,k}$ is not contained in the time series, it cannot be mapped back to the time series. 
Therefore, we show its general shape, by retrieving the indices $T_{j, k}$, and its respective observations, $X_{T_{j, k}}$, from several time series in the dataset. 
In practice, we use a background dataset of sorts to map SAX words onto the time series to see the real shape that these symbolic words assume in our dataset. 
In \Cref{fig:cbf_local}, we show the alignments of pattern `0,1,1,2' with a light gray color, and the medoid of these alignments is colored based on its SHAP value. This pattern closely resembles the increasing part of a \textit{Bell} instance, and its absence is a strong indicator for the model towards the \textit{Cylinder} class.

In summary, the local explanation comprises a saliency map explaining the contained patterns, and a feature importance vector, explaining the contribution of not-contained patterns.

\begin{figure}[t]
    \centering
    \subfloat[Pattern `0,1,1,2'.]{%
    \label{fig:cbf_global0}%
    \includegraphics[width=.99\linewidth, trim = 0mm 0mm 0mm 0mm, clip]{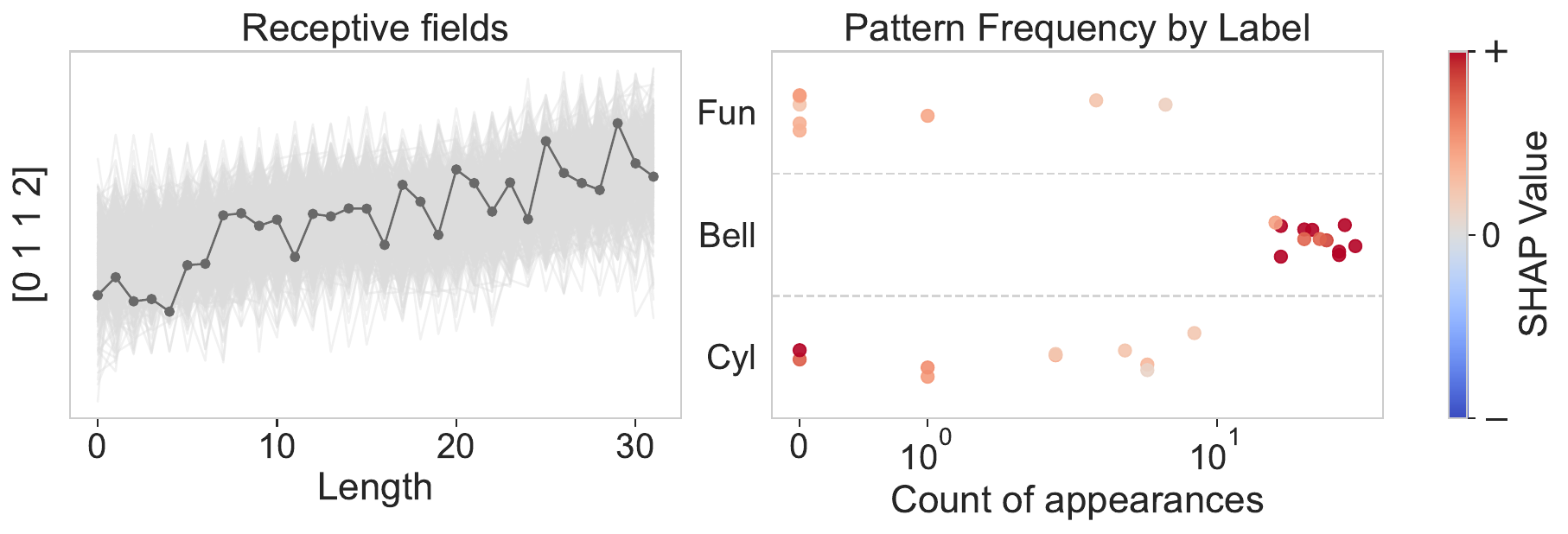}}\\
    \subfloat[Pattern `0,2'.]{%
    \label{fig:cbf_global1}%
    \includegraphics[width=.99\linewidth, trim = 0mm 0mm 0mm 0mm, clip]{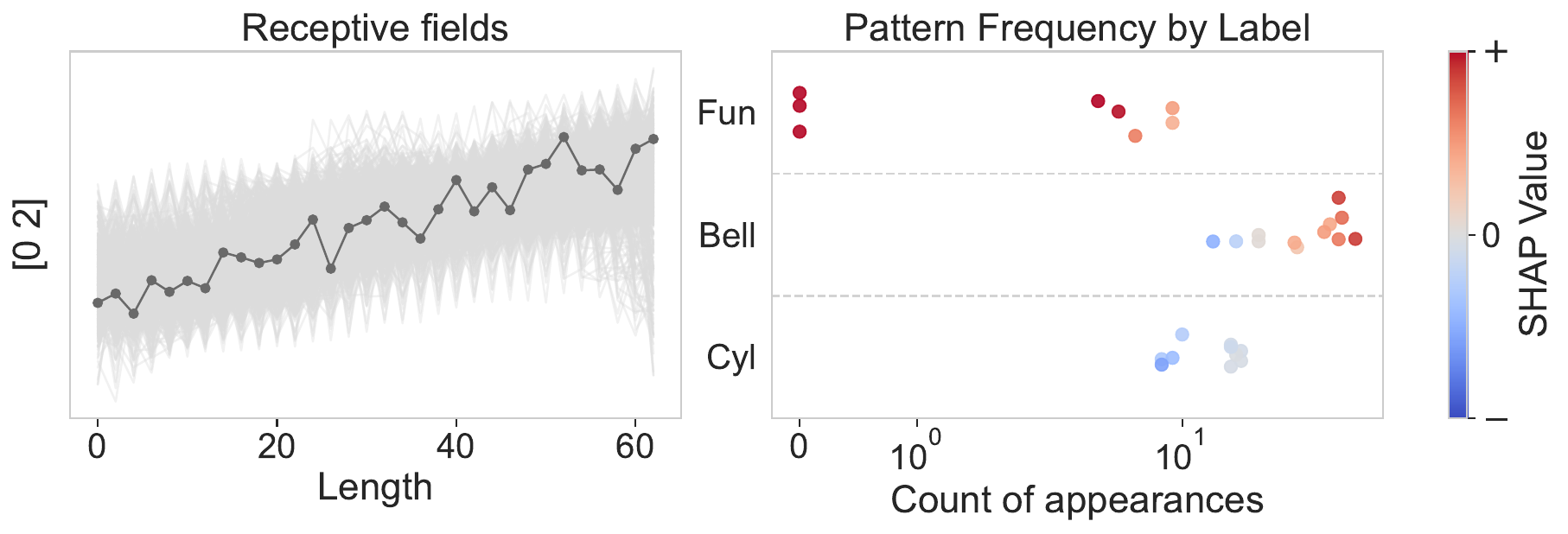}}\\
    \subfloat[Pattern `1,1,1,1'.]{%
    \label{fig:cbf_global2}%
    \includegraphics[width=.99\linewidth, trim = 0mm 0mm 0mm 0mm, clip]{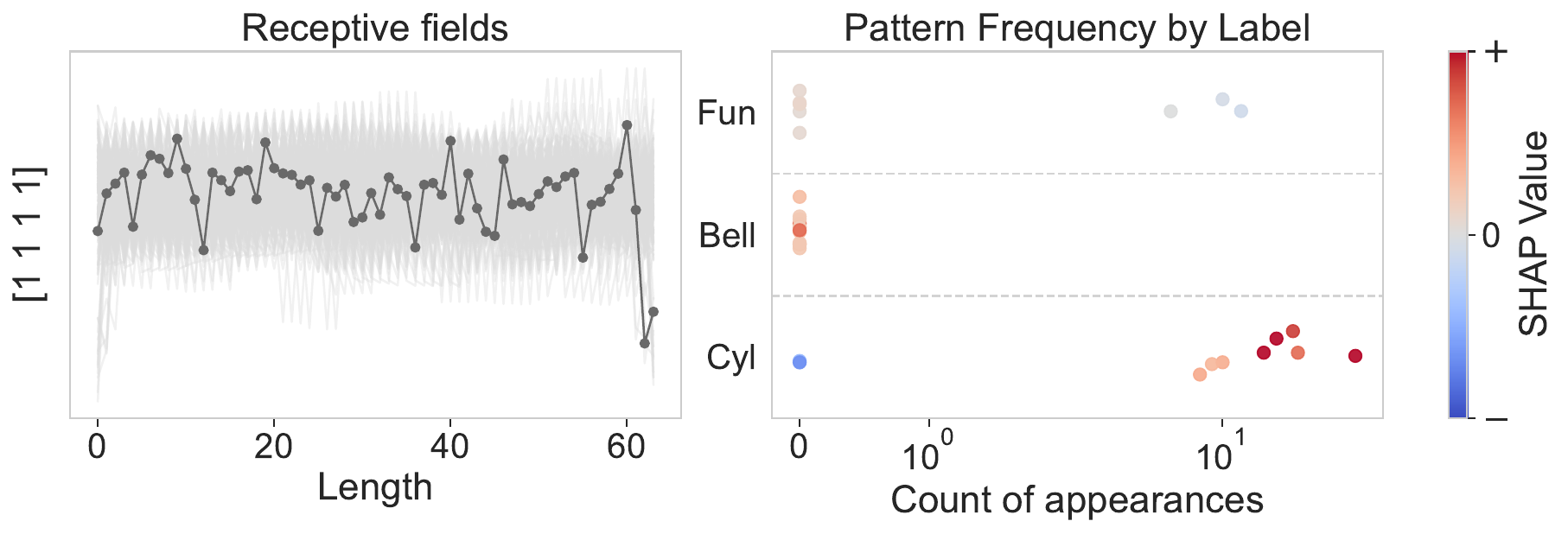}}\\
    \caption{Global explanation for three of the most important patterns for the CBF dataset. On the left, in dark gray, the medoid of all alignments of the pattern in the CBF dataset. On the right, the count of appearances of each pattern in each time series in the dataset, divided by label. The color represents the importance of the pattern count in terms of SHAP values.}
    \label{fig:cbf_global}
\end{figure}

\paragraph{Global Explanation.}
The local explanation highlights the importance of patterns for single time series. However, it does not tell anything about the general relevance of a given pattern. From the local explanation, a natural question could arise, i.e., if the pattern `0,1,1,2' is important in general or only for the given time series. In general, the explanation that arises from the Bag-Of-Receptive-Fields is the importance of frequency counts of the extracted patterns. 
To find the most important patterns in the dataset, we compute the average rank in terms of absolute SHAP values, i.e., we highlight the patterns that are the most important for most time series. 
We summarize this information for CBF in \Cref{fig:cbf_global} for three of the most relevant patterns in this dataset: `0,1,1,2', `0,2', and `1,1,1,1'. On the left, we depict their medoid in dark gray, while in light gray, we depict all the possible mappings of the same word in the given time series dataset.  
This part of the plot can help give an idea of what the classifier considers the most important shapes in our data. In our case, the first two patterns are increasing, resembling a \textit{Bell} instance of our dataset, while the latter is flatter, more similar to a \textit{Cylinder} instance.

In the right scatterplots of \Cref{fig:cbf_global}, each point represents a time series. 
Its x-axis position represents how many times a given pattern is contained in that time series, the y-axis position is the label of that time series, while the color is the SHAP value: red if it contributes toward its own class, gray if it is irrelevant, and blue if it contributes toward other classes. 
With this plot, we can easily see in which kind of time series the pattern is contained or not, and how many times, while also assessing the importance of its presence/absence. 

Let's take, for example, pattern `0,1,1,2' from \Cref{fig:cbf_global0}. This pattern is contained in \textit{Bell} time series many times (more than $10$), as shown by the cluster of red points on the right. This translates to: ``\textit{In general, the fact that pattern `0,1,1,2' is contained many times in the time series pushes the prediction toward Bell.}''. This makes a lot of sense as the pattern resembles the slowly increasing shape of a \textit{Bell}. `0,1,1,2' is sometimes also contained in \textit{Cylinder} and \textit{Funnel} instances, but the importance is lower (more toward the gray color), suggesting that if the pattern is contained only a few times, it is not really discriminative toward the class. Finally, there are a few bright red instances of the class \textit{Cylinder} in which this pattern is contained zero times, which translates to: ``\textit{In general, the fact that pattern `0,1,1,2' is not contained in the time series pushes the prediction toward Cylinder.}''. The instance depicted in the local explanation is one of them.

The blue points in \Cref{fig:cbf_global1} are also very interesting, as they indicate that `0,2' has a somewhat peculiar behavior in some instances. In fact, similarly to `0,1,1,2', the `0,2' pattern is normally contained many times (more than $10$) in \textit{Bell} instances. However, if `0,2' is contained around $10$ times, this actually contributes toward the class \textit{Funnel}. Therefore, the blue point instances indicate that if `0,2' is contained around $10$ times in \textit{Bell} or \textit{Cylinder} instances, this pushes the prediction toward the class \textit{Funnel}, i.e., the model recognized that this pattern frequency is atypical in \textit{Bells} and \textit{Cylinders}. 

Finally, pattern `1,1,1,1' in \Cref{fig:cbf_global2} represents a flat part of the time series, that usually appear around $10$ times in \textit{Cylinder} instances. The blue point represents an atypical \textit{Cylinder} instance, where `1,1,1,1' is not contained given that, in general, the model considers the absence of `1,1,1,1' as a strong indicator toward \textit{Bells} or \textit{Funnels}.

\begin{table}[t]
\caption{List of competitor classifiers, divided by classifier family, with information about feature space and classifier interpretability, and stochasticity. Models that have an interpretable feature space and classifier, and are also deterministic are highlighted in gray.}
\label{tab:competitors}
\begin{tabular}{rl|ccc}
\toprule
& & \multicolumn{2}{c}{\textit{interpretable}} & \\
\multicolumn{1}{l}{\textit{Model}}                &            & \textit{features} & \textit{classifier} & \textit{deterministic} \\
\midrule
\multicolumn{1}{l}{\textit{Dictionary-Based}} &  & & & \\
%\midrule
\rowcolor[gray]{0.9} {\Cmethod{BORF}}       &  (ours)     & \cmark   & \cmark     & \cmark        \\
\rowcolor[gray]{0.9} \Cmethod{BOP}        &    \cite{baydogan2013bag}   & \cmark   & \cmark     & \cmark        \\
\Cmethod{HYDRA}      &      \cite{dempster2023hydra} & \xmark   & \cmark     & \xmark        \\
\rowcolor[gray]{0.9} \Cmethod{MrSEQL}     &     \cite{nguyen2022fast}  & ~~\cmark*   & \cmark     & \cmark        \\
\Cmethod{MrSQM}      &     \cite{le2019interpretable}  & ~~\cmark*   & \cmark     & \xmark        \\
\Cmethod{REDCOM}     &     \cite{bennett2023red} & \xmark   & \cmark     & \xmark        \\
\rowcolor[gray]{0.9} \Cmethod{SAXVSM}     &    \cite{senin2013sax}   & \cmark   & \cmark     & \cmark        \\
\Cmethod{W+MUSE}     &     \cite{schafer2017multivariate}  & \xmark   & \cmark     & \xmark       \\
\midrule
\multicolumn{1}{l}{\textit{Interval-based}} &  & & & \\
%\midrule
\Cmethod{DrCIF}      &       \cite{middlehurst2021hive}  & \xmark   & \cmark     & \xmark        \\
\Cmethod{RSTSF}      &       \cite{cabello2024fast}  & \cmark   & \cmark     & \xmark        \\
\Cmethod{TSF}        &      \cite{deng2013time}   & \cmark   & \cmark     & \xmark        \\
\midrule
\multicolumn{1}{l}{\textit{Shapelet-based}} &  & & & \\
%\midrule
\Cmethod{RDST}       &      \cite{guillaume2022random}  & \cmark   & \cmark     & \xmark        \\
\Cmethod{ST}         &      \cite{lines2012shapelet}  & \cmark   & \cmark     & \xmark        \\
\midrule
\multicolumn{1}{l}{\textit{Kernel-based}} &  & & & \\
%\midrule
\Cmethod{MINIROCKET} &       \cite{dempster2021minirocket}    & \xmark   & \cmark     & \xmark        \\
\Cmethod{ROCKET}     &       \cite{dempster2020rocket}    & \xmark   & \cmark     & \xmark        \\
\midrule
\multicolumn{1}{l}{\textit{Neural Networks}} &  & & & \\
%\midrule
\Cmethod{CNN}        &   & \xmark   & \xmark     & \xmark        \\
\Cmethod{IT}         & \cite{ismail2020inceptiontime}  & \xmark   & \xmark     & \xmark        \\
\Cmethod{MCDCNN}     & \cite{zheng2014time}  & \xmark   & \xmark     & \xmark        \\
\Cmethod{TAPNET}     & \cite{zhang2020tapnet}  & \xmark   & \xmark     & \xmark        \\
\midrule
\multicolumn{1}{l}{\textit{Distance-based}} &  & & & \\
%\midrule
\rowcolor[gray]{0.9} \Cmethod{KNN-DTW}    &         & \cmark   & \cmark     & \cmark        \\
\midrule
\multicolumn{1}{l}{\textit{Hybrid}} &  & & & \\
%\midrule
\Cmethod{MR-HYDRA}   &   \cite{dempster2023hydra}    & \xmark   & \xmark     & \xmark        \\

\bottomrule
\end{tabular}
\begin{tablenotes}
\scriptsize \item[*] * Without SFA features.
\end{tablenotes}
\end{table}

\section{Experiments}
\label{sec:experiments}
In this section, we benchmark \Cmethodname{} in terms of classification performance and runtime against state-of-the-art baselines on the UEA/UCR repositories~\cite{dau2019ucr,bagnall2018uea}. 
We provide a comparison against different families of classifiers, also studying in which kind of datasets \Cmethodname{} perform best. 
Further, we assess scalability and provide an ablation study to understand which parameters affect \Cmethodname{} more. 
Finally, we show the flexibility of \Cmethodname{} for multivariate time series on a case study, and address interpretability with qualitative examples.

\paragraph{Datasets.}
The UCR and UEA TSC repositories comprise 158 datasets, 128 univariate and 30 multivariate. Some datasets contain time series with variable observations, and some contain missing values. We apply missing value imputation by linear interpolation and last value padding. Further, we concatenate all signals in a single axis for algorithms that do not support multivariate time series. For all tests, we adopt the default training and test splits. 

\paragraph{BORF Heuristic.}
Following the current trend in TSC~\cite{bagnall2017great,ruiz2021great,middlehurst2024bakeredux}, instead of fine-tuning \Cmethodname{} to maximize performance for specific applications, we propose a model that achieves good results in a wide range of problems. 
Since several hyperparameters exist, we provide a heuristic that computes a vector of window sizes, dilations, and word lengths, as follows:
$\Cwindowsizes=[2^2, 2^3, \dots, 2^{\lfloor\log_2\Cnpoints\rfloor}]$, $\Cdilations=[2^0, 2^1, \dots, 2^{\lfloor\log_2(\log_2\Cnpoints)\rfloor}]$, $\Cwordlengths=[2, 4, 8]$, $\Calphabet=3$, $\Cstride=1$. Each valid combination of these parameters is used as a configuration of \Cmethodname{}, resulting in a multi-resolution representation of the time series.
As commonly done in the literature~\cite{le2019interpretable}, to avoid blowing up noise with standardization, we set a minimum standard deviation ratio $\theta={\Cstd^{win}}/{\Cstd^\Csignal}$, between the standard deviation of a window, $\Cstd^{win}$, and the standard deviation of full signal, $\Cstd^\Csignal$. 
If $\theta\leq 0.15$, we consider the window constant (all zeros). 
Tuning is performed on the same $42$ \textit{development dataset} used to tune \Cmethod{rocket}~\cite{dempster2020rocket}, and is discussed in the ablation study. 
After the transformation, we apply an inverse hyperbolic sine function element-wise and use an interpretable ridge classifier as a predictive model\footnote{Code is available at: \url{https://github.com/fspinna/borf}}.

\paragraph{Competitors.}
We compare \Cmethodname{} against several state-of-the-art approaches presented in \Cref{sec:related}. A summary of these baselines is reported in \Cref{tab:competitors}.
First, we assess the performance of \Cmethodname{} against SAX-based methods, namely \Cmethod{bop}~\cite{baydogan2013bag}, \Cmethod{sax-vsm}~\cite{senin2013sax}, \Cmethod{mr-sqm}~\cite{le2019interpretable}, and \Cmethod{mr-seql}~\cite{nguyen2022fast}. Subsequently, we evaluate the performance against many different competitors. 
As a baseline, we use KNN with DTW as the distance metric and a Sakoe-Chiba band of $0.1$ (\Cmethod{knn-dtw}).
For dictionary-based approaches, we benchmark WEASEL+MUSE (\Cmethod{W+MUSE})~\cite{schafer2017multivariate}, RedComets (\Cmethod{redcom})~\cite{bennett2023red}, and \Cmethod{hydra}~\cite{dempster2023hydra}. 
For interval-based models, we compare Time Series Forest (\Cmethod{tsf})~\cite{deng2013time}, Random Supervised Time Series Forest (\Cmethod{RSTSF})~\cite{cabello2024fast}, and the Diverse Representation Canonical Interval Forest (\Cmethod{drcif})~\cite{middlehurst2021hive}. For shapelet-based classifiers, we assess the Shapelet Transform (\Cmethod{st})~\cite{lines2012shapelet} and its random dilated version (\Cmethod{rdst})~\cite{guillaume2022random}.
For deep learning-based models, we compare with a standard 1D Convolutional Neural Network (\Cmethod{CNN}), \Cmethod{tapnet}~\cite{zhang2020tapnet}, a multi-channels deep convolutional neural networks (\Cmethod{MCDCNN})\cite{zheng2014time}, and \Cmethod{InceptionTime} (\Cmethod{it})~\cite{ismail2020inceptiontime}. Additionally, we evaluate \Cmethod{rocket}~\cite{dempster2020rocket}, \Cmethod{minirocket}~\cite{dempster2021minirocket}, and the current absolute state-of-the-art in terms of accuracy/speed tradeoff, Multirocket-Hydra (\Cmethod{mr-hydra})~\cite{dempster2023hydra}.

All models are trained with the default library hyperparameters or values proposed in the respective papers\footnote{We used the Python libraries \texttt{pyts}~\cite{fauzi2020pyts} for \Cmethod{bop} and \Cmethod{sax-vsm}, and \texttt{aeon}~\cite{middlehurst2024aeon} and \texttt{sktime}~\cite{loning2019sktime} for all the other methods.}. 
Each model is allowed one day ($1440$ minutes) for training and inference on each dataset and is allocated $32$ cores and $64$ GB of memory\footnote{System: Lenovo SD650 nodes, with Intel Xeon Platinum 8268 CPUs.}. A run is considered failed if it exceeds the maximum time or crashes due to out-of-memory errors, and is assigned the lowest rank.

\begin{figure}[t]%
\centering
\subfloat[Accuracy]{%
\label{fig:cdplot_all_acc}%
\includegraphics[width=\linewidth,  trim = 20mm 15mm 20mm 15mm, clip]{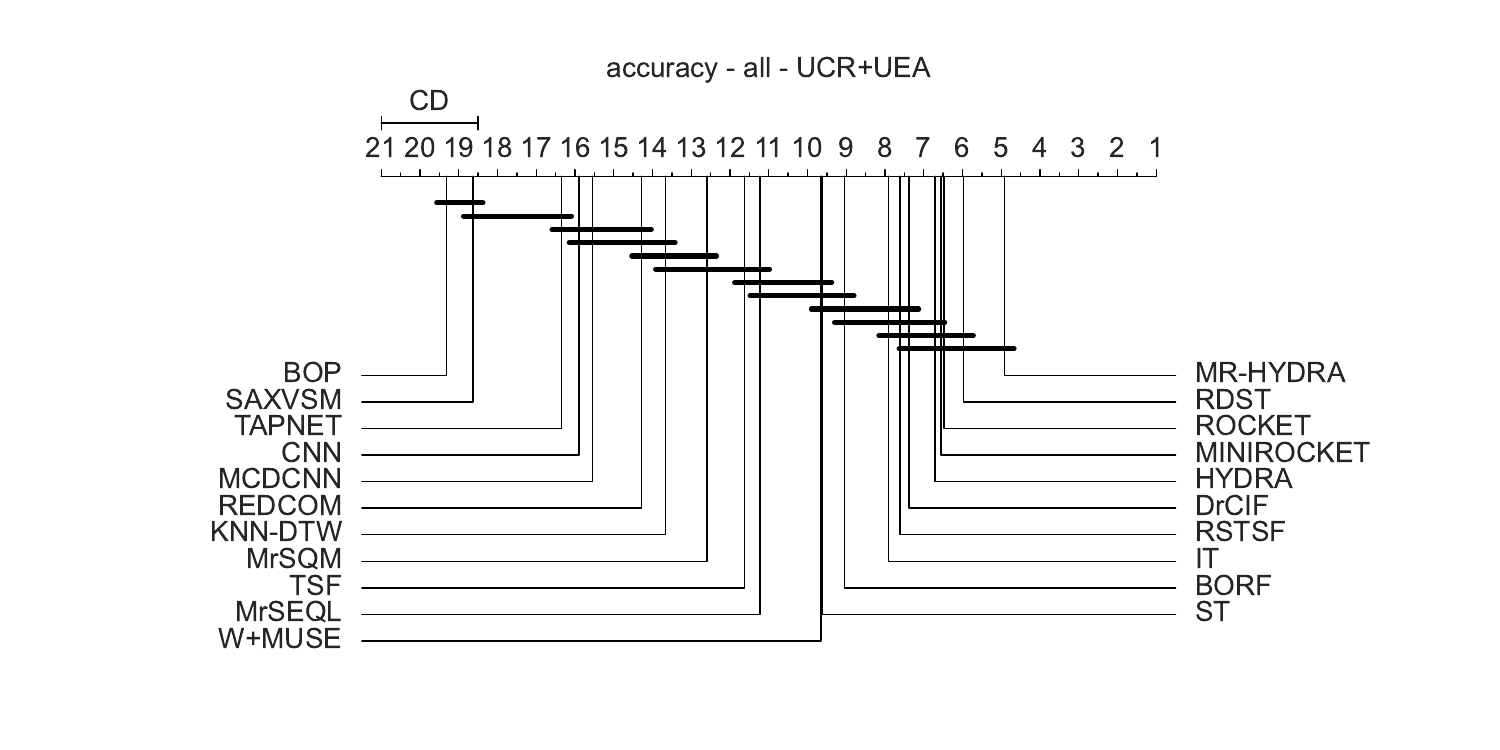}}\\
\subfloat[Runtime]{%
\label{fig:cdplot_all_time}%
\includegraphics[width=\linewidth, trim = 20mm 15mm 20mm 15mm, clip]{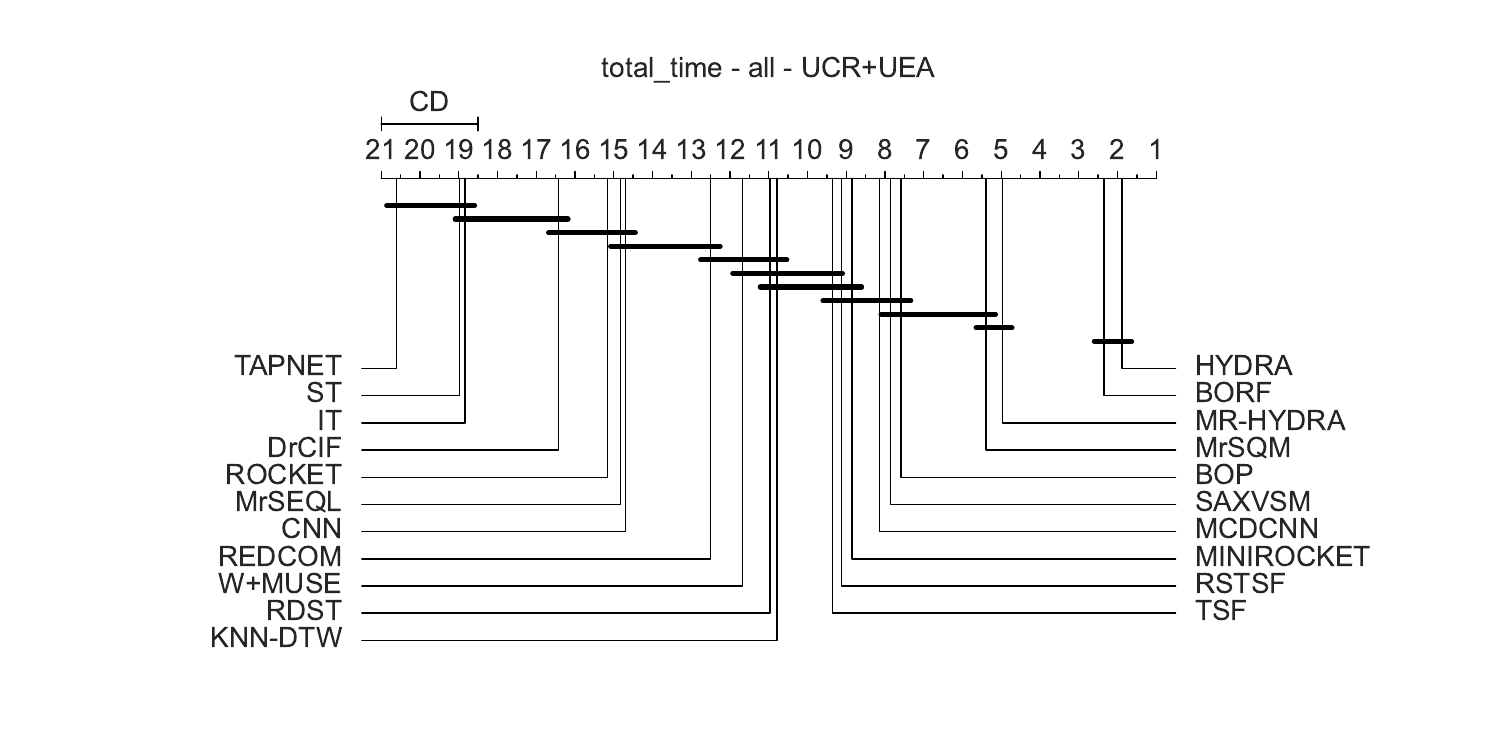}}\\
\caption{CD Plots for all benchmarked methods. Best models to the right.}
\label{fig:cdplot_all}
\end{figure}

\begin{figure}[t]%
\centering
\subfloat[Accuracy (SAX-based)]{%
\label{fig:cdplot_sax_acc}%
\includegraphics[width=.49\linewidth,  trim = 8mm 8mm 8mm 8mm, clip]{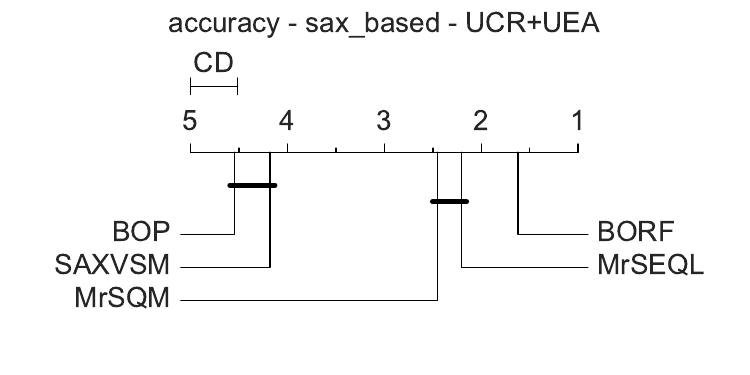}}%
~
\subfloat[Runtime (SAX-based)]{%
\label{fig:cdplot_sax_time}%
\includegraphics[width=.49\linewidth, trim = 8mm 8mm 8mm 8mm, clip]{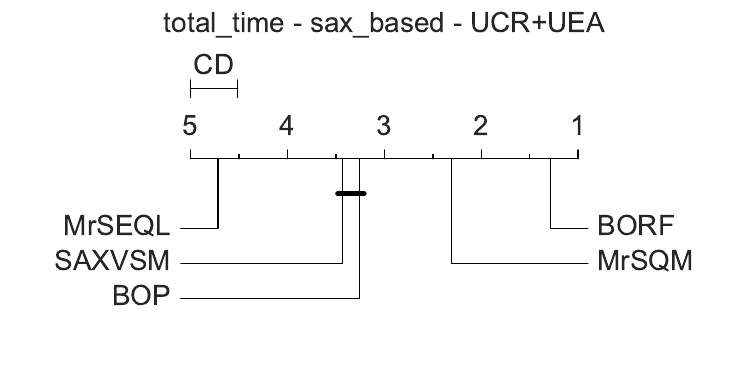}}\\
\subfloat[Accuracy (Dictionary-based)]{%
\label{fig:cdplot_dict_acc}%
\includegraphics[width=.49\linewidth,  trim = 8mm 8mm 5mm 7mm, clip]{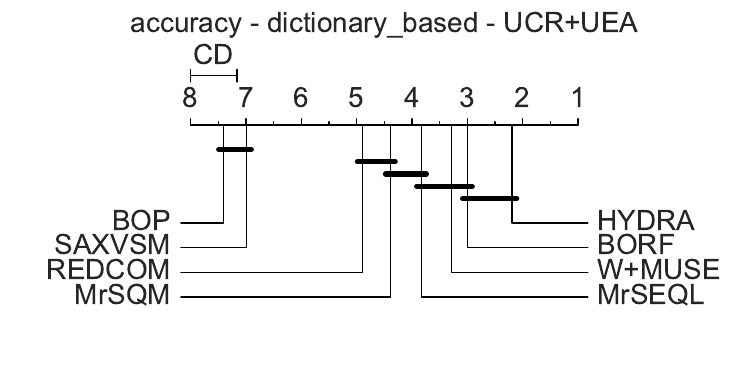}}%
~
\subfloat[Runtime (Dictionary-based)]{%
\label{fig:cdplot_dict_time}%
\includegraphics[width=.49\linewidth, trim = 8mm 8mm 5mm 7mm, clip]{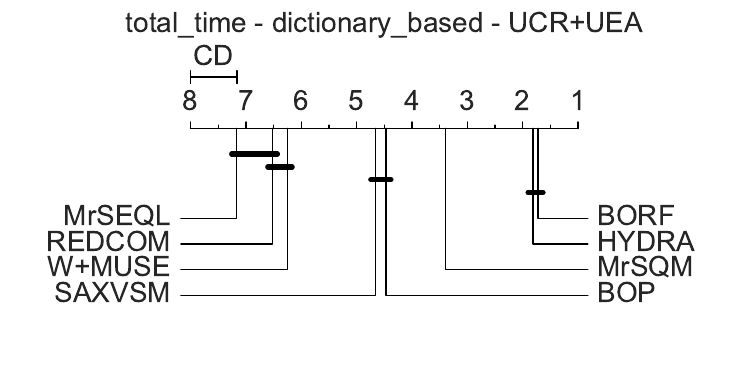}}\\
\subfloat[Accuracy (Interpretable)]{%
\label{fig:cdplot_int_acc}%
\includegraphics[width=.49\linewidth,  trim = 6mm 8mm 6mm 7mm, clip]{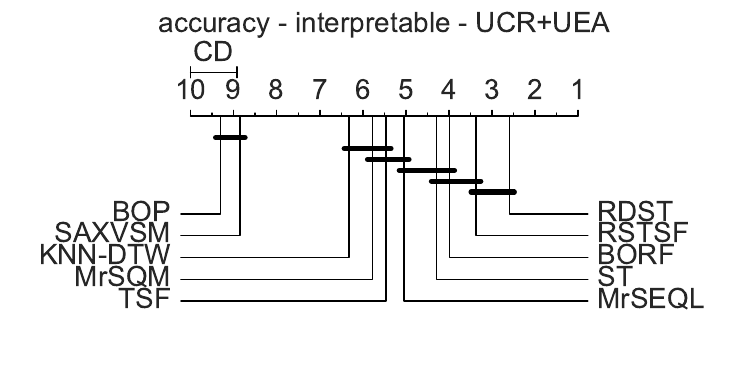}}%
~
\subfloat[Runtime (Interpretable)]{%
\label{fig:cdplot_int_time}%
\includegraphics[width=.49\linewidth, trim = 6mm 8mm 6mm 7mm, clip]{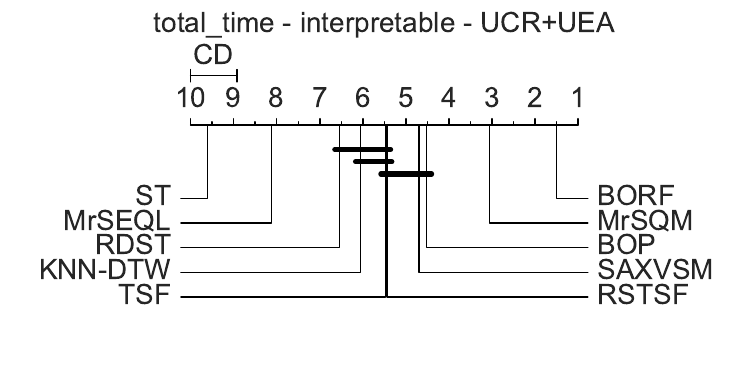}}\\
\caption{CD Plots for SAX-based, dictionary-based, and interpretable methods. Best models to the right.}
\label{fig:cdplot_families}
\end{figure}

\paragraph{Classification.}
In this section, we compare the predictive and runtime performance of \Cmethodname{} against competitor approaches. We report predictive performance in terms of accuracy, as it is the most commonly used metric, and runtime performance as the sum of training and inference runtime (in seconds). 
Performance in terms of F1-measure is reported in \Cref{sec:more_exp}, as it produces results very similar to accuracy. \Cref{sec:more_exp} also presents the full accuracy table on all datasets for the top 13 best-performing baselines.

To assess the statistical significance of these results, in \Cref{fig:cdplot_all}, we first report the Critical Difference (CD) plot comparing the pairwise rankings of all benchmarked methods. Two methods are considered tied if the null hypothesis that their performance is the same cannot be rejected using the Nemenyi test at $\alpha {=} 0.05$. The best models are reported on the right. In terms of accuracy (\Cref{fig:cdplot_all_acc}), \Cmethodname{} has the 9th average rank and is statistically tied to the top-performing models, such as \Cmethod{rocket} and \Cmethod{minirocket}, with the only statistically better methods being \Cmethod{rdst} and \Cmethod{mr-hydra}. Regarding runtime, \Cmethodname{} has the 2nd best rank, statistically tied with \Cmethod{hydra}. \Cmethodname{} and \Cmethod{hydra} are statistically faster than all other approaches, with the closest ones being \Cmethod{mr-hydra} and \Cmethod{mrsqm}. 
This plot shows that our approach achieves good performance that is close to, even if slightly below, the general state-of-the-art in terms of accuracy, and state-of-the-art speed. 
As state-of-the-art accuracy against all methods is not the claim of this work, but rather competitive performance against SAX-based, dictionary-based, and interpretable methods, we also report the CD plots for subsets of models belonging to different classifier families.

\Cref{fig:cdplot_sax_acc,fig:cdplot_sax_time} show a comparison against all SAX-based methods. 
For this subset, \Cmethodname{} is the clear winner, both in terms of accuracy and runtime, surpassing \Cmethod{mrseql} in terms of accuracy, while being even faster than \Cmethod{mrsqm}. 
In \Cref{fig:cdplot_dict_acc,fig:cdplot_dict_time}, we compare against dictionary-based methods. \Cmethod{hydra} is the most accurate model, but it is statistically tied with \Cmethodname{}. \Cmethod{W+MUSE} and \Cmethod{MrSEQL} are close to \Cmethodname{} in terms of accuracy rank, but they are significantly slower. Finally, in \Cref{fig:cdplot_int_acc,fig:cdplot_int_time}, we compare \Cmethodname{} against interpretable approaches, i.e., approaches that are widely regarded as interpretable or discuss interpretability in their respective papers. 
For this subgroup, \Cmethodname{} is still the fastest approach, while achieving a 3rd rank in accuracy, behind \Cmethod{RDST}, and statistically tied to \Cmethod{RSTSF} and \Cmethod{ST}. Note that \Cmethod{RDST} and \Cmethod{RSTSF}, while considered interpretable, are not deterministic. Therefore, while their predictions can be interpreted, they can also change from run to run given the same exact hyperparameters. On the contrary, \Cmethodname{} is interpretable and also stable across runs.
These experiments on subsets of classifier families show that \Cmethodname{} is always first in terms of runtime, while being the most accurate SAX-based approach and among the best dictionary-based and interpretable models.

\begin{table}[t]
    \centering
    \footnotesize
    \caption{Comparison Matrix between \Cmethodname{} and competitor approaches. Models are sorted by median difference in accuracy with \Cmethodname{}. Wins, ties, and losses are to be read ``against \Cmethodname{}''. Best models on top. Grey rows are statistically tied with \Cmethodname{} using the Wilcoxon signed-rank test.}
    \label{tab:mcm_comparison_acc}
\begin{tabular}{r|S[round-precision=2, round-mode=places, table-format=-1.2, detect-all,retain-explicit-plus]S[round-precision=2, round-mode=places, table-format=-1.2, detect-all,retain-explicit-plus]rrrS[round-precision=3, round-mode=places, table-format=<0.3, detect-all, round-minimum=0.001]}
\toprule
& \multicolumn{2}{c}{\textit{$\Delta$accuracy}} & & & & \\
 & {\textit{median}} & {\textit{mean}} & {\textit{win}} & {\textit{tie}} & {\textit{loss}} & {\textit{p-val}} \\
\midrule
\Cmethod{MR-HYDRA} & +0.022192 & +0.037400 & 109 & 22 & 27 & 0.000100 \\
\Cmethod{RDST} & +0.020242 & +0.030600 & 103 & 20 & 35 & 0.000100 \\
\Cmethod{ROCKET} & +0.016769 & +0.022100 & 100 & 17 & 41 & 0.000100 \\
\Cmethod{HYDRA} & +0.011855 & +0.022700 & 95 & 19 & 44 & 0.000100 \\
\Cmethod{MINIROCKET} & +0.010000 & +0.020500 & 96 & 17 & 45 & 0.000100 \\
\Cmethod{DrCIF} & +0.006530 & +0.017600 & 91 & 8 & 59 & 0.0012 \\
\rowcolor[gray]{0.95} \Cmethod{RSTSF} & +0.006480 & -0.000200 & 89 & 12 & 57 & 0.0160 \\
\rowcolor[gray]{0.9} \Cmethod{IT} & +0.005714 & +0.011800 & 86 & 9 & 63 & 0.0664 \\
\midrule
\Cmethod{BORF} & 0.855870 & 0.803461 &  &  &  &  \\
\midrule
\rowcolor[gray]{0.9} \Cmethod{ST} & 0.000000 & -0.014400 & 71 & 12 & 75 & 0.6036 \\
\rowcolor[gray]{0.9} \Cmethod{W+MUSE} & 0.000000 & -0.038900 & 75 & 15 & 68 & 0.1711 \\
\Cmethod{MrSEQL} & -0.012935 & -0.038700 & 45 & 9 & 104 & 0.000100 \\
\Cmethod{TSF} & -0.019281 & -0.037400 & 50 & 7 & 101 & 0.000100 \\
\Cmethod{MrSQM} & -0.030547 & -0.048800 & 34 & 11 & 113 & 0.000100 \\
\Cmethod{KNN-DTW} & -0.055175 & -0.077400 & 32 & 6 & 120 & 0.000100 \\
\Cmethod{REDCOM} & -0.061862 & -0.094100 & 28 & 6 & 124 & 0.000100 \\
\Cmethod{CNN} & -0.100000 & -0.205500 & 22 & 5 & 131 & 0.000100 \\
\Cmethod{MCDCNN} & -0.105270 & -0.194000 & 21 & 3 & 134 & 0.000100 \\
\Cmethod{TAPNET} & -0.150854 & -0.250000 & 14 & 6 & 138 & 0.000100 \\
\Cmethod{SAXVSM} & -0.384722 & -0.393200 & 5 & 1 & 152 & 0.000100 \\
\Cmethod{BOP} & -0.431526 & -0.438800 & 3 & 1 & 154 & 0.000100 \\
\bottomrule
\end{tabular}

\end{table}

\begin{table}[t]
    \centering
    \footnotesize
    \caption{Comparison Matrix between \Cmethodname{} and competitor approaches. Models are sorted by median difference in runtime with \Cmethodname{}. Wins, and losses are to be read ``against \Cmethodname{}''. Best models on top. Grey rows are statistically tied with \Cmethodname{} using the Wilcoxon signed-rank test.}
\label{tab:mcm_comparison_time}
\begin{tabular}{r|S[round-precision=1, round-mode=places, table-format=-4.1, detect-all,retain-explicit-plus]S[round-precision=1, round-mode=places, table-format=-5.1, detect-all,retain-explicit-plus]rrS[round-precision=3, round-mode=places, table-format=<0.3, detect-all, round-minimum=0.001]}
\toprule
& \multicolumn{2}{c}{\textit{$\Delta$runtime (sec)}} & & & \\
 & {\textit{median}} & {\textit{mean}} &{\textit{win}}  & {\textit{loss}} & {\textit{p-val}} \\
\midrule
\Cmethod{BORF} & 1.305673 & 298.748806 &  &  &  \\
\midrule
\rowcolor[gray]{0.9} \Cmethod{HYDRA} & +0.240600 & -294.475100 & 55 & 103 & 0.4150 \\
\Cmethod{MrSQM} & +1.445971 & +1717.956800 & 16 & 142 & 0.000100 \\
\Cmethod{MR-HYDRA} & +1.842159 & -286.187900 & 25 & 133 & 0.000100 \\
\Cmethod{SAXVSM} & +2.037075 & +8465.654600 & 14 & 144 & 0.000100 \\
\Cmethod{BOP} & +2.080840 & +7917.965700 & 14 & 144 & 0.000100 \\
\Cmethod{MCDCNN} & +5.178714 & +3067.028100 & 8 & 150 & 0.000100 \\
\Cmethod{MINIROCKET} & +6.569353 & -264.884700 & 15 & 143 & 0.000100 \\
\Cmethod{RSTSF} & +7.198198 & +1977.266800 & 3 & 155 & 0.000100 \\
\Cmethod{TSF} & +7.601070 & -247.130400 & 13 & 145 & 0.000100 \\
\Cmethod{KNN-DTW} & +9.648525 & +628.124900 & 11 & 147 & 0.000100 \\
\Cmethod{RDST} & +10.822470 & -157.059500 & 7 & 151 & 0.000100 \\
\Cmethod{W+MUSE} & +12.068445 & +3112.670400 & 2 & 156 & 0.000100 \\
\Cmethod{REDCOM} & +21.037976 & +304.791200 & 12 & 146 & 0.000100 \\
\Cmethod{ROCKET} & +42.215268 & -126.137100 & 4 & 154 & 0.000100 \\
\Cmethod{CNN} & +51.684931 & +1516.333400 & 5 & 153 & 0.000100 \\
\Cmethod{MrSEQL} & +74.944092 & +4104.123300 & 1 & 157 & 0.000100 \\
\Cmethod{DrCIF} & +102.230795 & +69.233300 & 4 & 154 & 0.000100 \\
\Cmethod{IT} & +488.190649 & +1275.131400 & 4 & 154 & 0.000100 \\
\Cmethod{ST} & +571.357490 & +4489.157700 & 0 & 158 & 0.000100 \\
\Cmethod{TAPNET} & +3292.977048 & +17514.254500 & 0 & 158 & 0.000100 \\
\bottomrule
\end{tabular}

\end{table}

While the CD-plot is widely used for performance comparison, it was criticized in~\cite{ismail2023approach} for being prone to manipulation, as the average rank of a model depends on the performance of other comparators. 
For this reason, in \Cref{tab:mcm_comparison_acc,tab:mcm_comparison_time}, we report the Comparison Matrices~\cite{ismail2023approach} in terms of accuracy and runtime. 
This representation, proposed in \cite{ismail2023approach}, computes the Wilcoxon Signed Rank Test, evaluating the median difference by ranking absolute differences and summing signed ranks. 
A significant p-value ($\leq0.01$ or $\leq0.05$) indicates a statistically significant difference between samples. 
We report both the mean and median difference, as well as the number of wins, ties, and losses against \Cmethod{borf}. 
Models are ranked based on the median performance, with the best models on top. 
Again, in terms of accuracy, \Cmethodname{} has the 9th rank, statistically tied to \Cmethod{rstsf}, \Cmethod{IT}, \Cmethod{ST}, and \Cmethod{w+muse}. 
As for median runtime, \Cmethodname{} is the best model, statistically tied to \Cmethod{hydra}. 
Against any competitor, \Cmethodname{} wins more than 100 times out of the total, showing its superior median runtime performance. 
However, the mean runtime performance indicates that there are some outlier datasets in which \Cmethodname{} does not perform as well. 
For example, \Cmethodname{} has a median difference with \Cmethod{rocket} that is 42.2 seconds faster, but 126.1 seconds slower on average. 
This difference is due to only four losses overall, among which the only ones with significant runtime deltas are on the ``\textit{FaceDetection}'' and the ``\textit{InsectWingbeat}'' datasets. 
Discarding these two datasets would improve the average runtime tenfold, from $298.7$ to only $27.9$ seconds. 
``\textit{FaceDetection}'' and ``\textit{InsectWingbeat}'' are highly multivariate ($\geq100$ signals) and short ($\leq65$ points) datasets. 
We believe this difference in performance on these datasets is due to two reasons. 
First, 1D convolution is extremely well-optimized for multivariate time series datasets, so models like \Cmethod{rocket}, \Cmethod{minirocket}, \Cmethod{hydra}, and \Cmethod{multirocket-hydra} greatly benefit from this. 
Second, \Cmethodname{} is much more efficient on longer datasets, where $\Cwordlength\ll\Cnpoints$, i.e., when the word length is much lower than the number of points of the time series.

\begin{figure}[t]%
\centering
\subfloat[All datasets]{%
\label{fig:acc_vs_time_all}%
\includegraphics[width=\linewidth,  trim = 0mm 0mm 0mm 0mm, clip]{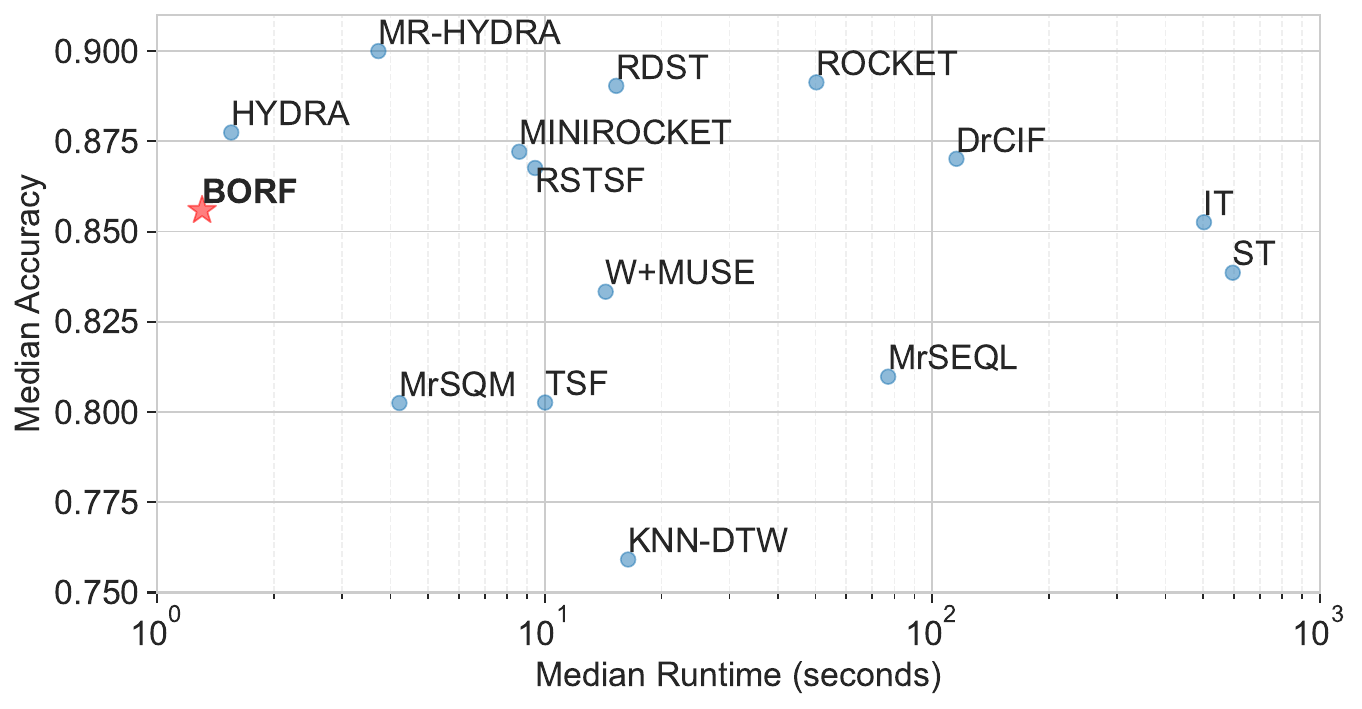}}\\
\subfloat[Long ($m\geq500$), univariate ($c=1$) datasets]{%
\label{fig:acc_vs_time_long}%
\includegraphics[width=\linewidth, trim = 0mm 0mm 0mm 0mm, clip]{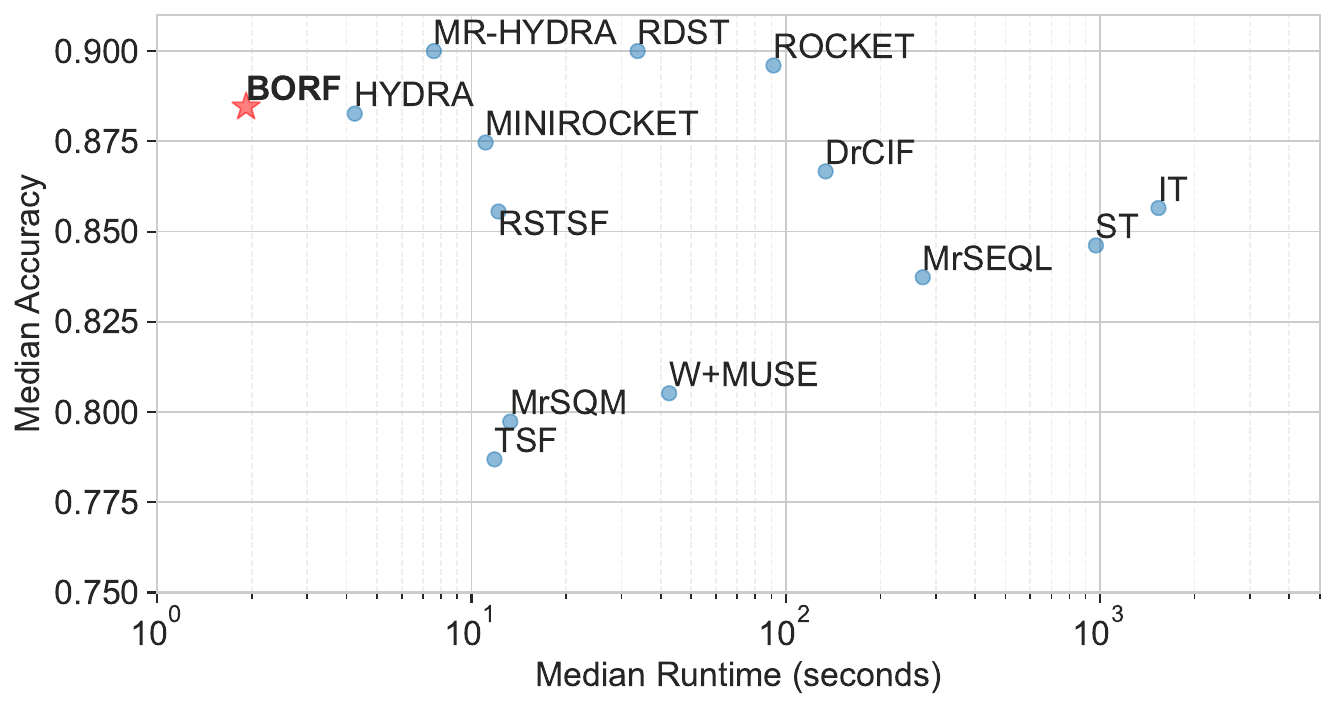}}\\
\caption{Comparison of median accuracy performance against median runtime (seconds). Best models are on the top left.}
\label{fig:acc_vs_time}
\end{figure}

We further prove this point in \Cref{fig:acc_vs_time}. 
In \Cref{fig:acc_vs_time_all}, we show median runtime performance against median accuracy for all datasets. The best models are on the top left. Again, \Cmethodname{} is in the top-10 regarding accuracy, while being the fastest in terms of runtime, with the closest one being \Cmethod{hydra}. 
In \Cref{fig:acc_vs_time_long}, we show the same plot, but on the subset of 45 longest univariate datasets, having more than 500 observations. 
Overall, for this subset, \Cmethodname{} is in the top 5, outperforming even \Cmethod{minirocket} and \Cmethod{hydra} in terms of accuracy and speed. This further proves that \Cmethodname{} tends to be more efficient on longer datasets, which are arguably the most common in real-world scenarios.

\begin{figure}[t]
    \centering
    \includegraphics[width=.99\columnwidth, trim = 0mm 0mm 0mm 0mm, clip]{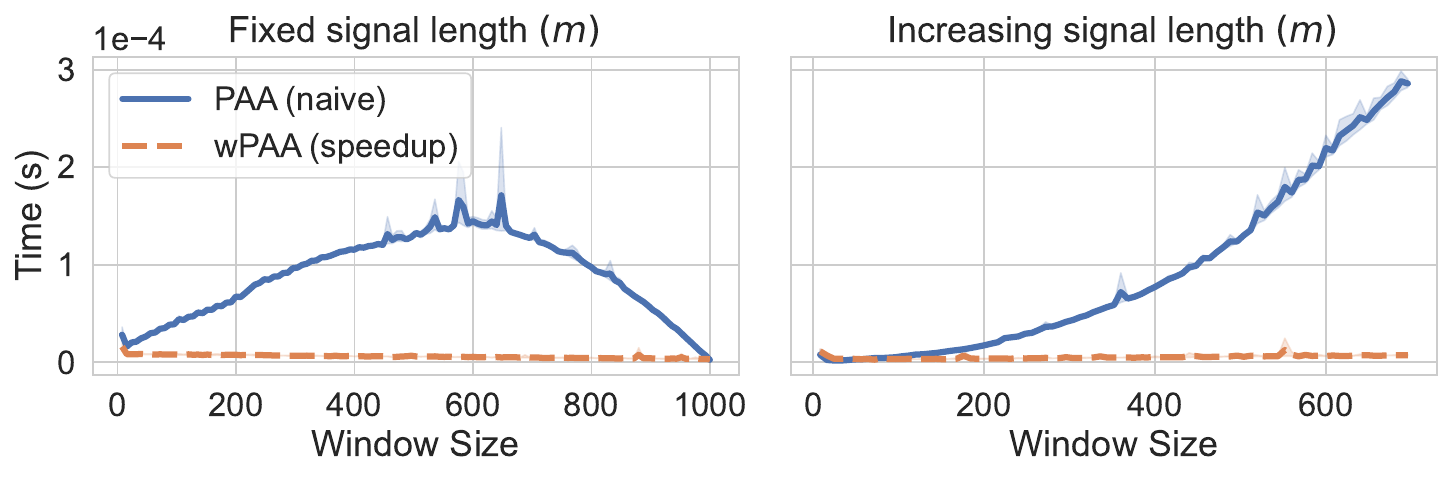}
    \caption{Runtime comparison between naive PAA and our proposal (wPAA). On the left: runtime when changing the window size, and keeping time series length fixed. On the right: runtime when changing the time series length and setting the window size to half the number of points.}
    \label{fig:timecompl}
\end{figure}

\paragraph{Scalability.}
Given the state-of-the-art speed that \Cmethodname{} achieves, we specifically analyze the impact of our proposed algorithmic speedup. That is, we assess the scalability of wPAA against the classical naive PAA (used by methods such as \Cmethod{mrsqm} and \Cmethod{mrseql}), when varying the window size, $\Cwindowsize$.
First, in \Cref{fig:timecompl} (left), we conduct experiments on randomly sampled signals, by fixing their length to $1000$, while varying the window size from $8$ to $1000$ and measuring average runtime on $10$ trials. The word length is kept fixed at $8$. The runtime of naive PAA, as expected given its theoretical complexity of $(\Cnpoints-\Cwindowsize+1)\cdot\Cwindowsize$, grows until $\Cwindowsize\approx\frac{\Cnpoints+1}{2}$, and then decreases. Conversely, wPAA, with the exception of a small overhead at the beginning due to the moving average computation, constantly decreases as the window size increases, and is superior to naive PAA at each point of the curve. On the right, we generate random signals at increasing lengths $\Cnpoints$, setting the window size at half the signal length. This represents the worst-case scenario for PAA, and its quadratic behaviour clearly evidences it. Meanwhile, the runtime of wPAA increases very slowly, as the complexity depends not on the window size but (almost) linearly on the number of observations.

\begin{figure}[t]
    \centering
    \includegraphics[width=0.99\columnwidth, trim = 0mm 0mm 0mm 0mm, clip]{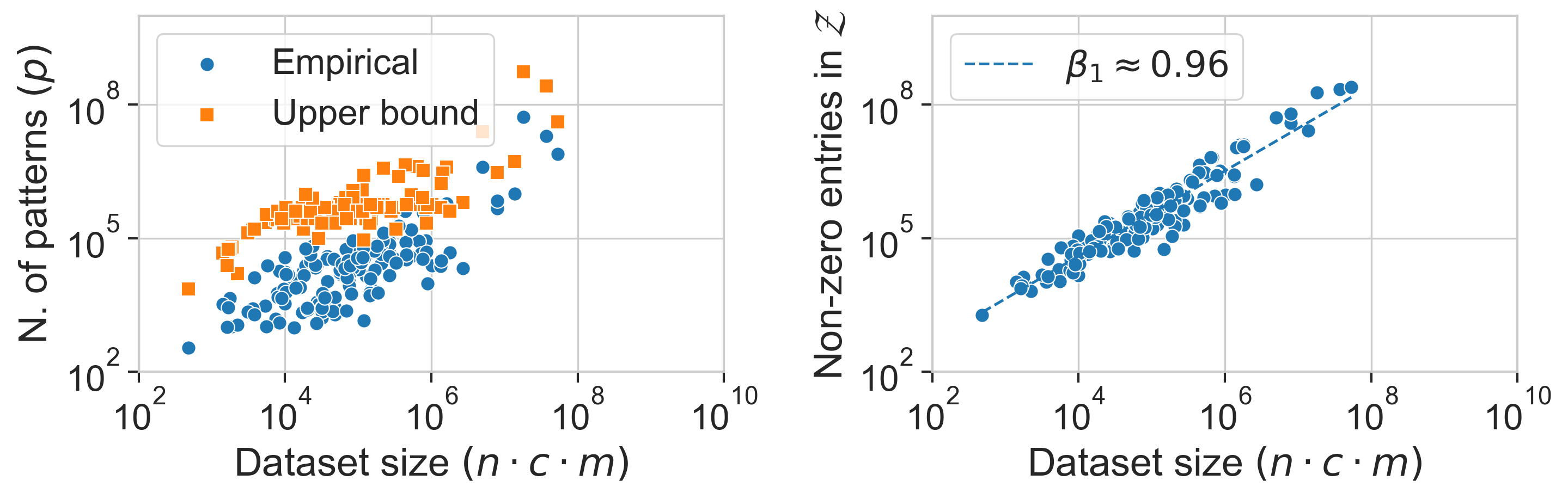}
    \caption{Space complexity. On the left: dataset size against the number of extracted patterns. Each blue point represents the empirical number of extracted features for each dataset, and each orange point is the upper bound on the number of features for that specific dataset. On the right, the dataset size against non-zero elements in $\Ctabtensor$ and fitted log-log regression line.}
    \label{fig:spacecompl}
\end{figure}

Regarding space complexity, in \Cref{fig:spacecompl}, we compare the dataset size ($\Cnmts \cdot \Cnsignals \cdot \Cnpoints$) with the number of extracted patterns (left) and the number of non-zero elements in $\Ctabtensor$ (right) for all datasets in the repositories. Specifically, the plot on the left displays the number of extracted features for each dataset in blue, and the number of theoretically possible features in orange. The upper bound, $O(\Cnsignals \cdot \min(\Calphabet^\Cwordlength, \Cnmts \cdot \Cnpoints))$, is significantly higher than the number of extracted patterns, more than an order of magnitude in most cases. On the right, we illustrate the number of non-zero elements in $\Ctabtensor$, representing the overall empirical space complexity of the algorithm against the dataset size. A linear relationship between the dataset size and the number of non-zero elements becomes evident when visualizing this scatterplot on a log-log scale. To better assess this relationship, we perform a linear regression analysis. The result, $\textit{nnz} \approx 5.74 \cdot (\Cnmts \cdot \Cnsignals \cdot \Cnpoints)^{0.96}$, highlights a nearly proportional relationship, suggesting that the non-zero elements scale almost linearly with the dataset size.

\begin{table}[t]
\centering
\footnotesize
\caption{
Performance delta in median Accuracy (higher is better) and median runtime (lower is better) for various alternative hyperparameter configurations with respect to the \Cmethodname{} baseline heuristic presented in \Cref{sec:experiments}. The two best-performing models for each metric are in bold.}
    \label{tab:ablation}
\setlength{\tabcolsep}{1.2mm}
\begin{tabular}{l|lllll|S[round-precision=3, round-mode=places, table-format=-1.3, detect-all,retain-explicit-plus]S[round-precision=1, round-mode=places, table-format=-2.1, detect-all, retain-explicit-plus]}
\toprule
 & {$\theta$} & {$d$} & {\textit{ash}} & {$l$} & {$w$} & {Acc} & {Time (s)} \\
\midrule 
baseline & \cmark & \cmark & \cmark & 2,4,8 & all & \bfseries 0.820175 & 17.954614 \\
\midrule 
w/o $\theta$ & \xmark & \cmark & \cmark & 2,4,8 & all & -0.012570 & +7.954003 \\
w/o ash & \cmark & \cmark & \xmark & 2,4,8 & all & -0.019762 & -0.053850 \\
w/o dilations & \cmark & \xmark & \cmark & 2,4,8 & all & -0.009656 & -10.144182 \\
$\max{l}=2$ & \cmark & \cmark & \cmark & 2 & all & -0.185247 &  \bfseries -17.196495 \\
$\max{l}=4$ & \cmark & \cmark & \cmark & 2,4 & all & -0.050413 & \bfseries -14.253021 \\
$\max{l}=16$ & \cmark & \cmark & \cmark & 2,4,8,16 & all & \bfseries +0.002720 & +27.502504 \\
$\max{w}=8$ & \cmark & \cmark & \cmark & 2,4,8 & 4,8 & -0.043034 & -11.914014 \\
$\max{w}=16$ & \cmark & \cmark & \cmark & 2,4,8 & 4,8,16 & -0.014399 & -7.452427 \\
$\max{w}=32$ & \cmark & \cmark & \cmark & 2,4,8 & 4,8,16,32 &  -0.000989 & -2.546247 \\
\bottomrule
\end{tabular}
\end{table}

\paragraph{Ablation Study.}
\label{sec:ablation}

We present here an ablation study on \Cmethodname{} to assess which heuristic parameters affect most accuracy and total runtime. 
Starting from our baseline configuration, we change one parameter at a time. 
Specifically, we benchmark \Cmethodname{} without the parameter $\theta$ (w/o $\theta$), without the inverse hyperbolic sine function (w/o ash), and without performing dilated discretization (w/o dilations). 
We then test by setting different maximum word lengths, ($\max{l}=2$, $\max{l}=4$, $\max{l}=16$). 
Additionally, we test limiting the maximum number of window sizes ($\max{l}=8$, $\max{l}=16$, $\max{l}=32$). 

The results are reported in \Cref{tab:ablation}.
On the top row, we report the baseline's median accuracy and median running time, on the $42$ \textit{development datasets}~\cite{dempster2020rocket}.
In all other rows, we report the performance delta between the alternatives and the baseline. 
Regarding accuracy, almost all alternatives perform worse than the baseline, with the exception of increasing the word length to $16$, which slightly improves accuracy. However, we deemed the runtime increase to be too significant to consider this configuration in the heuristic.
Decreasing the word length significantly impacts performance negatively; accuracy decreases by $0.05$ if we remove $\Cwordlength=8$ and $0.185$ if we also remove $\Cwordlength=4$, but it positively impacts runtime. However, we believe the performance drop is too severe to ignore. 
The parameter $\theta$ is particularly interesting because its removal deteriorates both the accuracy and runtime. This parameter impacts the \textit{unique} function in \Cref{alg:wsax}, which receives fewer unique words to hash, while simultaneously removing noisy subsequences. 
Overall, the proposed baseline appears to be a good compromise between accuracy and runtime performance. 
The inverse hyperbolic sine has limited effect, but is still noticeable, at a very small performance gap. 
The only parameter that could be worth limiting is the window size ($\max{l}=32$). However, the performance drop is still present, with minimal runtime improvement. 
Moreover, the development datasets tend to be short in terms of length, so we postulate we can obtain better generalizability with our proposed heuristic.

\begin{figure}[t]
    \centering
    \includegraphics[width=.80\columnwidth, trim = 0mm 0mm 0mm 0mm, clip]{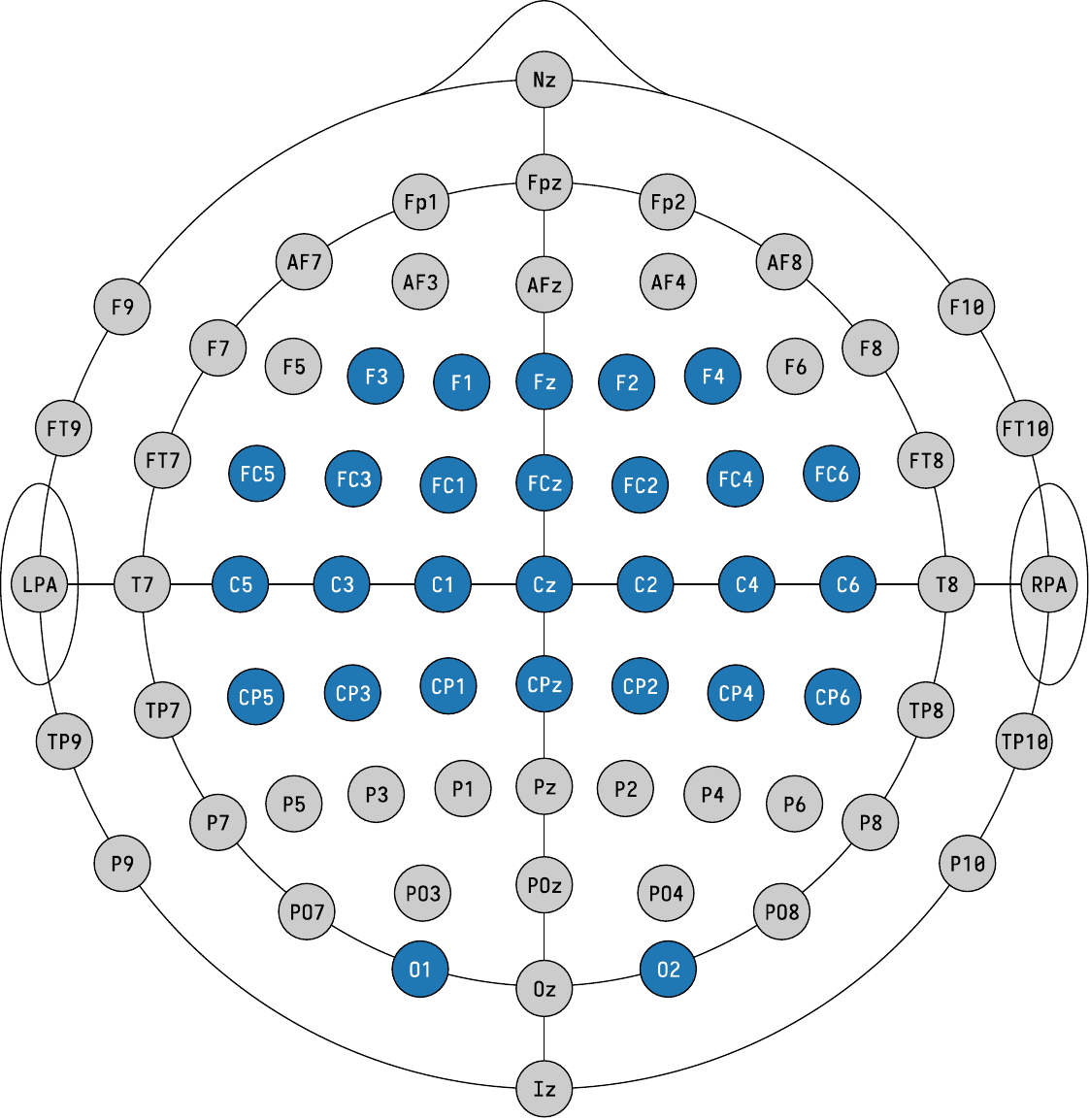}
    \caption{Electrode locations based on the International 10-20 system for encephalography recording. The FingerMovements dataset contains recordings from the blue regions. 
    }
    \label{fig:brain}
\end{figure}

\begin{figure}[t]
    \centering
    \includegraphics[width=.99\columnwidth, trim = 0mm 0mm 0mm 10mm, clip]{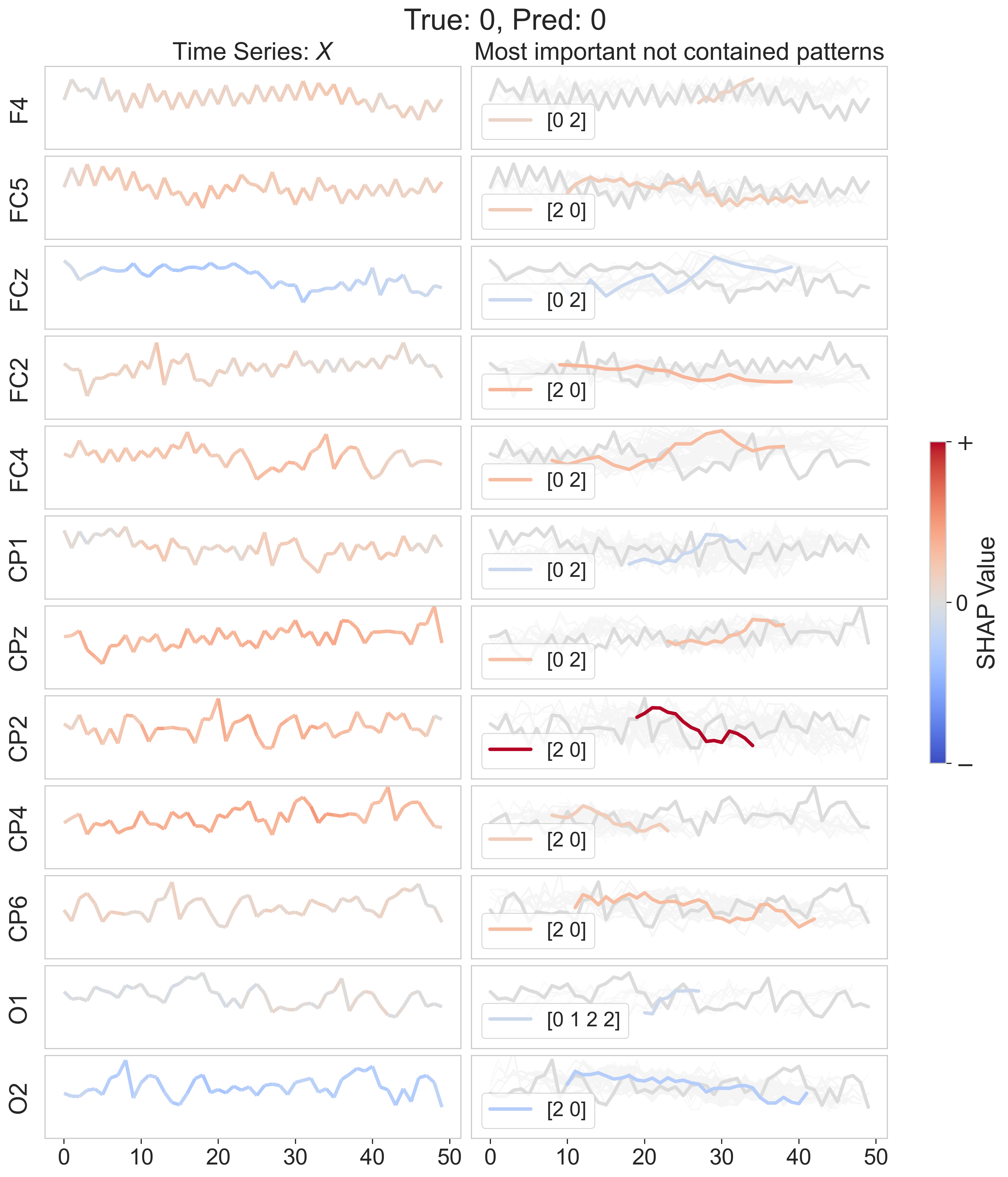}
    \caption{Local explanation for a \textit{Left-Hand} instance of the FingerMovements dataset. On the left, $12$ out of the $28$ signals of the original time series colored based on the saliency map, highlighting the most relevant observations for the classification. On the right, for each signal, the medoid of the most important pattern that is not contained in that signal of the time series.}
    \label{fig:fm_local}
\end{figure}

\begin{figure}[t]
    \centering
    \subfloat[Pattern `2,0' from FCz.]{%
    \label{fig:fm_global0}%
    \includegraphics[width=.99\linewidth, trim = 0mm 0mm 0mm 0mm, clip]{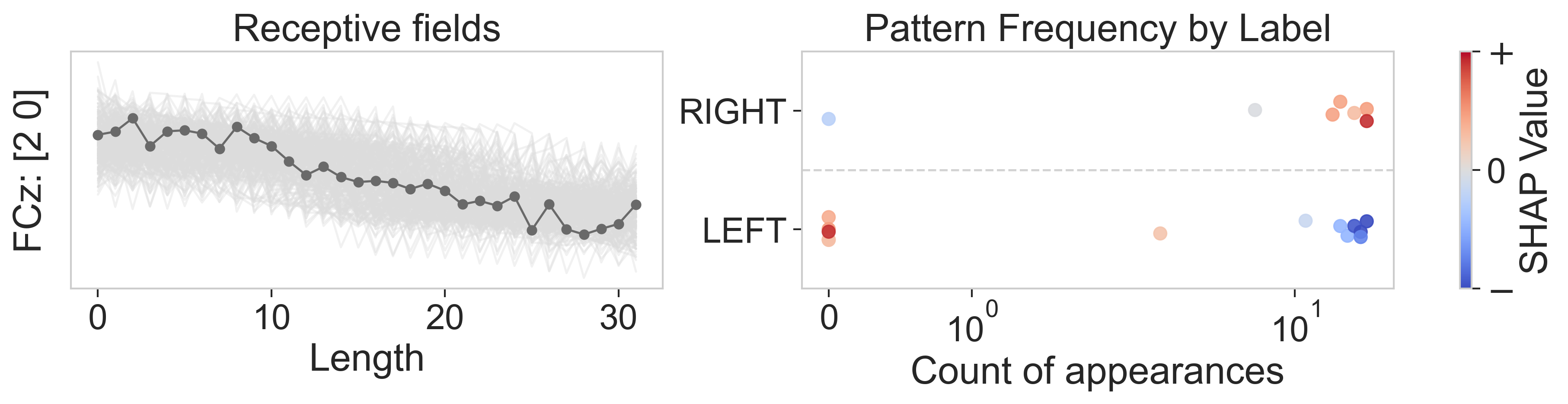}}\\
    \subfloat[Pattern `2,0' from CP2.]{%
    \label{fig:fm_global1}%
    \includegraphics[width=.99\linewidth, trim = 0mm 0mm 0mm 0mm, clip]{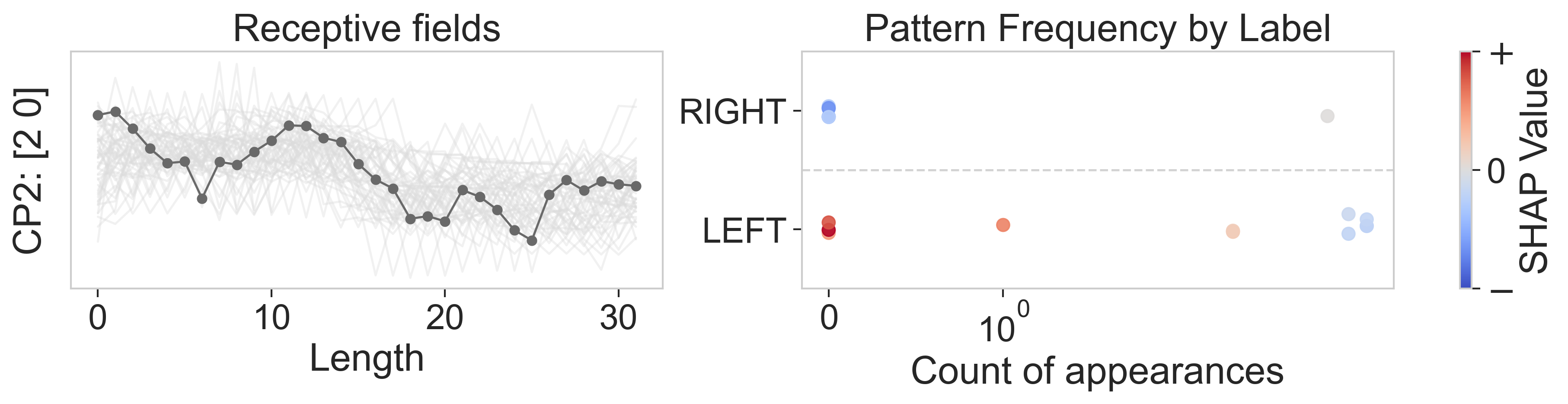}}\\
    \subfloat[Pattern `2,0' from FC3.]{%
    \label{fig:fm_global2}%
    \includegraphics[width=.99\linewidth, trim = 0mm 0mm 0mm 0mm, clip]{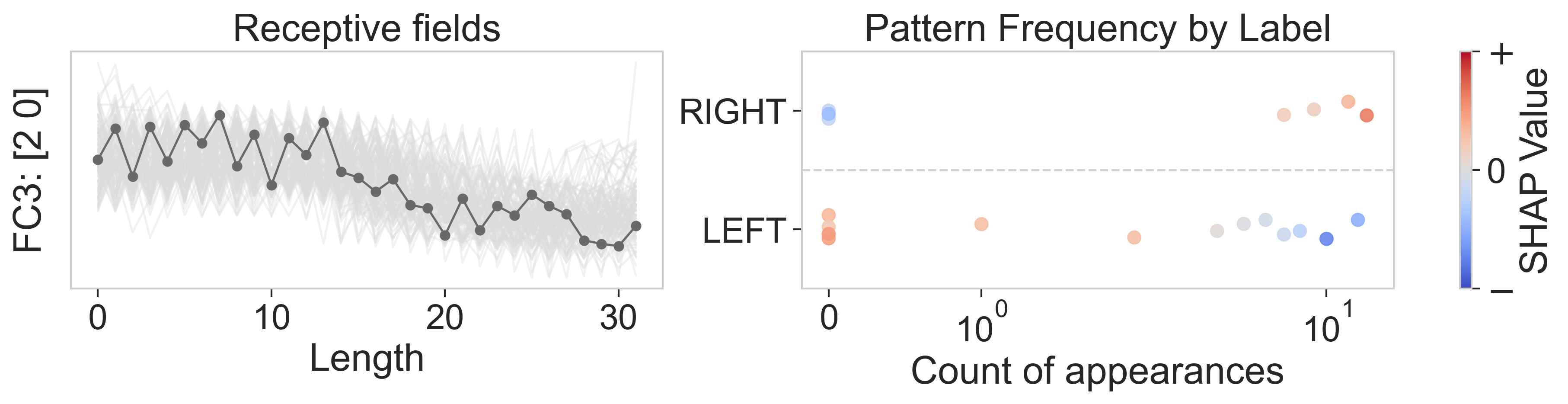}}\\
    \subfloat[Pattern `2,0' from O2.]{%
    \label{fig:fm_global3}%
    \includegraphics[width=.99\linewidth, trim = 0mm 0mm 0mm 0mm, clip]{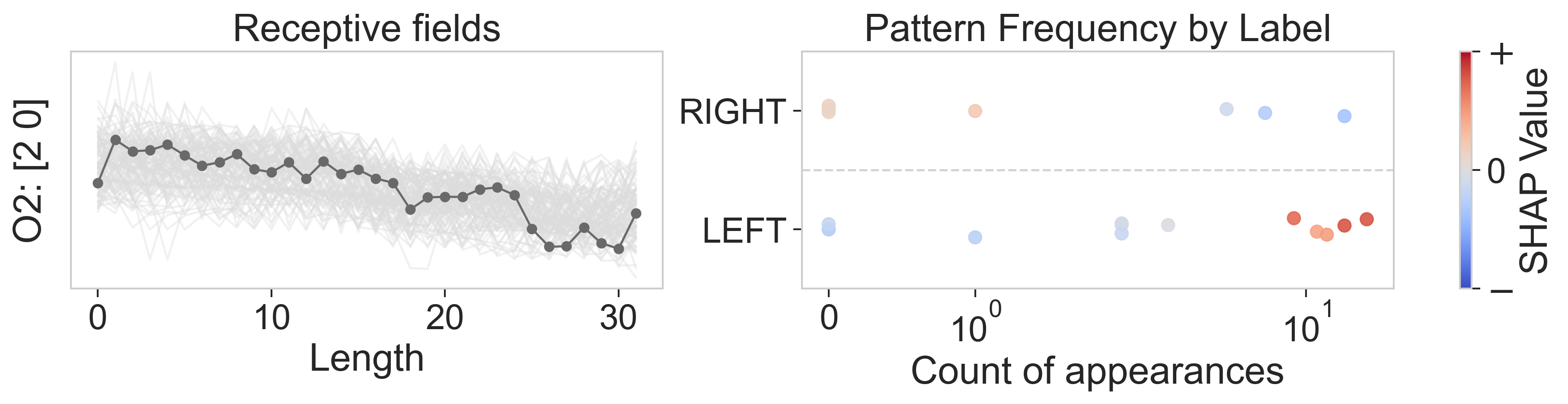}}\\
    \caption{Global explanation for four of the most important `2,0' patterns for the FingerMovements dataset. On the left, in dark gray, the medoid of all alignments of the pattern in the FingerMovements dataset. On the right, the count of appearances of each pattern in each time series in the dataset, divided by label. The color represents the importance of the pattern count in terms of SHAP values.}
    \label{fig:fm_global}
\end{figure}

\paragraph{Case Study: FingerMovements}

We showcase the explainability of the \Cmethodname{} representation on the multivariate \textit{FingerMovements} dataset~\cite{blankertz2001classifying}. This dataset contains EEG sessions of a subject seated with their arms resting on a table, fingers positioned on a keyboard. The task required self-paced typing using the index and little fingers in any chosen order. Each sample comprises recordings from 28 EEG channels, representing signals from distinct brain regions, as depicted in \Cref{fig:brain}. The goal, as defined in the UEA repository, is binary classification: $0$ for imminent \textit{Left-Hand} movements and $1$ for \textit{Right-Hand} movements.

For the local explanation, we choose the \textit{Left-Hand} instance reported in \Cref{fig:fm_local}. On the left side of the picture, we show 12 of the most relevant signals out of the 28, colored based on the importance of each observation toward the classification. For this instance, the saliency map is mostly spread and does not indicate that any part of the time series is particularly important by itself for the classification. However, it is still useful, as it highlights the most relevant signals with a more intense color, in this case, CPz, CP2, and CP4 for positive contribution and O2 for negative contribution. In this case, the right plot, showing the most important not contained patterns, is more interesting, as it shows that the most relevant pattern is `2,0' for half of the depicted signals. This seems to be a very important shape for many signals in the dataset. 

Indeed, by using the global explanation, we can easily see that `2,0' is relevant for many instances of the \textit{FingerMovements} dataset, as shown in \Cref{fig:fm_global}. Analyzing the explanation, it seems, however, that this pattern behaves differently in various signals. For example, if this pattern is not contained in the FCz, CP2, or FC3 signals, it seems to push toward the \textit{Left-Hand} class, while the opposite is true for the `2,0' pattern contained in the O2 signal.

In this sense, a significant advantage of \Cmethodname{} is the flexibility, after transforming the time series, in deciding which patterns have the same semantics, i.e., whether the counts of patterns from certain channels should be summed together or kept separate. In particular, after using \Cmethodname{} for transforming the dataset into $\Ctabtensor\in\mathbb{N}^{\Cnmts\times 28 \times \Cnsubseq}$, we can ask ourselves if patterns for certain brain regions should be summed together or not, i.e., \textit{``Does it make sense to sum the frequency of pattern `2,0' extracted from the occipital lobe with the same pattern `2,0' extracted from the frontal lobe?''}. From the global explanation, it seems like frequencies from some signals have the same semantics and some do not.

Conventionally, pattern count in each signal is maintained separately, i.e., \Cmethodname{} preserves the counts of each pattern for each signal independently. In this setting, the accuracy of \Cmethodname{} is $0.56$, which ranks in the top 5 among the competitor classifiers tested earlier. An alternative method involves summing patterns across all channels, effectively treating pattern frequencies from various parts of the brain as semantically equivalent, obtaining $\Ctabtensor'\in\mathbb{N}^{\Cnmts\times 1 \times\Cnsubseq}$, where $\Ctab'_{i,1,k} = \sum_{j=1}^\Cnsignals \Ctab_{i,j,k}$. This method has improved performance on this dataset, achieving an accuracy of $0.58$, which is the best, tied with \Cmethod{w+muse}.
With \Cmethodname{}, it is possible to supervise pattern aggregation. For instance, as illustrated in \Cref{fig:brain}, the two occipital regions O1 and O2 (at the bottom of the picture) are spatially remote from the 26 recordings of the parietal and frontal lobes, which cluster together in this dataset. Therefore, we can sum pattern counts from these different regions separately, i.e., $\Ctab''_{i,1,k} = \sum_{j=1}^{26} \Ctab_{i,j,k}$ (for the parietal and frontal lobes), and $\Ctab''_{i,2,k} = \sum_{j=27}^{28} \Ctab_{i,j,k}$ (for the occipital regions), obtaining $\Ctabtensor''\in\mathbb{N}^{\Cnmts\times 2 \times \Cnsubseq}$. Using this dataset as our classifier input, we achieve a superior accuracy of $0.61$, which is better than that of any classifiers tested earlier.
This experiment shows that the interpretability, paired with the capability of defining aggregation after the Bag-Of-Receptive-Fields transformation, offers many insights and substantial flexibility, allowing an expert user to use the explanation as a qualitative and quantitative tool to answer research questions.

\section{Conclusion}
\label{sec:conclusion}
In this work, we proposed \Cmethodname{}, a fast, interpretable, and deterministic transformation for univariate and multivariate time series data. We formalized the window-wise Symbolic Aggregate Approximation (wSAX) and augmented it with dilation and stride convolutional operators. Our approach significantly reduces the computational complexity traditionally associated with window-wise SAX-based classifiers and extends the Bag-Of-Patterns framework to a more flexible Bag-Of-Receptive-Fields. 
We have shown that the proposed approach is statistically superior in runtime and predictive performance compared to SAX-based methods while being very competitive among dictionary-based and interpretable classifiers. Moreover, the proposed heuristic is scalable both in time and memory, allowing the accurate analysis of long datasets. We have assessed the interpretability and flexibility of our approach in a case study, showcasing it as a very promising tool for answering research questions.

A limitation of our proposal is that it is still subpar compared to ensemble hybrids such as Multirocket-Hydra, which achieve better predictive accuracy but are uninterpretable. Beating these kinds of models was never the claim of this work, as they use multiple algorithms and complex feature spaces to achieve that level of performance. However, an interesting future research direction would be to build a fast ensemble hybrid that is also interpretable, combining, for example, the feature space of \Cmethodname{} with other interpretable approaches such as shapelets or interval-based methods.
Additionally, like all SAX-based methods, \Cmethodname{} faces the challenge of managing longer word lengths as the feature space expands rapidly. Our ablation study indicates that this does not seem to be a problem for this set of datasets; however, with very long time series, larger word lengths would probably be needed. To address this, we plan to hash the extracted patterns into a pre-fixed size, for example, using Bloom filters, to mitigate this issue.

Finally, given that the \Cmethodname{} transformation is unsupervised, we plan on tackling time series regression, forecasting, and clustering to achieve fast and deterministic explanations in different types of tasks.

\appendices
\crefalias{section}{appendix}

\section{Notation}
\label{sec:notation}
In \Cref{tab:notation}, we summarize the main symbols used in the paper.

\begin{table}[h]
\caption{Summary of notation.}
\footnotesize
\label{tab:notation}
%\color{c1}
\begin{tabular}{ll}
\toprule
\textbf{Time Series Data} & \\ \midrule
$\mathcal{X}, X, \textbf{x}, x$ & time series dataset, instance, signal, entry \\
$\textbf{y}, y$ &  labels vector, value \\
$n$ & number of instances in a dataset \\
$c$ & number of signals in a time series \\
$m$ & number of observations in a signal \\ \midrule
\textbf{Convolution} & \\ \midrule
$w$ & window/receptive field size \\
$d$ & dilation \\
$s$ & stride \\
$v$ & number of windows/receptive fields \\ \midrule
\textbf{Discretization} & \\ \midrule
$l$ & word length \\
$q$ & segment size \\
$\alpha,\textbf{b}$ & alphabet size, Gaussian breakpoints\\
$\mu^{\textit{seg}}_i, \mu^{\textit{win}}_i$ & moving average for segment, window $i$\\
$\sigma^{\textit{win}}_i$ & moving std for window $i$\\
$M, \mu_{i,j}$ & wPAA matrix, entry\\
$M^*, \mu^*_{i,j}$ & standardized wPAA matrix, entry\\
$\tilde{M}, \tilde{\mu}_{i,j}$  & wSAX matrix, entry\\
$\textbf{k}, \textbf{k}^*, \textbf{a}$ & words, unique words, counts\\ \midrule
\textbf{Bag-Of-Receptive-Fields} & \\ \midrule
$\mathcal{Z}, Z, z$ & 3d tensor, 2d dataset, entry \\
$p$ & number of features in \Cmethodname{}\\ \midrule
\textbf{XAI} & \\ \midrule
$\Phi, \phi$ & feature importance matrix, entry \\
$\Psi, \psi$ & saliency matrix, entry \\
\bottomrule
\end{tabular}
\end{table}

\section{Complexity}
\label{sec:complapp}

Here, we show in greater detail that the worst-case complexity of the window-wise naive PAA is quadratic. 
For each window, for each segment, we need to average the values in each segment, therefore \Cref{eq:paadil} is repeated $\Cnwindows \Cwordlength$ times, for a total of $\Cnwindows \Cwordlength \Cseglength$ iterations. Given that, by definition, $\Cseglength \Cwordlength = \Cwindowsize$, it follows that $\Cnwindows \Cwordlength \Cseglength = \Cnwindows \Cwindowsize$. The highest number of receptive fields is obtained when $\Cdilation = 1$ and $\Cstride = 1$, i.e., when dilation and stride are both absent, given that they reduce the number of receptive fields by skipping some observations~\cite{dumoulin2016guide}.
Thus, the total number of iterations simplifies to:
\begin{align*}
\Cnwindows \Cwindowsize &= \left(1 + \left\lfloor \frac{(\Cnpoints - \Cwindowsize - (\Cdilation - 1)(\Cwindowsize - 1))}{\Cstride} \right\rfloor \right) \Cwindowsize \\
&= (\Cnpoints - \Cwindowsize + 1) \Cwindowsize = \Cnpoints \Cwindowsize - \Cwindowsize^2 + \Cwindowsize,
\end{align*}

\noindent with \(1 \leq w \leq m\). Given $f(\Cwindowsize) = \Cnpoints \Cwindowsize - \Cwindowsize^2 + \Cwindowsize$, we find the critical point $\frac{df}{dw} = \Cnpoints - 2\Cwindowsize + 1 = 0$, which indicates that $\Cwindowsize = \frac{\Cnpoints + 1}{2}$ is a local maximum, given that the second derivative is negative. The maximum number of iterations is:

\begin{align*}
    f\left(\frac{m+1}{2}\right) &= m\left(\frac{m+1}{2}\right) -\left(\frac{m+1}{2}\right)^2 + \left(\frac{m+1}{2}\right)\\
    &= \frac{(m+1)^2}{4}.
\end{align*}

\noindent Thus the worst-case complexity is $O(\frac{(m+1)^2}{4}) = O(\Cnpoints^2)$.

\section{More Experiments}
\begin{figure}[h]%
\centering
\subfloat[Accuracy - UCR]{%
\includegraphics[width=.49\linewidth,  trim = 8mm 8mm 8mm 8mm, clip]{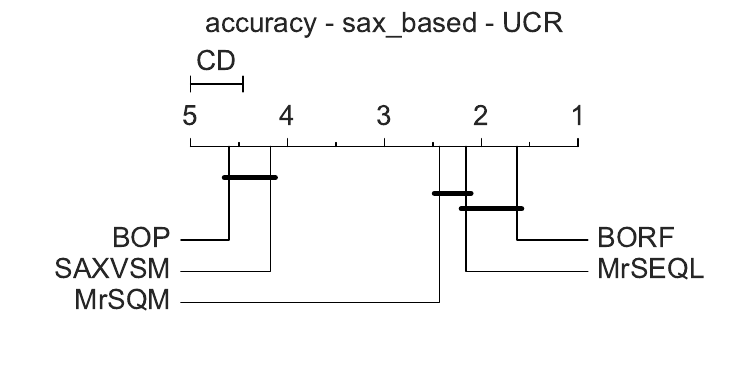}}%
~
\subfloat[Accuracy - UEA]{%
\includegraphics[width=.49\linewidth,  trim = 8mm 8mm 8mm 8mm, clip]{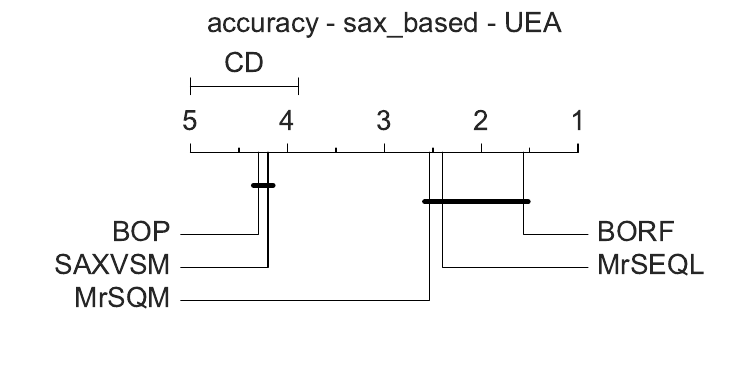}}\\
\subfloat[F1 - UCR]{%
\includegraphics[width=.49\linewidth,  trim = 8mm 8mm 8mm 8mm, clip]{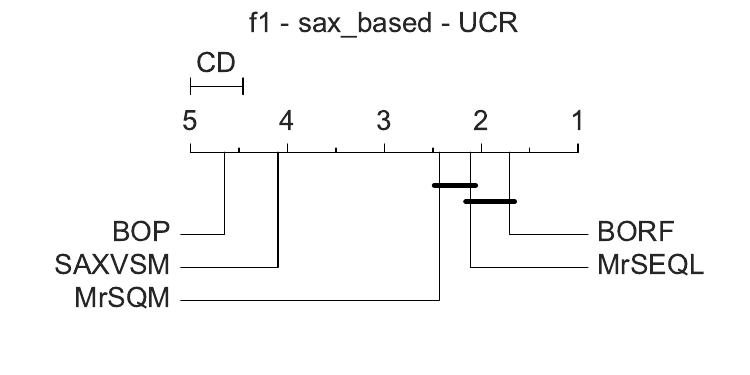}}%
~
\subfloat[F1 - UEA]{%
\includegraphics[width=.49\linewidth,  trim = 8mm 8mm 8mm 8mm, clip]{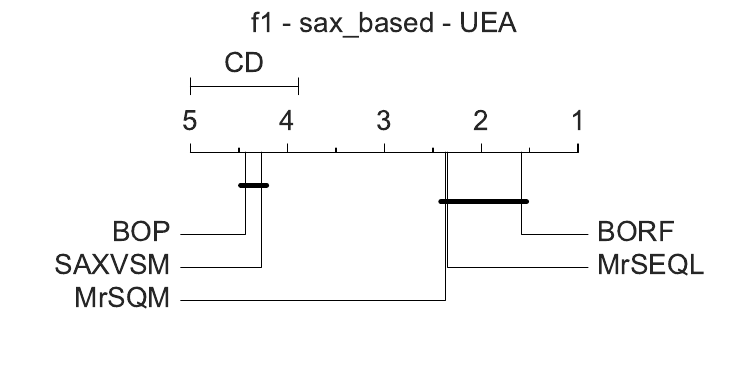}}\\
\subfloat[Runtime - UCR]{%
\includegraphics[width=.49\linewidth,  trim = 8mm 8mm 8mm 8mm, clip]{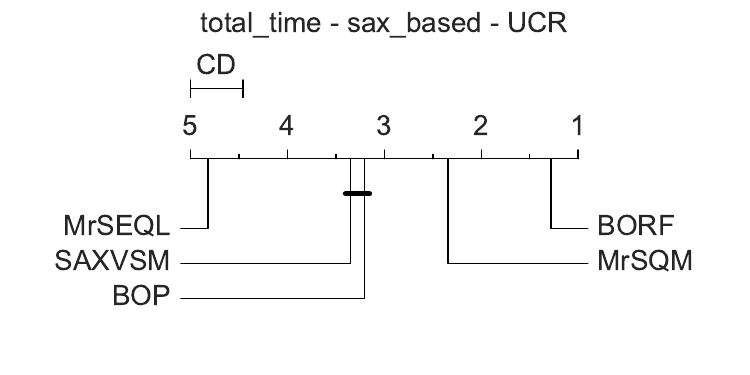}}%
~
\subfloat[Runtime - UEA]{%
\includegraphics[width=.49\linewidth,  trim = 8mm 8mm 8mm 8mm, clip]{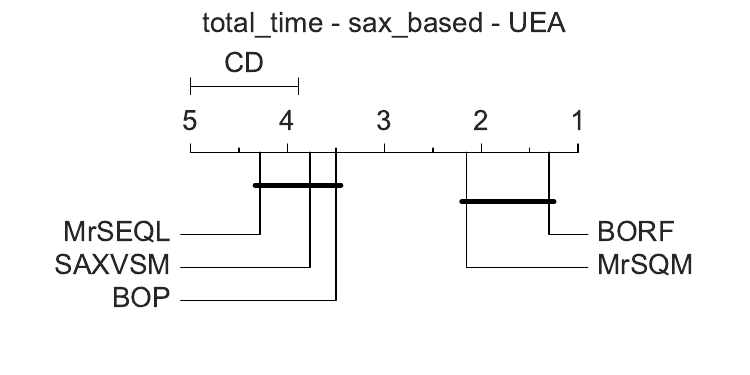}}
\caption{CD Plots for SAX-based methods. Best models to the right.}
\label{fig:cdplots_sax_divided}
\end{figure}

\begin{figure}[h]%
\centering
\subfloat[Accuracy - UCR]{%
\includegraphics[width=.49\linewidth,  trim = 8mm 8mm 5mm 7mm, clip]{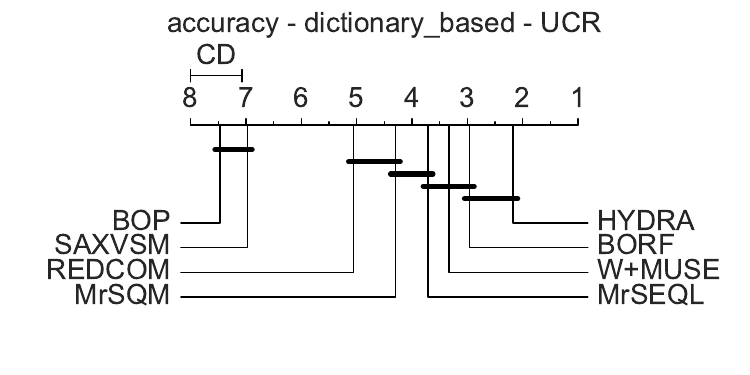}}%
~
\subfloat[Accuracy - UEA]{%
\includegraphics[width=.49\linewidth,  trim = 8mm 8mm 5mm 7mm, clip]{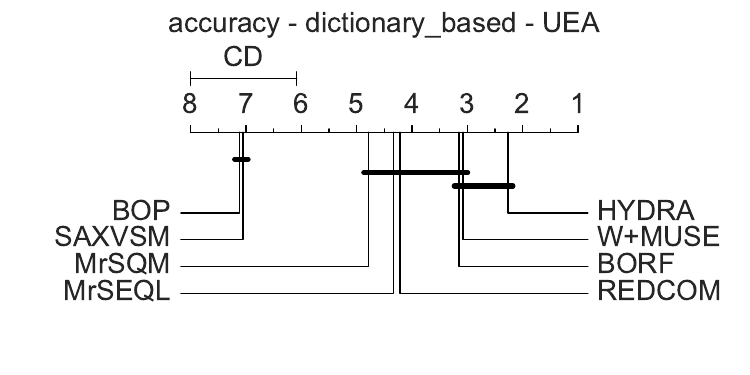}}\\
\subfloat[F1 - UCR]{%
\includegraphics[width=.49\linewidth,  trim = 8mm 8mm 5mm 7mm, clip]{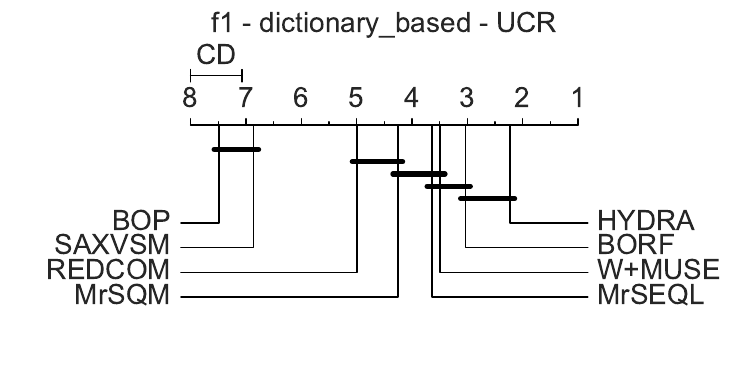}}%
~
\subfloat[F1 - UEA]{%
\includegraphics[width=.49\linewidth,  trim = 8mm 8mm 5mm 7mm, clip]{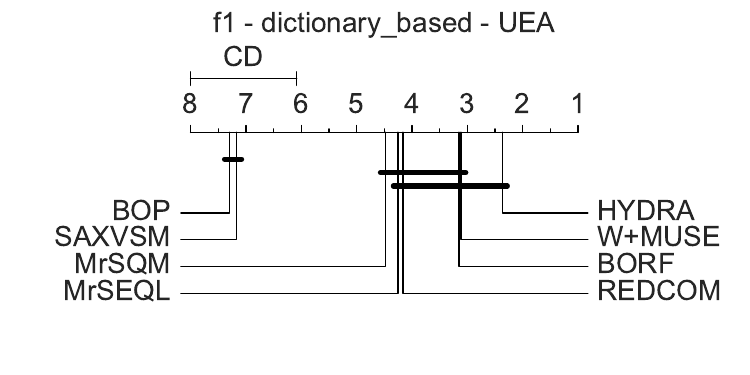}}\\
\subfloat[Runtime - UCR]{%
\includegraphics[width=.49\linewidth,  trim = 8mm 8mm 5mm 7mm, clip]{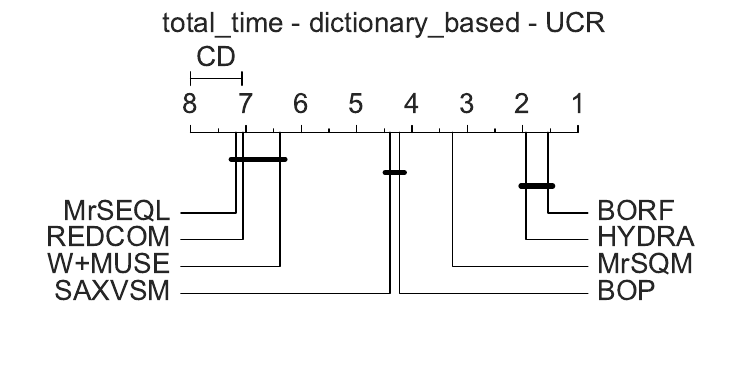}}%
~
\subfloat[Runtime - UEA]{%
\includegraphics[width=.49\linewidth,  trim = 8mm 8mm 5mm 7mm, clip]{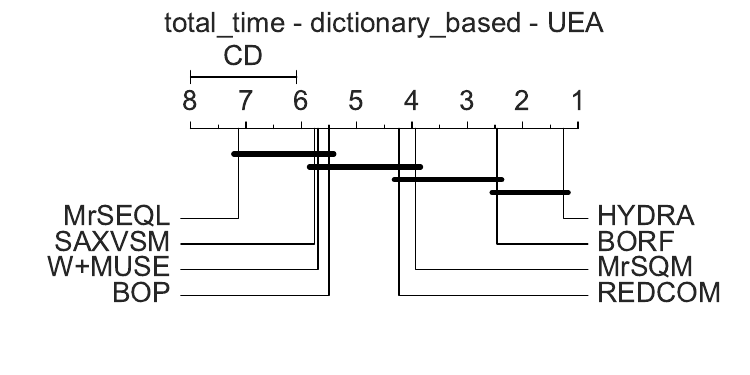}}
\caption{CD Plots for dictionary-based methods. Best models to the right.}
\label{fig:cdplots_dict_divided}
\end{figure}

\begin{figure}[h]%
\centering
\subfloat[Accuracy - UCR]{%
\includegraphics[width=.49\linewidth,  trim = 6mm 8mm 8mm 6mm, clip]{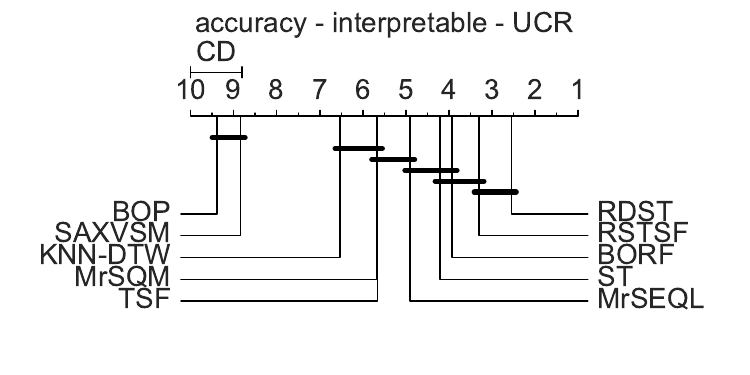}}%
~
\subfloat[Accuracy - UEA]{%
\includegraphics[width=.49\linewidth,  trim = 6mm 8mm 8mm 6mm, clip]{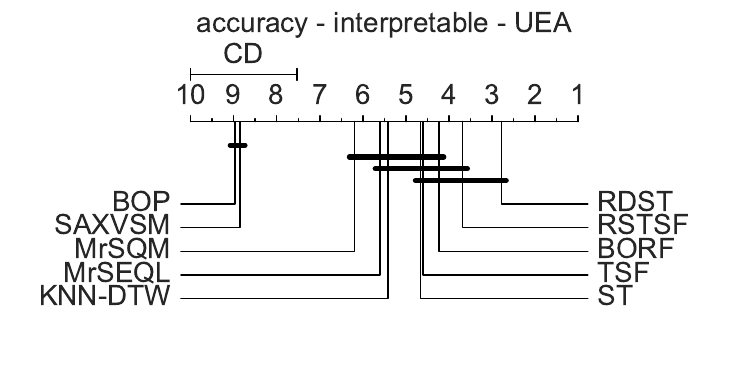}}\\
\subfloat[F1 - UCR]{%
\includegraphics[width=.49\linewidth,  trim = 6mm 8mm 8mm 6mm, clip]{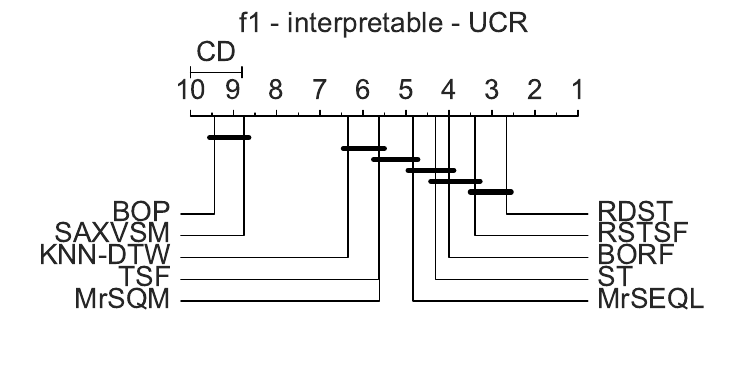}}%
~
\subfloat[F1 - UEA]{%
\includegraphics[width=.49\linewidth,  trim = 6mm 8mm 8mm 6mm, clip]{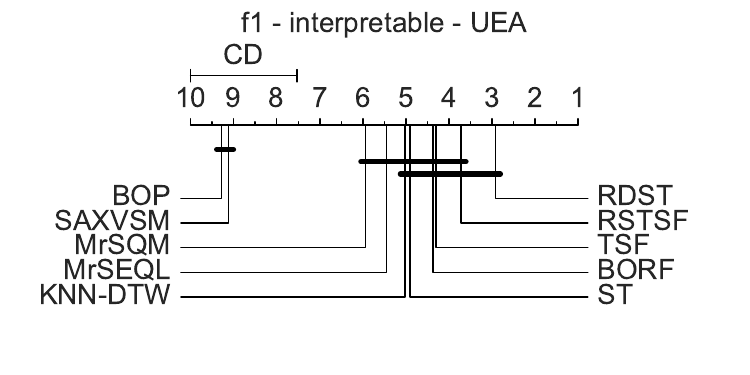}}\\
\subfloat[Runtime - UCR]{%
\includegraphics[width=.49\linewidth,  trim = 6mm 8mm 8mm 7mm, clip]{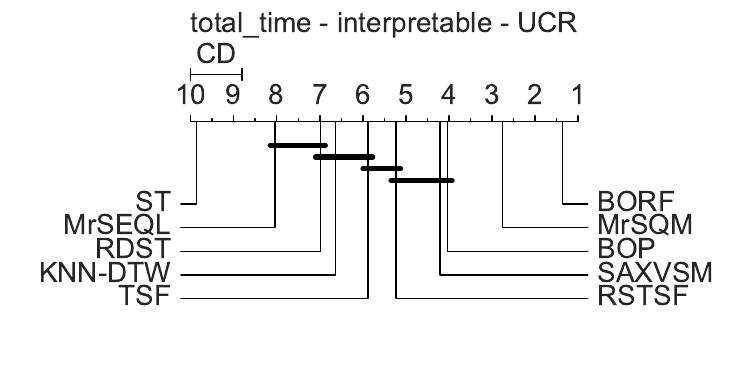}}%
~
\subfloat[Runtime - UEA]{%
\includegraphics[width=.49\linewidth,  trim = 6mm 8mm 8mm 7mm, clip]{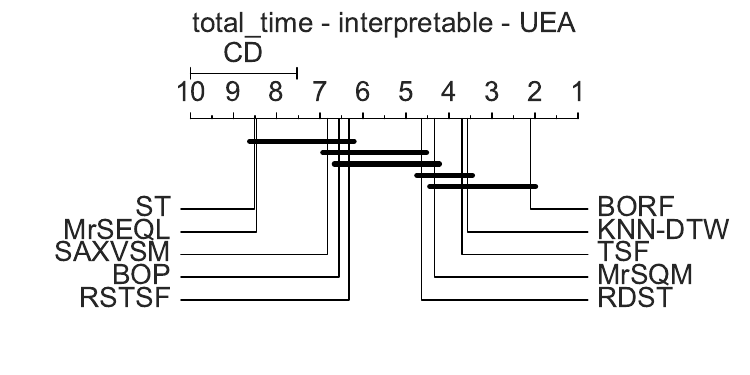}}
\caption{CD Plots for interpretable methods. Best models to the right.}
\label{fig:cdplots_int_divided}
\end{figure}

\begin{figure}[h]%
\centering
\subfloat[F1]{%
\includegraphics[width=\linewidth,  trim = 20mm 15mm 20mm 15mm, clip]{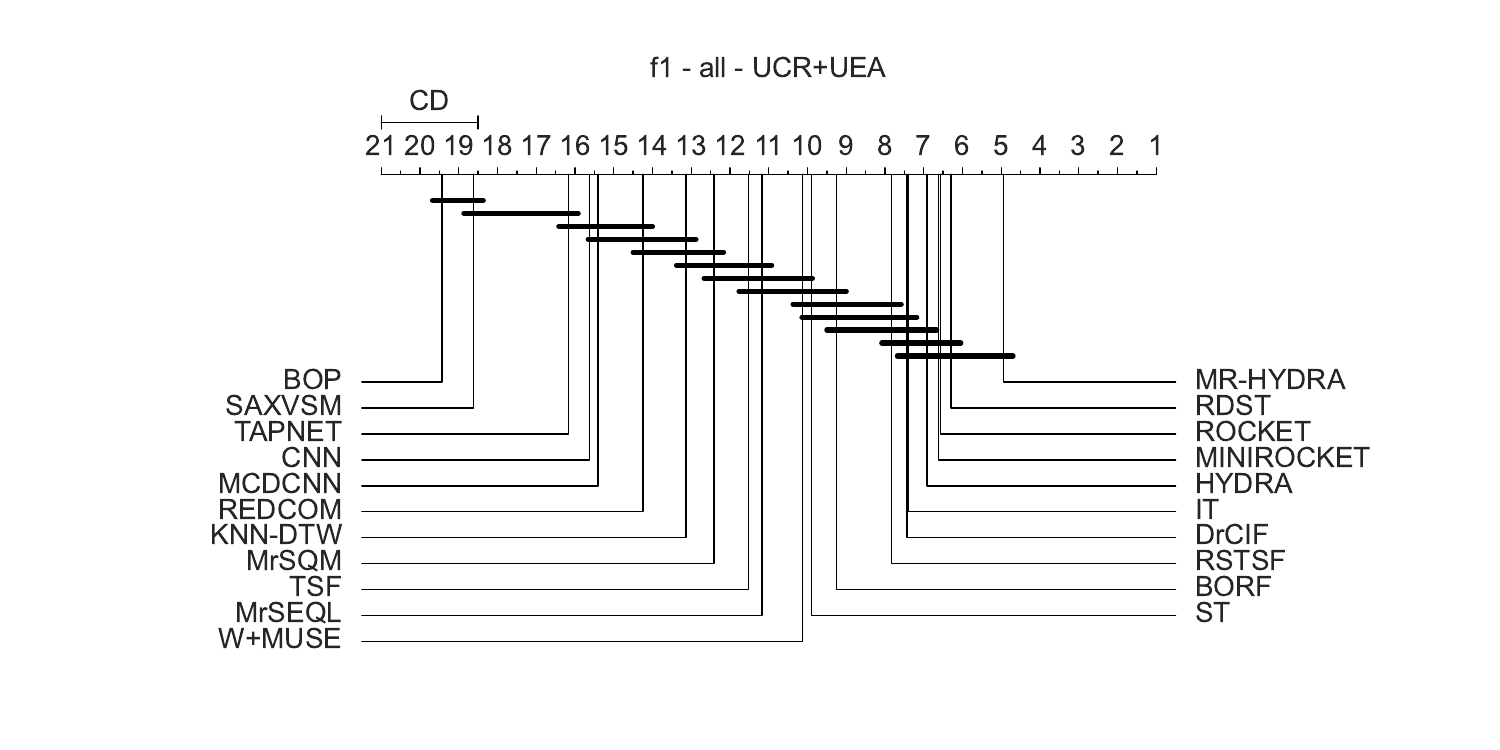}}\\
\caption{CD Plots for all benchmarked methods. Best models to the right.}
\label{fig:cdplots_all_divided}
\end{figure}

\begin{figure}[h]
    \centering
    \includegraphics[width=.99\columnwidth, trim = 0mm 0mm 0mm 0mm, clip]{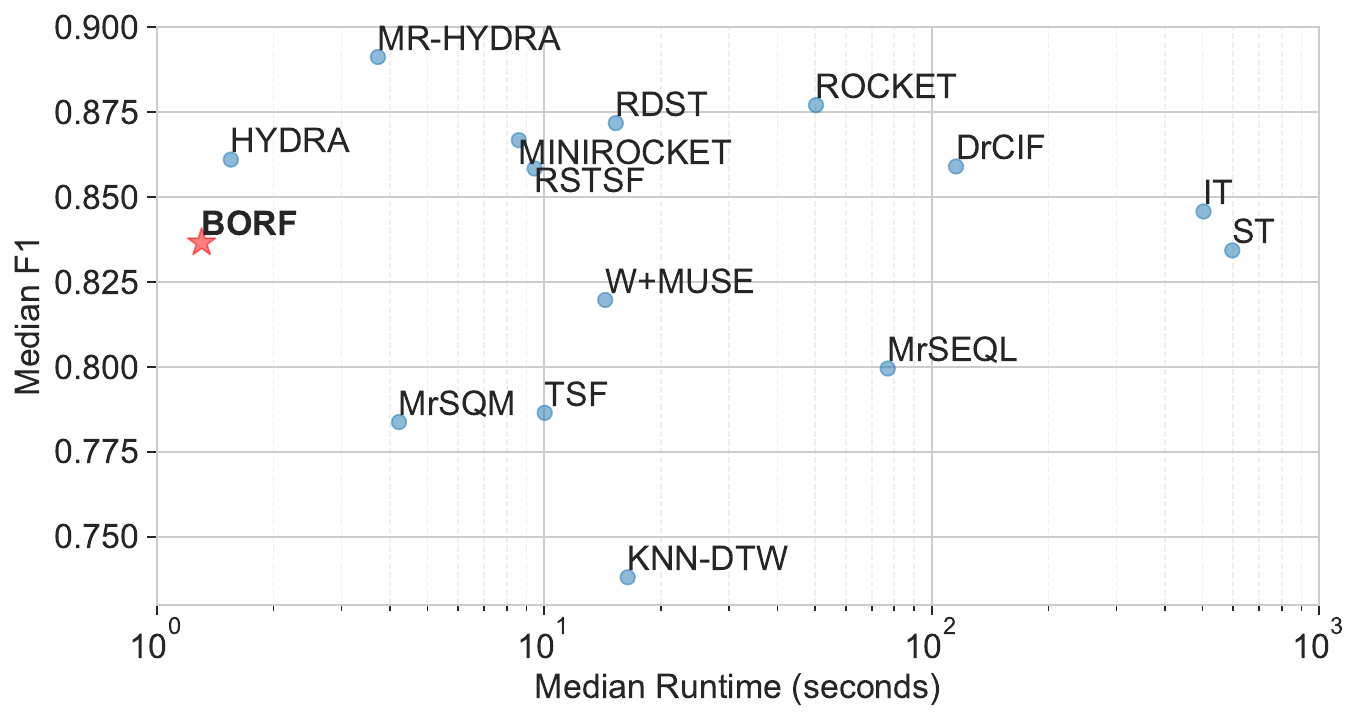}
    \caption{Comparison of median F1 performance against median runtime (seconds). Best models are on the top left.}
    \label{fig:perfvstime2}
\end{figure}

\begin{figure}[h]%
\centering
\includegraphics[width=\linewidth,  trim = 0mm 0mm 0mm 0mm, clip]{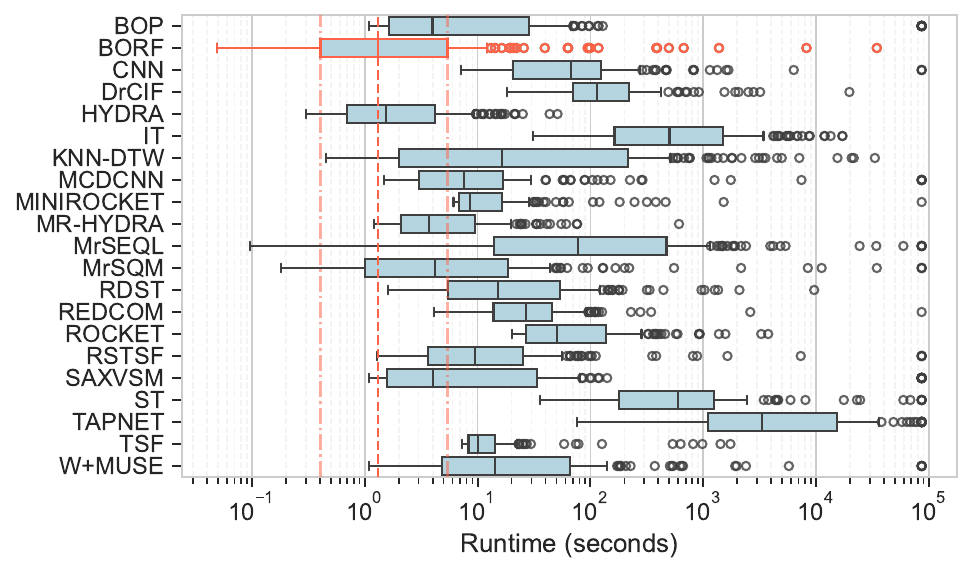}
\caption{Runtime boxplots (seconds), Lower is better. \Cmethodname{} median and interquartile range are highlighted in red.}
\label{fig:boxplots}
\end{figure}

In \Cref{fig:cdplots_sax_divided,fig:cdplots_dict_divided,fig:cdplots_int_divided,fig:cdplots_all_divided}, we report the CD plots from different classifier families for the UEA and UCR repositories separately, also showing performance in terms of F1 score. \Cref{fig:perfvstime2} shows a comparison between median runtime in seconds and median F1 score against competitor models. \Cref{fig:boxplots} shows the boxplot of runtime performance (in seconds). Finally, \Cref{tab:accuracy} reports the results in terms of accuracy of the top-13 models.

\label{sec:more_exp}

{
\begin{table*}
\setlength{\tabcolsep}{7.5pt}
\centering
\scriptsize
\caption{Accuracy of the top 13 best-performing models on all datasets. Missing values are due to exceeded runtime limits or out-of-memory errors.}
\label{tab:accuracy}
\begin{tabular}{l|ccccccccccccc}
\toprule
& {\Cmethod{BORF}} & \Cmethod{\shortstack{dr\\cif}} & \Cmethod{HYDRA} & \Cmethod{IT} & \Cmethod{\shortstack{{MINI}\\ROCKET}} & \Cmethod{\shortstack{{MR}\\HYDRA}} & \Cmethod{\shortstack{mr\\seql}} & \Cmethod{\shortstack{mr\\sqm}} & \Cmethod{RDST} & \Cmethod{ROCKET} & \Cmethod{\shortstack{rs\\tsf}} & \Cmethod{ST} & \Cmethod{\shortstack{W+\\MUSE}} \\
\midrule
ACSF1 & 0.610 & 0.880 & 0.850 & 0.920 & 0.910 & 0.880 & 0.710 & 0.770 & 0.920 & 0.880 & 0.890 & 0.890 & 0.920 \\
Adiac & 0.652 & 0.829 & 0.818 & 0.813 & 0.816 & 0.844 & 0.739 & 0.509 & 0.742 & 0.785 & 0.803 & 0.801 & 0.824 \\
AGW-X & 0.656 & 0.647 & 0.686 & 0.789 & 0.703 & 0.736 & 0.601 & 0.590 & 0.693 & 0.699 & 0.643 & 0.590 & 0.487 \\
AGW-Y & 0.706 & 0.661 & 0.750 & 0.817 & 0.769 & 0.791 & 0.680 & 0.624 & 0.754 & 0.774 & 0.711 & 0.626 & 0.563 \\
AGW-Z & 0.633 & 0.661 & 0.671 & 0.784 & 0.679 & 0.724 & 0.579 & 0.546 & 0.653 & 0.716 & 0.674 & 0.589 & 0.516 \\
ArrowHead & 0.829 & 0.823 & 0.834 & 0.834 & 0.863 & 0.863 & 0.754 & 0.709 & 0.857 & 0.823 & 0.749 & 0.806 & 0.874 \\
AWRecog & 0.993 & 0.983 & 0.990 & 0.983 & 0.990 & 0.993 & 0.967 & 0.950 & 0.993 & 0.993 & 0.980 & 0.967 & 0.993 \\
AFib & 0.267 & 0.400 & 0.133 & 0.200 & 0.133 & 0.067 & 0.400 & 0.267 & 0.200 & 0.067 & 0.333 & 0.267 & 0.200 \\
BME & 0.980 & 1.000 & 1.000 & 0.993 & 1.000 & 1.000 & 0.900 & 0.840 & 0.993 & 1.000 & 1.000 & 0.893 & 0.887 \\
BasicMotions & 0.975 & 1.000 & 1.000 & 1.000 & 1.000 & 1.000 & 1.000 & 0.900 & 1.000 & 1.000 & 1.000 & 1.000 & 1.000 \\
Beef & 0.733 & 0.833 & 0.833 & 0.667 & 0.833 & 0.767 & 0.600 & 0.767 & 0.867 & 0.800 & 0.733 & 0.733 & 0.833 \\
BeetleFly & 0.900 & 0.900 & 0.900 & 0.800 & 0.850 & 0.900 & 0.850 & 0.900 & 0.950 & 0.900 & 0.900 & 0.800 & 0.950 \\
BirdChick & 0.900 & 0.950 & 0.900 & 0.900 & 0.900 & 0.900 & 0.800 & 0.900 & 0.900 & 0.900 & 0.900 & 0.900 & 0.900 \\
CBF & 1.000 & 0.999 & 0.993 & 0.999 & 0.999 & 0.996 & 0.999 & 1.000 & 0.992 & 1.000 & 0.992 & 0.970 & 0.992 \\
Car & 0.917 & 0.867 & 0.933 & 0.883 & 0.917 & 0.933 & 0.850 & 0.750 & 0.917 & 0.900 & 0.833 & 0.833 & 0.900 \\
CharTraj & 0.969 & 0.981 & 0.986 & 0.995 & 0.993 & 0.992 & 0.978 & 0.953 & 0.992 & 0.991 & 0.985 & 0.974 & 0.984 \\
Chinatown & 0.962 & 0.985 & 0.983 & 0.985 & 0.983 & 0.980 & 0.950 & 0.968 & 0.977 & 0.980 & 0.983 & 0.971 & 0.939 \\
ChlConc & 0.721 & 0.738 & 0.760 & 0.870 & 0.755 & 0.789 & 0.718 & 0.682 & 0.754 & 0.812 & 0.743 & 0.729 & 0.772 \\
CinCECGT & 0.751 & 1.000 & 0.995 & 0.857 & 0.857 & 0.969 & 0.846 & 0.960 & 0.993 & 0.827 & 0.982 & 0.938 & 0.999 \\
Coffee & 1.000 & 1.000 & 1.000 & 1.000 & 1.000 & 1.000 & 1.000 & 0.964 & 0.964 & 1.000 & 1.000 & 1.000 & 1.000 \\
Computers & 0.772 & 0.732 & 0.716 & 0.816 & 0.732 & 0.772 & 0.728 & 0.736 & 0.760 & 0.772 & 0.752 & 0.720 & 0.732 \\
Cricket & 1.000 & 0.986 & 0.986 & 0.986 & 0.986 & 0.986 & 0.986 & 0.958 & 1.000 & 1.000 & 0.986 & 0.972 & 1.000 \\
CricketX & 0.821 & 0.746 & 0.800 & 0.800 & 0.821 & 0.803 & 0.769 & 0.705 & 0.821 & 0.826 & 0.746 & 0.800 & 0.718 \\
CricketY & 0.821 & 0.790 & 0.841 & 0.836 & 0.828 & 0.851 & 0.792 & 0.710 & 0.841 & 0.854 & 0.800 & 0.785 & 0.826 \\
CricketZ & 0.838 & 0.797 & 0.805 & 0.810 & 0.828 & 0.838 & 0.764 & 0.697 & 0.864 & 0.859 & 0.792 & 0.808 & 0.764 \\
Crop & 0.678 & 0.780 & 0.734 & 0.777 & 0.766 & 0.778 & 0.642 & 0.641 & 0.766 & 0.757 & 0.777 & - & 0.754 \\
DiatomSR & 0.980 & 0.928 & 0.948 & 0.944 & 0.928 & 0.964 & 0.941 & 0.889 & 0.922 & 0.977 & 0.928 & 0.938 & 0.889 \\
DP-OAG & 0.770 & 0.755 & 0.763 & 0.727 & 0.734 & 0.770 & 0.748 & 0.727 & 0.770 & 0.777 & 0.734 & 0.741 & 0.806 \\
DP-OC & 0.743 & 0.786 & 0.786 & 0.761 & 0.790 & 0.808 & 0.775 & 0.754 & 0.768 & 0.764 & 0.783 & 0.812 & 0.757 \\
DP-TW & 0.712 & 0.669 & 0.712 & 0.683 & 0.655 & 0.698 & 0.676 & 0.705 & 0.719 & 0.705 & 0.655 & 0.669 & 0.662 \\
DLDay & 0.537 & 0.637 & 0.512 & 0.512 & 0.662 & 0.600 & 0.613 & 0.487 & 0.625 & 0.588 & 0.625 & 0.550 & 0.350 \\
DLGame & 0.855 & 0.899 & 0.855 & 0.877 & 0.870 & 0.884 & 0.746 & 0.696 & 0.833 & 0.877 & 0.891 & 0.848 & 0.630 \\
DLWeekend & 0.978 & 0.986 & 0.978 & 0.978 & 0.986 & 0.978 & 0.949 & 0.957 & 0.978 & 0.971 & 0.986 & 0.964 & 0.906 \\
DuckDG & 0.440 & 0.480 & 0.680 & 0.620 & 0.700 & 0.640 & - & - & 0.420 & 0.460 & - & 0.560 & 0.460 \\
ECG200 & 0.900 & 0.870 & 0.870 & 0.940 & 0.910 & 0.910 & 0.890 & 0.830 & 0.900 & 0.920 & 0.910 & 0.900 & 0.850 \\
ECG5000 & 0.944 & 0.942 & 0.949 & 0.939 & 0.945 & 0.946 & 0.943 & 0.938 & 0.945 & 0.946 & 0.944 & 0.945 & 0.944 \\
ECG5D & 1.000 & 0.999 & 1.000 & 1.000 & 1.000 & 1.000 & 0.995 & 0.994 & 0.999 & 1.000 & 0.997 & 0.997 & 1.000 \\
EOG-HS & 0.610 & 0.599 & 0.572 & 0.633 & 0.613 & 0.633 & 0.630 & 0.398 & 0.677 & 0.616 & 0.599 & 0.619 & 0.500 \\
EOG-VS & 0.506 & 0.594 & 0.492 & 0.530 & 0.522 & 0.514 & 0.517 & 0.398 & 0.550 & 0.552 & 0.552 & 0.522 & 0.326 \\
ERing & 0.952 & 0.993 & 0.985 & 0.870 & 0.978 & 0.989 & 0.837 & 0.896 & 0.985 & 0.981 & 0.978 & 0.952 & 0.959 \\
Earthquakes & 0.748 & 0.748 & 0.734 & 0.727 & 0.727 & 0.741 & 0.683 & 0.705 & 0.734 & 0.741 & 0.748 & 0.748 & 0.748 \\
EigenWorms & 0.870 & 0.931 & 0.985 & 0.725 & 0.939 & 0.962 & - & 0.809 & 0.908 & 0.916 & 0.954 & 0.779 & 0.947 \\
ElecDev & 0.757 & 0.736 & 0.752 & 0.705 & 0.742 & 0.743 & 0.696 & 0.727 & 0.743 & 0.729 & 0.742 & - & 0.765 \\
Epilepsy & 0.993 & 1.000 & 1.000 & 0.993 & 1.000 & 1.000 & 0.986 & 0.993 & 0.993 & 0.986 & 1.000 & 1.000 & 1.000 \\
EthanolConc & 0.369 & 0.631 & 0.548 & 0.232 & 0.483 & 0.589 & 0.456 & 0.354 & 0.620 & 0.433 & 0.620 & 0.760 & 0.540 \\
EthanolLvl & 0.508 & 0.596 & 0.588 & 0.830 & 0.578 & 0.638 & 0.648 & 0.502 & 0.664 & 0.598 & 0.542 & 0.696 & 0.558 \\
FaceAll & 0.802 & 0.764 & 0.898 & 0.764 & 0.807 & 0.808 & 0.780 & 0.783 & 0.798 & 0.947 & 0.864 & 0.750 & 0.778 \\
FaceDet & 0.589 & 0.591 & 0.523 & 0.650 & 0.582 & 0.619 & - & 0.491 & 0.629 & 0.606 & - & - & - \\
FaceFour & 1.000 & 0.977 & 0.886 & 0.966 & 0.989 & 0.955 & 1.000 & 0.989 & 0.989 & 0.977 & 1.000 & 0.886 & 0.841 \\
FacesUCR & 0.942 & 0.897 & 0.956 & 0.961 & 0.960 & 0.960 & 0.929 & 0.893 & 0.967 & 0.961 & 0.903 & 0.889 & 0.906 \\
FiftyWords & 0.730 & 0.804 & 0.842 & 0.815 & 0.837 & 0.862 & 0.716 & 0.593 & 0.840 & 0.833 & 0.763 & 0.730 & 0.769 \\
FingMvmnt & 0.560 & 0.470 & 0.550 & 0.500 & 0.570 & 0.560 & 0.480 & 0.550 & 0.540 & 0.550 & 0.570 & 0.570 & 0.580 \\
Fish & 0.966 & 0.954 & 0.989 & 0.971 & 0.977 & 0.983 & 0.966 & 0.949 & 0.989 & 0.983 & 0.909 & 0.960 & 0.994 \\
FordA & 0.937 & 0.971 & 0.961 & 0.950 & 0.948 & 0.955 & 0.934 & 0.911 & 0.950 & 0.935 & 0.975 & 0.939 & - \\
FordB & 0.801 & 0.806 & 0.825 & 0.847 & 0.820 & 0.831 & 0.790 & 0.735 & 0.833 & 0.810 & 0.828 & 0.826 & - \\
FRTrain & 0.998 & 0.999 & 0.998 & 0.997 & 1.000 & 1.000 & 0.998 & 0.995 & 0.999 & 0.997 & 0.999 & 0.996 & 0.995 \\
FSTrain & 0.991 & 0.999 & 0.926 & 0.845 & 0.967 & 0.995 & 0.996 & 0.960 & 0.996 & 0.951 & 0.980 & 0.959 & 0.949 \\
Fungi & 1.000 & 0.930 & 1.000 & 0.995 & 1.000 & 1.000 & 0.930 & 0.806 & 1.000 & 1.000 & 0.919 & 0.839 & 1.000 \\
GMD1 & 0.746 & 0.723 & 0.754 & 0.692 & 0.715 & 0.769 & 0.762 & 0.646 & 0.785 & 0.785 & 0.731 & 0.769 & 0.615 \\
GMD2 & 0.662 & 0.685 & 0.638 & 0.562 & 0.631 & 0.669 & 0.677 & 0.577 & 0.715 & 0.723 & 0.615 & 0.646 & 0.462 \\
GMD3 & 0.469 & 0.554 & 0.500 & 0.377 & 0.392 & 0.492 & 0.500 & 0.377 & 0.554 & 0.485 & 0.531 & 0.577 & 0.323 \\
GPZ1 & 0.872 & 0.878 & 0.826 & 0.907 & 0.890 & 0.890 & 0.849 & 0.878 & 0.924 & 0.878 & 0.866 & 0.808 & 0.651 \\
GPZ2 & 0.873 & 0.810 & 0.810 & 0.835 & 0.835 & 0.823 & 0.854 & 0.810 & 0.861 & 0.842 & 0.804 & 0.816 & 0.620 \\
GunPoint & 1.000 & 0.993 & 1.000 & 1.000 & 0.993 & 1.000 & 0.993 & 0.987 & 1.000 & 1.000 & 0.973 & 1.000 & 1.000 \\
GP-AS & 0.987 & 0.991 & 1.000 & 0.975 & 0.997 & 1.000 & 0.991 & 0.994 & 0.997 & 0.997 & 0.991 & 0.984 & 1.000 \\
GP-MVF & 1.000 & 1.000 & 1.000 & 0.997 & 1.000 & 1.000 & 0.987 & 0.987 & 1.000 & 0.997 & 1.000 & 0.968 & 1.000 \\
GP-OY & 0.997 & 1.000 & 1.000 & 1.000 & 1.000 & 1.000 & 0.968 & 0.965 & 1.000 & 0.994 & 1.000 & 0.975 & 1.000 \\
Ham & 0.743 & 0.705 & 0.743 & 0.714 & 0.714 & 0.743 & 0.676 & 0.762 & 0.705 & 0.714 & 0.752 & 0.762 & 0.686 \\
HandMovDir & 0.284 & 0.514 & 0.216 & 0.405 & 0.392 & 0.365 & 0.351 & 0.351 & 0.324 & 0.514 & 0.459 & 0.338 & 0.311 \\
HandOutl & 0.930 & 0.922 & 0.941 & 0.962 & 0.935 & 0.946 & 0.943 & 0.911 & 0.946 & 0.943 & 0.914 & 0.922 & 0.924 \\
Handwrit & 0.520 & 0.362 & 0.466 & 0.648 & 0.518 & 0.514 & 0.461 & 0.322 & 0.600 & 0.587 & 0.365 & 0.449 & 0.366 \\
Haptics & 0.487 & 0.519 & 0.513 & 0.539 & 0.529 & 0.529 & 0.510 & 0.513 & 0.575 & 0.526 & 0.500 & 0.529 & 0.390 \\
Heartbeat & 0.741 & 0.790 & 0.761 & 0.678 & 0.741 & 0.732 & 0.702 & 0.727 & 0.756 & 0.756 & 0.780 & 0.732 & 0.746 \\
Herring & 0.609 & 0.578 & 0.750 & 0.656 & 0.688 & 0.734 & 0.609 & 0.578 & 0.688 & 0.594 & 0.578 & 0.672 & 0.641 \\
Hse20 & 0.992 & 0.916 & 0.958 & 0.941 & 0.958 & 0.992 & 0.899 & 0.941 & 0.966 & 0.966 & 0.916 & 0.899 & 0.975 \\
InlSkate & 0.405 & 0.531 & 0.491 & 0.489 & 0.476 & 0.496 & 0.445 & 0.329 & 0.447 & 0.469 & 0.615 & 0.425 & 0.578 \\
IEPG-RT & 0.988 & 1.000 & 1.000 & 1.000 & 1.000 & 1.000 & 0.976 & 0.948 & 1.000 & 1.000 & 1.000 & 0.996 & 1.000 \\
\bottomrule
\end{tabular}
\end{table*}
}
{
\begin{table*}\ContinuedFloat
\centering
\scriptsize
\setlength{\tabcolsep}{7.5pt}
\renewcommand{\arraystretch}{0.99}
\begin{tabular}{l|ccccccccccccc}
\toprule
& \Cmethod{BORF} & \Cmethod{\shortstack{dr\\cif}} & \Cmethod{HYDRA} & \Cmethod{IT} & \Cmethod{\shortstack{MINI\\ROCKET}} & \Cmethod{\shortstack{MR\\HYDRA}} & \Cmethod{\shortstack{mr\\seql}} & \Cmethod{\shortstack{mr\\sqm}} & \Cmethod{RDST} & \Cmethod{ROCKET} & \Cmethod{\shortstack{rs\\tsf}} & \Cmethod{ST} & \Cmethod{\shortstack{W+\\MUSE}} \\
\midrule
IEPG-ST & 0.928 & 1.000 & 1.000 & 1.000 & 0.992 & 1.000 & 0.896 & 0.795 & 0.992 & 0.980 & 1.000 & 0.976 & 1.000 \\
IWingImp & 0.519 & 0.664 & 0.679 & 0.692 & 0.673 & 0.667 & - & 0.303 & 0.482 & 0.399 & - & - & - \\
IWingSound & 0.633 & 0.682 & 0.655 & 0.627 & 0.672 & 0.667 & 0.599 & 0.552 & 0.665 & 0.666 & 0.668 & 0.648 & 0.627 \\
ItaPwrDem & 0.954 & 0.968 & 0.966 & 0.971 & 0.963 & 0.968 & 0.907 & 0.926 & 0.856 & 0.970 & 0.972 & 0.962 & 0.948 \\
JapVow & 0.795 & 0.973 & 0.978 & 0.989 & 0.992 & 0.978 & 0.965 & 0.930 & 0.965 & 0.854 & 0.965 & 0.762 & 0.954 \\
LSST & 0.603 & 0.565 & 0.607 & 0.506 & 0.651 & 0.636 & 0.681 & 0.576 & 0.647 & 0.636 & 0.628 & 0.639 & 0.603 \\
LrgKitchAppl & 0.923 & 0.821 & 0.883 & 0.888 & 0.875 & 0.896 & 0.872 & 0.915 & 0.824 & 0.896 & 0.811 & 0.851 & 0.672 \\
Libras & 0.844 & 0.894 & 0.933 & 0.878 & 0.917 & 0.928 & 0.817 & 0.800 & 0.944 & 0.906 & 0.917 & 0.872 & 0.906 \\
Lightn2 & 0.803 & 0.754 & 0.721 & 0.770 & 0.754 & 0.672 & 0.672 & 0.623 & 0.738 & 0.787 & 0.787 & 0.787 & 0.672 \\
Lightn7 & 0.795 & 0.753 & 0.808 & 0.822 & 0.795 & 0.822 & 0.795 & 0.630 & 0.781 & 0.822 & 0.753 & 0.753 & 0.644 \\
Mallat & 0.939 & 0.925 & 0.941 & 0.944 & 0.946 & 0.929 & 0.938 & 0.897 & 0.954 & 0.956 & 0.972 & 0.982 & 0.942 \\
Meat & 0.850 & 0.933 & 0.917 & 0.933 & 0.967 & 0.933 & 0.917 & 0.883 & 0.950 & 0.950 & 0.967 & 0.933 & 0.833 \\
MedImgs & 0.696 & 0.786 & 0.778 & 0.780 & 0.791 & 0.797 & 0.701 & 0.617 & 0.763 & 0.807 & 0.826 & 0.742 & 0.734 \\
MelbPed & 0.836 & 0.964 & 0.926 & 0.972 & 0.966 & 0.967 & 0.812 & 0.845 & 0.965 & 0.908 & 0.971 & 0.932 & 0.887 \\
MP-OAG & 0.623 & 0.610 & 0.597 & 0.487 & 0.552 & 0.584 & 0.519 & 0.552 & 0.571 & 0.552 & 0.623 & 0.571 & 0.571 \\
MP-OC & 0.759 & 0.832 & 0.835 & 0.797 & 0.849 & 0.845 & 0.729 & 0.691 & 0.808 & 0.838 & 0.845 & 0.838 & 0.790 \\
MP-TW & 0.591 & 0.591 & 0.539 & 0.506 & 0.526 & 0.513 & 0.532 & 0.487 & 0.558 & 0.539 & 0.591 & 0.526 & 0.539 \\
MxShpRT & 0.952 & 0.956 & 0.981 & 0.967 & 0.971 & 0.982 & 0.963 & 0.962 & 0.983 & 0.973 & 0.952 & 0.962 & 0.986 \\
MxShpST & 0.920 & 0.922 & 0.965 & 0.904 & 0.950 & 0.967 & 0.903 & 0.930 & 0.967 & 0.936 & 0.898 & 0.938 & 0.960 \\
MoteStrain & 0.946 & 0.932 & 0.929 & 0.888 & 0.927 & 0.946 & 0.894 & 0.923 & 0.918 & 0.917 & 0.951 & 0.903 & 0.933 \\
MotImag & 0.460 & 0.500 & 0.500 & 0.570 & 0.500 & 0.480 & - & - & 0.540 & 0.540 & 0.500 & 0.510 & 0.570 \\
NATOPS & 0.917 & 0.828 & 0.872 & 0.967 & 0.944 & 0.917 & 0.911 & 0.867 & 0.878 & 0.872 & 0.861 & 0.878 & 0.922 \\
NIFECGT1 & 0.902 & 0.931 & 0.943 & 0.910 & 0.949 & 0.964 & 0.925 & 0.780 & 0.924 & 0.958 & 0.926 & 0.927 & 0.913 \\
NIFECGT2 & 0.916 & 0.935 & 0.958 & 0.959 & 0.964 & 0.968 & 0.946 & 0.838 & 0.938 & 0.966 & 0.942 & 0.947 & 0.926 \\
OSULeaf & 0.950 & 0.847 & 0.988 & 0.897 & 0.959 & 0.963 & 0.942 & 0.967 & 0.988 & 0.942 & 0.872 & 0.880 & 0.988 \\
OliveOil & 0.900 & 0.933 & 0.933 & 0.833 & 0.933 & 0.967 & 0.867 & 0.933 & 0.900 & 0.900 & 0.933 & 0.800 & 0.867 \\
PEMS-SF & 0.908 & 1.000 & 0.803 & 0.809 & 0.827 & 0.775 & - & - & 0.919 & 0.844 & 1.000 & 0.971 & - \\
PLAID & 0.885 & 0.907 & 0.819 & 0.534 & 0.877 & 0.909 & 0.883 & 0.890 & 0.905 & 0.858 & 0.888 & 0.784 & 0.736 \\
PenDigits & 0.920 & 0.973 & 0.968 & 0.987 & - & 0.968 & 0.872 & 0.941 & 0.985 & 0.979 & - & - & 0.953 \\
POC & 0.748 & 0.844 & 0.823 & 0.831 & 0.840 & 0.844 & 0.742 & 0.703 & 0.819 & 0.837 & 0.844 & 0.837 & 0.807 \\
Phoneme & 0.317 & 0.400 & 0.324 & 0.334 & 0.294 & 0.354 & 0.276 & 0.341 & 0.341 & 0.281 & 0.369 & 0.346 & 0.339 \\
PhonSpect & 0.251 & 0.296 & 0.329 & 0.306 & 0.317 & 0.332 & 0.232 & 0.220 & 0.313 & 0.279 & 0.287 & 0.307 & - \\
PGWZ & 0.700 & 0.720 & 0.700 & 0.640 & 0.780 & 0.780 & 0.680 & 0.680 & 0.780 & 0.740 & 0.720 & 0.580 & 0.520 \\
PAP & 0.659 & 0.245 & 0.769 & 0.933 & 0.865 & 0.712 & 0.587 & 0.957 & 0.822 & 0.394 & 0.245 & 0.668 & 0.851 \\
PAPress & 0.986 & 0.885 & 0.986 & 1.000 & 0.981 & 0.957 & 0.904 & 0.971 & 0.942 & 0.962 & 0.750 & 0.962 & 1.000 \\
PCVP & 0.952 & 0.591 & 0.966 & 0.947 & 0.947 & 0.894 & 0.707 & 0.952 & 0.938 & 0.933 & 0.644 & 0.846 & 0.976 \\
Plane & 1.000 & 1.000 & 1.000 & 1.000 & 1.000 & 1.000 & 1.000 & 1.000 & 1.000 & 1.000 & 1.000 & 1.000 & 1.000 \\
PwrCons & 0.889 & 1.000 & 0.978 & 1.000 & 0.994 & 0.983 & 0.878 & 0.872 & 0.994 & 0.933 & 1.000 & 0.983 & 0.933 \\
PP-OAG & 0.844 & 0.849 & 0.868 & 0.829 & 0.834 & 0.868 & 0.820 & 0.844 & 0.844 & 0.859 & 0.854 & 0.834 & 0.844 \\
PP-OC & 0.859 & 0.897 & 0.911 & 0.918 & 0.904 & 0.924 & 0.804 & 0.811 & 0.883 & 0.900 & 0.911 & 0.893 & 0.900 \\
PP-TW & 0.800 & 0.795 & 0.810 & 0.776 & 0.800 & 0.805 & 0.815 & 0.761 & 0.820 & 0.805 & 0.810 & 0.746 & 0.815 \\
RacketS & 0.908 & 0.888 & 0.914 & 0.882 & 0.855 & 0.908 & 0.816 & 0.829 & 0.882 & 0.908 & 0.901 & 0.822 & 0.862 \\
RefDev & 0.608 & 0.613 & 0.528 & 0.533 & 0.480 & 0.517 & 0.616 & 0.635 & 0.568 & 0.512 & 0.600 & 0.536 & 0.552 \\
Rock & 0.880 & 0.700 & 0.920 & 0.640 & 0.820 & 0.860 & 0.820 & 0.880 & 0.880 & 0.900 & 0.800 & 0.940 & 0.800 \\
ScrnType & 0.515 & 0.555 & 0.493 & 0.603 & 0.461 & 0.565 & 0.552 & 0.509 & 0.509 & 0.467 & 0.541 & 0.533 & 0.472 \\
SRSCP1 & 0.758 & 0.870 & 0.867 & 0.829 & 0.918 & 0.952 & 0.635 & 0.652 & 0.898 & 0.853 & 0.867 & 0.836 & 0.799 \\
SRSCP2 & 0.472 & 0.511 & 0.589 & 0.550 & 0.544 & 0.567 & 0.533 & 0.533 & 0.511 & 0.528 & 0.567 & 0.467 & 0.600 \\
SHG-Ch2 & 0.892 & 0.932 & 0.847 & 0.865 & 0.888 & 0.950 & 0.840 & 0.805 & 0.923 & 0.928 & 0.973 & 0.957 & 0.797 \\
SHM-Ch2 & 0.638 & 0.849 & 0.496 & 0.456 & 0.676 & 0.773 & 0.638 & 0.367 & 0.702 & 0.629 & 0.856 & 0.876 & 0.284 \\
SHS-Ch2 & 0.867 & 0.913 & 0.771 & 0.711 & 0.867 & 0.902 & 0.882 & 0.758 & 0.853 & 0.887 & 0.922 & 0.949 & 0.611 \\
SWGZ & 0.820 & 0.860 & 0.840 & 0.780 & 0.860 & 0.900 & 0.760 & 0.880 & 0.880 & 0.900 & 0.800 & 0.860 & 0.680 \\
ShapeSim & 1.000 & 0.972 & 0.989 & 0.956 & 1.000 & 1.000 & 0.672 & 1.000 & 1.000 & 1.000 & 0.972 & 0.978 & 1.000 \\
ShapesAll & 0.897 & 0.867 & 0.928 & 0.917 & 0.923 & 0.928 & 0.928 & 0.822 & 0.927 & 0.905 & 0.868 & 0.838 & 0.945 \\
SmKitchAppl & 0.757 & 0.821 & 0.808 & 0.763 & 0.827 & 0.805 & 0.837 & 0.797 & 0.827 & 0.808 & 0.832 & 0.811 & 0.773 \\
SmthSubsp & 0.880 & 0.980 & 0.927 & 0.993 & 0.953 & 0.980 & 0.800 & 0.913 & 0.987 & 0.973 & 0.980 & 0.953 & 0.800 \\
SonySurf1 & 0.759 & 0.917 & 0.920 & 0.907 & 0.887 & 0.887 & 0.747 & 0.790 & 0.892 & 0.925 & 0.899 & 0.937 & 0.517 \\
SonySurf2 & 0.935 & 0.918 & 0.948 & 0.955 & 0.921 & 0.942 & 0.879 & 0.937 & 0.901 & 0.919 & 0.881 & 0.925 & 0.940 \\
SpArabDig & 0.986 & 0.978 & 0.989 & 0.993 & 0.993 & 0.993 & 0.961 & 0.933 & 0.997 & 0.996 & 0.983 & 0.945 & 0.986 \\
SWJ & 0.400 & 0.400 & 0.400 & 0.467 & 0.400 & 0.600 & 0.667 & 0.667 & 0.467 & 0.467 & 0.467 & 0.467 & 0.400 \\
StarLC & 0.976 & 0.981 & 0.982 & 0.979 & 0.982 & 0.981 & 0.980 & 0.969 & 0.983 & 0.981 & 0.980 & 0.978 & 0.979 \\
Strawberry & 0.968 & 0.965 & 0.973 & 0.849 & 0.984 & 0.976 & 0.962 & 0.954 & 0.976 & 0.981 & 0.968 & 0.973 & 0.973 \\
SweLeaf & 0.926 & 0.974 & 0.978 & 0.970 & 0.962 & 0.976 & 0.942 & 0.918 & 0.965 & 0.962 & 0.958 & 0.917 & 0.971 \\
Symbols & 0.956 & 0.967 & 0.980 & 0.983 & 0.983 & 0.981 & 0.963 & 0.972 & 0.980 & 0.973 & 0.968 & 0.942 & 0.974 \\
SynthCtrl & 0.977 & 1.000 & 0.957 & 0.997 & 0.987 & 0.997 & 0.957 & 0.967 & 0.993 & 1.000 & 0.993 & 0.997 & 0.980 \\
ToeSeg1 & 0.969 & 0.925 & 0.961 & 0.956 & 0.961 & 0.956 & 0.965 & 0.947 & 0.969 & 0.952 & 0.899 & 0.956 & 0.961 \\
ToeSeg2 & 0.923 & 0.885 & 0.908 & 0.954 & 0.923 & 0.915 & 0.869 & 0.908 & 0.908 & 0.931 & 0.885 & 0.815 & 0.938 \\
Trace & 1.000 & 1.000 & 1.000 & 1.000 & 1.000 & 1.000 & 1.000 & 1.000 & 1.000 & 1.000 & 1.000 & 0.990 & 1.000 \\
2LeadECG & 0.999 & 0.996 & 0.999 & 0.995 & 0.998 & 0.998 & 0.993 & 0.998 & 0.996 & 0.999 & 0.986 & 0.996 & 1.000 \\
TwoPatt & 0.995 & 1.000 & 0.987 & 1.000 & 0.997 & 1.000 & 0.998 & 0.921 & 1.000 & 1.000 & 0.997 & 0.988 & 0.996 \\
UMD & 0.993 & 0.972 & 0.993 & 0.993 & 0.993 & 0.993 & 0.938 & 0.972 & 1.000 & 0.993 & 0.986 & 0.972 & 0.986 \\
UWGL & 0.944 & 0.938 & 0.909 & 0.891 & 0.931 & 0.941 & 0.778 & 0.591 & 0.931 & 0.941 & 0.906 & 0.891 & 0.909 \\
UWGL-All & 0.940 & 0.972 & 0.973 & 0.941 & 0.971 & 0.980 & 0.919 & 0.734 & 0.980 & 0.976 & 0.961 & 0.960 & 0.889 \\
UWGL-X & 0.811 & 0.843 & 0.854 & 0.827 & 0.848 & 0.875 & 0.807 & 0.691 & 0.854 & 0.853 & 0.829 & 0.811 & 0.826 \\
UWGL-Y & 0.731 & 0.769 & 0.792 & 0.753 & 0.776 & 0.810 & 0.711 & 0.619 & 0.764 & 0.779 & 0.758 & 0.731 & 0.732 \\
UWGL-Z & 0.748 & 0.781 & 0.803 & 0.765 & 0.801 & 0.821 & 0.759 & 0.642 & 0.800 & 0.797 & 0.776 & 0.780 & 0.768 \\
Wafer & 1.000 & 0.999 & 1.000 & 0.999 & 0.999 & 1.000 & 0.995 & 1.000 & 1.000 & 0.998 & 1.000 & 1.000 & 0.999 \\
Wine & 0.648 & 0.815 & 0.907 & 0.667 & 0.852 & 0.926 & 0.741 & 0.815 & 0.889 & 0.778 & 0.870 & 0.944 & 0.833 \\
WordSyn & 0.654 & 0.680 & 0.743 & 0.730 & 0.760 & 0.777 & 0.629 & 0.487 & 0.745 & 0.752 & 0.636 & 0.654 & 0.657 \\
Worms & 0.688 & 0.753 & 0.753 & 0.805 & 0.740 & 0.779 & 0.766 & 0.753 & 0.688 & 0.727 & 0.753 & 0.766 & 0.805 \\
Worms2C & 0.727 & 0.805 & 0.766 & 0.766 & 0.779 & 0.779 & 0.805 & 0.740 & 0.740 & 0.805 & 0.792 & 0.844 & 0.792 \\
Yoga & 0.857 & 0.873 & 0.927 & 0.904 & 0.910 & 0.929 & 0.869 & 0.761 & 0.932 & 0.910 & 0.877 & 0.890 & 0.934 \\
\bottomrule
\end{tabular}
\end{table*}
}

\bibliographystyle{unsrt}
\bibliography{bibliography}

@article{lin2007experiencing,
  title={Experiencing SAX: a novel symbolic representation of time series},
  author={Lin, Jessica and Keogh, Eamonn and Wei, Li and Lonardi, Stefano},
  journal={Data Mining and knowledge discovery},
  volume={15},
  pages={107--144},
  year={2007},
  publisher={Springer}
}

@article{lin2012rotation,
  title={Rotation-invariant similarity in time series using bag-of-patterns representation},
  author={Lin, Jessica and Khade, Rohan and Li, Yuan},
  journal={Journal of Intelligent Information Systems},
  volume={39},
  pages={287--315},
  year={2012},
  publisher={Springer}
}

@inproceedings{senin2013sax,
  title={Sax-vsm: Interpretable time series classification using sax and vector space model},
  author={Senin, Pavel and Malinchik, Sergey},
  booktitle={2013 IEEE 13th international conference on data mining},
  pages={1175--1180},
  year={2013},
  organization={IEEE}
}

@inproceedings{nguyen2022fast,
  title={Fast time series classification with random symbolic subsequences},
  author={Nguyen, Thach Le and Ifrim, Georgiana},
  booktitle={International Workshop on Advanced Analytics and Learning on Temporal Data},
  pages={50--65},
  year={2022},
  organization={Springer}
}

@article{schafer2017multivariate,
  title={Multivariate time series classification with WEASEL+ MUSE},
  author={Sch{\"a}fer, Patrick and Leser, Ulf},
  journal={arXiv preprint arXiv:1711.11343},
  year={2017}
}

@inproceedings{schafer2012sfa,
  title={SFA: a symbolic fourier approximation and index for similarity search in high dimensional datasets},
  author={Sch{\"a}fer, Patrick and H{\"o}gqvist, Mikael},
  booktitle={Proceedings of the 15th international conference on extending database technology},
  pages={516--527},
  year={2012}
}

@article{deng2013time,
  title={A time series forest for classification and feature extraction},
  author={Deng, Houtao and Runger, George and Tuv, Eugene and Vladimir, Martyanov},
  journal={Information Sciences},
  volume={239},
  pages={142--153},
  year={2013},
  publisher={Elsevier}
}

@article{baydogan2013bag,
  title={A bag-of-features framework to classify time series},
  author={Baydogan, Mustafa Gokce and Runger, George and Tuv, Eugene},
  journal={IEEE transactions on pattern analysis and machine intelligence},
  volume={35},
  number={11},
  pages={2796--2802},
  year={2013},
  publisher={IEEE}
}

@inproceedings{middlehurst2020canonical,
  title={The canonical interval forest (CIF) classifier for time series classification},
  author={Middlehurst, Matthew and Large, James and Bagnall, Anthony},
  booktitle={2020 IEEE international conference on big data (big data)},
  pages={188--195},
  year={2020},
  organization={IEEE}
}

@inproceedings{lines2012shapelet,
  title={A shapelet transform for time series classification},
  author={Lines, Jason and Davis, Luke M and Hills, Jon and Bagnall, Anthony},
  booktitle={Proceedings of the 18th ACM SIGKDD international conference on Knowledge discovery and data mining},
  pages={289--297},
  year={2012}
}

@article{ye2011time,
  title={Time series shapelets: a novel technique that allows accurate, interpretable and fast classification},
  author={Ye, Lexiang and Keogh, Eamonn},
  journal={Data mining and knowledge discovery},
  volume={22},
  pages={149--182},
  year={2011},
  publisher={Springer}
}

@inproceedings{grabocka2014learning,
  title={Learning time-series shapelets},
  author={Grabocka, Josif and Schilling, Nicolas and Wistuba, Martin and Schmidt-Thieme, Lars},
  booktitle={Proceedings of the 20th ACM SIGKDD international conference on Knowledge discovery and data mining},
  pages={392--401},
  year={2014}
}

@article{karlsson2016generalized,
  title={Generalized random shapelet forests},
  author={Karlsson, Isak and Papapetrou, Panagiotis and Bostr{\"o}m, Henrik},
  journal={Data mining and knowledge discovery},
  volume={30},
  pages={1053--1085},
  year={2016},
  publisher={Springer}
}

@article{bagnall2017great,
  title={The great time series classification bake off: a review and experimental evaluation of recent algorithmic advances},
  author={Bagnall, Anthony and Lines, Jason and Bostrom, Aaron and Large, James and Keogh, Eamonn},
  journal={Data mining and knowledge discovery},
  volume={31},
  pages={606--660},
  year={2017},
  publisher={Springer}
}

@article{dempster2020rocket,
  title={ROCKET: exceptionally fast and accurate time series classification using random convolutional kernels},
  author={Dempster, Angus and Petitjean, Fran{\c{c}}ois and Webb, Geoffrey I},
  journal={Data Mining and Knowledge Discovery},
  volume={34},
  number={5},
  pages={1454--1495},
  year={2020},
  publisher={Springer}
}

@article{ruiz2021great,
  title={The great multivariate time series classification bake off: a review and experimental evaluation of recent algorithmic advances},
  author={Ruiz, Alejandro Pasos and Flynn, Michael and Large, James and Middlehurst, Matthew and Bagnall, Anthony},
  journal={Data Mining and Knowledge Discovery},
  volume={35},
  number={2},
  pages={401--449},
  year={2021},
  publisher={Springer}
}

@article{le2019interpretable,
  title={Interpretable time series classification using linear models and multi-resolution multi-domain symbolic representations},
  author={Le Nguyen, Thach and Gsponer, Severin and Ilie, Iulia and O’reilly, Martin and Ifrim, Georgiana},
  journal={Data mining and knowledge discovery},
  volume={33},
  pages={1183--1222},
  year={2019},
  publisher={Springer}
}

@article{theissler2022explainable,
  title={Explainable AI for Time Series Classification: A review, taxonomy and research directions},
  author={Theissler, Andreas and Spinnato, Francesco and Schlegel, Udo and Guidotti, Riccardo},
  journal={IEEE Access},
  year={2022},
  publisher={IEEE}
}

@article{dau2019ucr,
  title={The UCR time series archive},
  author={Dau, Hoang Anh and Bagnall, Anthony and Kamgar, Kaveh and Yeh, Chin-Chia Michael and Zhu, Yan and Gharghabi, Shaghayegh and Ratanamahatana, Chotirat Ann and Keogh, Eamonn},
  journal={IEEE/CAA Journal of Automatica Sinica},
  volume={6},
  number={6},
  pages={1293--1305},
  year={2019},
  publisher={IEEE}
}

@article{bagnall2018uea,
  title={The UEA multivariate time series classification archive, 2018},
  author={Bagnall, Anthony and Dau, Hoang Anh and Lines, Jason and Flynn, Michael and Large, James and Bostrom, Aaron and Southam, Paul and Keogh, Eamonn},
  journal={arXiv preprint arXiv:1811.00075},
  year={2018}
}

@article{dumoulin2016guide,
  title={A guide to convolution arithmetic for deep learning},
  author={Dumoulin, Vincent and Visin, Francesco},
  journal={arXiv preprint arXiv:1603.07285},
  year={2016}
}

@article{keogh2001dimensionality,
  title={Dimensionality reduction for fast similarity search in large time series databases},
  author={Keogh, Eamonn and Chakrabarti, Kaushik and Pazzani, Michael and Mehrotra, Sharad},
  journal={Knowledge and information Systems},
  volume={3},
  pages={263--286},
  year={2001},
  publisher={Springer}
}

@article{lundberg2017unified,
  title={A unified approach to interpreting model predictions},
  author={Lundberg, Scott M and Lee, Su-In},
  journal={Advances in neural information processing systems},
  volume={30},
  year={2017}
}

@inproceedings{hung2007combining,
  title={Combining sax and piecewise linear approximation to improve similarity search on financial time series},
  author={Hung, Nguyen Quoc Viet and Anh, Duong Tuan},
  booktitle={2007 International Symposium on Information Technology Convergence (ISITC 2007)},
  pages={58--62},
  year={2007},
  organization={IEEE}
}

@article{welford1962note,
author = {B. P. Welford},
title = {Note on a Method for Calculating Corrected Sums of Squares and Products},
journal = {Technometrics},
volume = {4},
number = {3},
pages = {419--420},
year = {1962},
publisher = {Taylor \& Francis},
doi = {10.1080/00401706.1962.10490022},
}

@book{knuth2014art,
  title={The Art of Computer Programming: Seminumerical Algorithms, Volume 2},
  author={Knuth, Donald E},
  year={2014},
  publisher={Addison-Wesley Professional}
}

@article{ismail2020inceptiontime,
  title={Inceptiontime: Finding alexnet for time series classification},
  author={Ismail Fawaz, Hassan and Lucas, Benjamin and Forestier, Germain and Pelletier, Charlotte and Schmidt, Daniel F and Weber, Jonathan and Webb, Geoffrey I and Idoumghar, Lhassane and Muller, Pierre-Alain and Petitjean, Fran{\c{c}}ois},
  journal={Data Mining and Knowledge Discovery},
  volume={34},
  number={6},
  pages={1936--1962},
  year={2020},
  publisher={Springer}
}

@article{harris1954distributional,
	title        = {Distributional structure},
	author       = {Harris, Zellig S},
	year         = 1954,
	journal      = {Word},
	publisher    = {Taylor \& Francis},
	volume       = 10,
	number       = {2-3},
	pages        = {146--162}
}

@inproceedings{dempster2021minirocket,
	title        = {Minirocket: A very fast (almost) deterministic transform for time series classification},
	author       = {Dempster, Angus and Schmidt, Daniel F and Webb, Geoffrey I},
	year         = 2021,
	booktitle    = {Proceedings of the 27th ACM SIGKDD Conference on Knowledge Discovery \& Data Mining},
	pages        = {248--257}
}

@article{blankertz2001classifying,
  title={Classifying single trial EEG: Towards brain computer interfacing},
  author={Blankertz, Benjamin and Curio, Gabriel and M{\"u}ller, Klaus-Robert},
  journal={Advances in neural information processing systems},
  volume={14},
  year={2001}
}

@article{fauzi2020pyts,
  author  = {Johann Faouzi and Hicham Janati},
  title   = {pyts: A Python Package for Time Series Classification},
  journal = {Journal of Machine Learning Research},
  year    = {2020},
  volume  = {21},
  number  = {46},
  pages   = {1-6},
  url     = {http://jmlr.org/papers/v21/19-763.html}
}

@article{loning2019sktime,
  title={sktime: A unified interface for machine learning with time series},
  author={L{\"o}ning, Markus and Bagnall, Anthony and Ganesh, Sajaysurya and Kazakov, Viktor and Lines, Jason and Kir{\'a}ly, Franz J},
  journal={arXiv preprint arXiv:1909.07872},
  year={2019}
}

@inproceedings{zhang2020tapnet,
  title={Tapnet: Multivariate time series classification with attentional prototypical network},
  author={Zhang, Xuchao and Gao, Yifeng and Lin, Jessica and Lu, Chang-Tien},
  booktitle={Proceedings of the AAAI conference on artificial intelligence},
  volume={34},
  pages={6845--6852},
  year={2020}
}

@article{middlehurst2024bakeredux,
	title = {Bake off redux: a review and experimental evaluation of recent time series classification algorithms},
	issn = {1384-5810, 1573-756X},
	shorttitle = {Bake off redux},
	url = {https://link.springer.com/10.1007/s10618-024-01022-1},
	doi = {10.1007/s10618-024-01022-1},
	language = {en},
	urldate = {2024-06-19},
	journal = {Data Mining and Knowledge Discovery},
	author = {Middlehurst, Matthew and Schäfer, Patrick and Bagnall, Anthony},
	month = apr,
	year = {2024},
}

@article{middlehurst2021hive,
  title={HIVE-COTE 2.0: a new meta ensemble for time series classification},
  author={Middlehurst, Matthew and Large, James and Flynn, Michael and Lines, Jason and Bostrom, Aaron and Bagnall, Anthony},
  journal={Machine Learning},
  volume={110},
  number={11},
  pages={3211--3243},
  year={2021},
  publisher={Springer}
}

@article{dempster2023hydra,
  title={Hydra: Competing convolutional kernels for fast and accurate time series classification},
  author={Dempster, Angus and Schmidt, Daniel F and Webb, Geoffrey I},
  journal={Data Mining and Knowledge Discovery},
  volume={37},
  number={5},
  pages={1779--1805},
  year={2023},
  publisher={Springer}
}

@inproceedings{guillaume2022random,
  title={Random dilated shapelet transform: A new approach for time series shapelets},
  author={Guillaume, Antoine and Vrain, Christel and Elloumi, Wael},
  booktitle={International Conference on Pattern Recognition and Artificial Intelligence},
  pages={653--664},
  year={2022},
  organization={Springer}
}

@article{cabello2024fast,
  title={Fast, accurate and explainable time series classification through randomization},
  author={Cabello, Nestor and Naghizade, Elham and Qi, Jianzhong and Kulik, Lars},
  journal={Data Mining and Knowledge Discovery},
  volume={38},
  number={2},
  pages={748--811},
  year={2024},
  publisher={Springer}
}

@article{tan2022multirocket,
  title={MultiRocket: multiple pooling operators and transformations for fast and effective time series classification},
  author={Tan, Chang Wei and Dempster, Angus and Bergmeir, Christoph and Webb, Geoffrey I},
  journal={Data Mining and Knowledge Discovery},
  volume={36},
  number={5},
  pages={1623--1646},
  year={2022},
  publisher={Springer}
}

@inproceedings{middlehurst2023extracting,
  title={Extracting features from random subseries: A hybrid pipeline for time series classification and extrinsic regression},
  author={Middlehurst, Matthew and Bagnall, Anthony},
  booktitle={International Workshop on Advanced Analytics and Learning on Temporal Data},
  pages={113--126},
  year={2023},
  organization={Springer}
}

@inproceedings{bostrom2015binary,
  title={Binary shapelet transform for multiclass time series classification},
  author={Bostrom, Aaron and Bagnall, Anthony},
  booktitle={Big Data Analytics and Knowledge Discovery: 17th International Conference, DaWaK 2015, Valencia, Spain, September 1-4, 2015, Proceedings 17},
  pages={257--269},
  year={2015},
  organization={Springer}
}

@article{schafer2023weasel,
  title={WEASEL 2.0: a random dilated dictionary transform for fast, accurate and memory constrained time series classification},
  author={Sch{\"a}fer, Patrick and Leser, Ulf},
  journal={Machine Learning},
  volume={112},
  number={12},
  pages={4763--4788},
  year={2023},
  publisher={Springer}
}

@inproceedings{schafer2017fast,
  title={Fast and accurate time series classification with weasel},
  author={Sch{\"a}fer, Patrick and Leser, Ulf},
  booktitle={Proceedings of the 2017 ACM on Conference on Information and Knowledge Management},
  pages={637--646},
  year={2017}
}

@inproceedings{ismail2022deep,
  title={Deep learning for time series classification using new hand-crafted convolution filters},
  author={Ismail-Fawaz, Ali and Devanne, Maxime and Weber, Jonathan and Forestier, Germain},
  booktitle={2022 IEEE International Conference on Big Data (Big Data)},
  pages={972--981},
  year={2022},
  organization={IEEE}
}

@article{shifaz2020ts,
  title={TS-CHIEF: a scalable and accurate forest algorithm for time series classification},
  author={Shifaz, Ahmed and Pelletier, Charlotte and Petitjean, Fran{\c{c}}ois and Webb, Geoffrey I},
  journal={Data Mining and Knowledge Discovery},
  volume={34},
  number={3},
  pages={742--775},
  year={2020},
  publisher={Springer}
}

@article{lubba2019catch22,
  title={catch22: CAnonical Time-series CHaracteristics: Selected through highly comparative time-series analysis},
  author={Lubba, Carl H and Sethi, Sarab S and Knaute, Philip and Schultz, Simon R and Fulcher, Ben D and Jones, Nick S},
  journal={Data Mining and Knowledge Discovery},
  volume={33},
  number={6},
  pages={1821--1852},
  year={2019},
  publisher={Springer}
}

@article{ismail2023approach, title={An approach to multiple comparison benchmark evaluations that is stable under manipulation of the comparate set}, author={Ismail-Fawaz, Ali and Dempster, Angus and Tan, Chang Wei and Herrmann, Matthieu and Miller, Lynn and Schmidt, Daniel F and Berretti, Stefano and Weber, Jonathan and Devanne, Maxime and Forestier, Germain and others}, journal={arXiv preprint arXiv:2305.11921}, year={2023} }

@inproceedings{bennett2023red,
  title={RED CoMETS: An ensemble classifier for symbolically represented multivariate time series},
  author={Bennett, Luca A and Abdallah, Zahraa S},
  booktitle={International Workshop on Advanced Analytics and Learning on Temporal Data},
  pages={76--91},
  year={2023},
  organization={Springer}
}

@article{middlehurst2024aeon,
  title={aeon: a Python toolkit for learning from time series},
  author={Middlehurst, Matthew and Ismail-Fawaz, Ali and Guillaume, Antoine and Holder, Christopher and Rubio, David Guijo and Bulatova, Guzal and Tsaprounis, Leonidas and Mentel, Lukasz and Walter, Martin and Sch{\"a}fer, Patrick and others},
  journal={arXiv preprint arXiv:2406.14231},
  year={2024}
}

@inproceedings{zheng2014time,
  title={Time series classification using multi-channels deep convolutional neural networks},
  author={Zheng, Yi and Liu, Qi and Chen, Enhong and Ge, Yong and Zhao, J Leon},
  booktitle={International conference on web-age information management},
  pages={298--310},
  year={2014},
  organization={Springer}
}

@book{saito1994local,
  title={Local feature extraction and its applications using a library of bases},
  author={Saito, Naoki},
  year={1994},
  publisher={Yale University}
}

\begin{IEEEbiography}[{\includegraphics[width=1in,height=1.25in,clip,keepaspectratio]{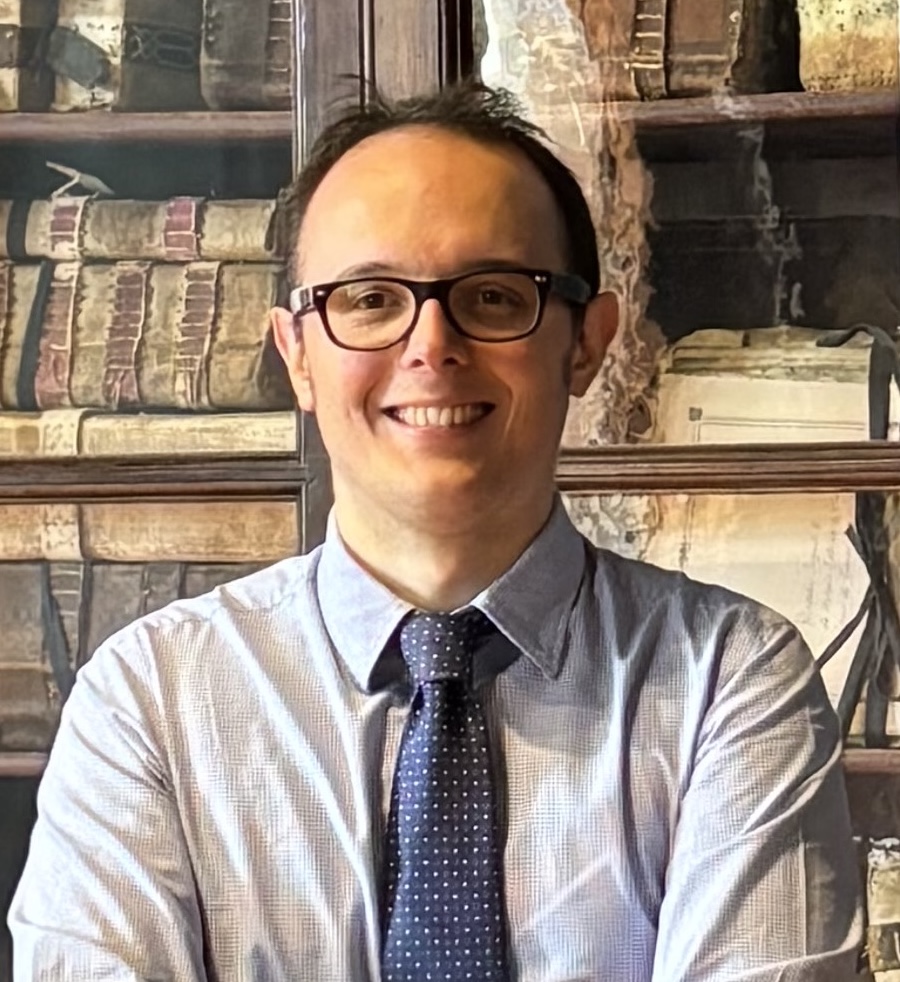}}]{Francesco Spinnato} earned his Bachelor's degree in Economics and Management from the University of Padua in 2017, and he received his Master's degree in Data Science from the University of Pisa in 2020. He obtained his Ph.D. in Data Science from the Scuola Normale Superiore in 2024.

He is currently a researcher at the University of Pisa. His research focuses on explainable AI (XAI) applied to sequential data, particularly on interpreting black-box models for univariate and multivariate time series. In 2024, he collaborated with the Visualization and Data Analytics (VIDA) Research Center at New York University.

Dr. Spinnato was a recipient of the 2023-2024 Socint G-Research Italian National Prize for his Ph.D. thesis: \textit{``Explanation Methods for Sequential Data Models''}.
\end{IEEEbiography}

\begin{IEEEbiography}[{\includegraphics[width=1in,height=1.25in,clip,keepaspectratio]{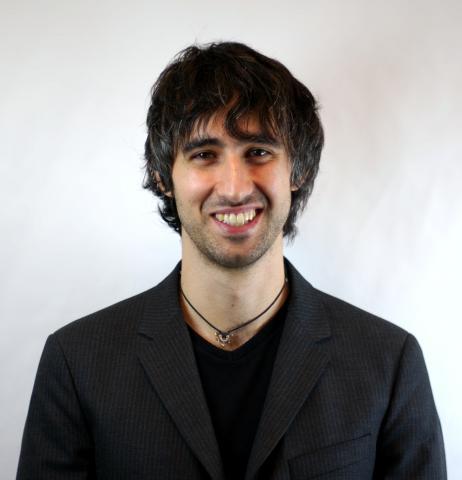}}]{Riccardo Guidotti} received the B.S. and M.S. degrees in computer science, both with cum laude, from the University of Pisa, Pisa, Italy, in 2010 and 2013, respectively. He received his Ph.D. degree in computer science from the same institution, where his thesis focused on personal data analytics.

He is currently an Assistant Professor (RTD-B) at the Department of Computer Science, University of Pisa, and a member of the Knowledge Discovery and Data Mining Laboratory (KDDLab), a joint research group with the Information Science and Technology Institute of the National Research Council in Pisa. His research interests include personal data mining, clustering, explainable models, and the analysis of transactional data.

Dr. Guidotti won the IBM fellowship program and was an intern at IBM Research, Dublin, Ireland, in 2015.
\end{IEEEbiography}

\begin{IEEEbiography}[{\includegraphics[width=1in,height=1.25in,clip,keepaspectratio]{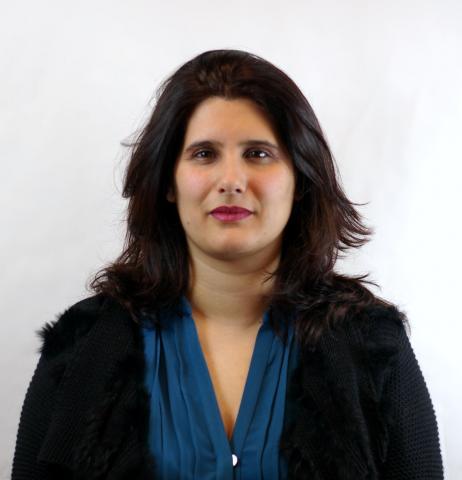}}]{Anna Monreale} received her Ph.D. degree in computer science from the University of Pisa, Pisa, Italy, in June 2011, with a dissertation on privacy-by-design in data mining. 

She is currently an Associate Professor at the Department of Computer Science, University of Pisa, and a member of the Knowledge Discovery and Data Mining Laboratory (KDD-Lab), a joint research group with the Information Science and Technology Institute of the National Research Council in Pisa.
Her research interests include big data analytics, social networks, and the privacy issues arising in mining these types of social and human-sensitive data. She focuses particularly on evaluating privacy risks during analytical processes and designing privacy-by-design technologies in the era of big data.

\end{IEEEbiography}

\begin{IEEEbiography}[{\includegraphics[width=1in,height=1.25in,clip,keepaspectratio]{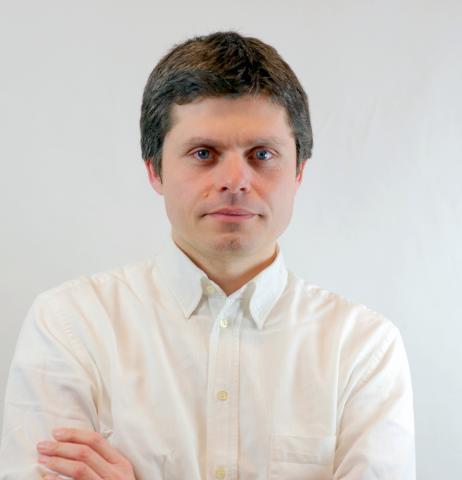}}]{Mirco Nanni} 
holds a degree and a PhD in Computer Science (University of Pisa, in 1997 and 2002). He is currently a senior researcher at ISTI - CNR in Pisa, head of the KDD Laboratory. 
His research interests cover several aspects of human mobility, such as data mining on spatio-temporal data, applications on transportation and smart cities, mobile phone data analysis and Big Data for Official Statistics.

\end{IEEEbiography}

\EOD

\end{document}